\documentclass[10pt]{article} % For LaTeX2e

\usepackage[preprint]{rlj} % Should be uncommented for preprint versions

\usepackage{amssymb}   
\usepackage{mathtools}   
\usepackage{mathrsfs}    
\usepackage{graphicx}      
\usepackage{subcaption}   
\usepackage[space]{grffile}   
\usepackage{url}          
\usepackage{lipsum}

\usepackage{amsthm} 

\usepackage{algorithm2e}
\RestyleAlgo{ruled}
\SetKwComment{Comment}{/* }{ */}
\usepackage{algpseudocode}
\usepackage{arydshln}
\usepackage{tabularray}

\newcommand{\R}{\ensuremath{\mathbb{R}}}   % reelle Zahlen
\newcommand{\IE}{\ensuremath{\mathbb{E}}}

\def\A{{\cal A}}
\def\S{{\cal S}}

\newtheorem{lemma*}{Lemma}
\theoremstyle{definition}

\definecolor{forestgreen}{rgb}{0.13, 0.55, 0.13}
\definecolor{deepred}{rgb}{0.7215, 0.0, 0.5882}
\definecolor{royalpurple}{rgb}{0.0, 0.0, 0.5}

\title{On the Effect of Regularization in Policy Mirror Descent}
\setrunningtitle{On the Effect of Regularization in Policy Mirror Descent}

\author{Jan Felix Kleuker\textsuperscript{1}, Aske Plaat\textsuperscript{1}, Thomas Moerland\textsuperscript{1}}

\emails{kleukerjf@vuw.leidenuniv.nl}
\affiliations{
$^{1}$\textbf{Leiden Institute of Advanced Computer Science, Leiden University, the Netherlands}}

\contribution{
    We conduct an empirical analysis of the interaction between the regularization components in PMD, running over 500k training seeds and providing insights into the fragility of these algorithms to regularization temperatures.
    }
    {
    Empirical studies on PMD-based RL algorithms are limited.
    }

\contribution{
    We examined the impact of temperature scales across different reward scales and identified simple heuristics to aid in tuning these temperatures. Additionally, we evaluated dynamic adaptation strategies for temperature parameters, indicating that maintaining constant values is often more effective than adapting them.
    }
    {
    RL algorithms are sensitive to hyperparameter tuning, making the selection of appropriate temperatures challenging.
    }

\contribution{
    We examine the effects of different regularization combinations on robustness to temperature tuning and performance, indicating a notable impact.
    }
    {
    To the best of our knowledge, no existing study examines the effects of the interplay between these two components for different choices, leaving a gap in the literature.
    }

\keywords{Policy Mirror Descent, Regularization, Reinforcement Learning} % Your keywords

\summary{Policy Mirror Descent (PMD) has emerged as a unifying framework in reinforcement learning (RL) by linking policy gradient methods with a first-order optimization method known as mirror descent. At its core, PMD incorporates two key regularization components: (i) a distance term that enforces a trust region for stable policy updates and (ii) an MDP regularizer that augments the reward function to promote structure and robustness. While PMD has been extensively studied in theory, empirical investigations remain scarce. This work provides a large-scale empirical analysis of the interplay between these two regularization techniques, running over 500k training seeds on small RL environments.
}

\begin{document}

% \makeCover  % Create the cover page
\maketitle  % Make the title section

\begin{abstract}
    Policy Mirror Descent (PMD) has emerged as a unifying framework in reinforcement learning (RL) by linking policy gradient methods with a first-order optimization method known as mirror descent. At its core, PMD incorporates two key regularization components: (i) a distance term that enforces a trust region for stable policy updates and (ii) an MDP regularizer that augments the reward function to promote structure and robustness. 
    While PMD has been extensively studied in theory, empirical investigations remain scarce. This work provides a large-scale empirical analysis of the interplay between these two regularization techniques, running over 500k training seeds on small RL environments. Our results demonstrate that, although the two regularizers can partially substitute each other, their precise combination is critical for achieving robust performance. These findings highlight the potential for advancing research on more robust algorithms in RL, particularly with respect to hyperparameter sensitivity.
\end{abstract}

\section{Introduction}
\label{sec:Intro}

Recent research has revealed a deep connection between policy gradient methods in RL and mirror descent — a first-order optimization technique \citep{beckMirr2003, beck2017first}. This insight has led to the development of the PMD framework \citep{geistTheo2019, tomarMirr2021, kubaMirr, lanPoli2023, xiaoConv, vaswaniGene2023, alfano2023novel, zhan2023policy}, which encompasses a broad class of algorithms distinguished by their choice of regularization, such as Trust Region Learning (TRL; \citealt{schulman2015trust}), or Soft Actor-Critic (SAC; \citealt{haarnojaSoft2018}).

In the PMD framework, two regularization components are central. The first is a distance term, referred to as the \textit{Drift regularizer}, that ensures the updated policy remains sufficiently close to its predecessor -- an idea that was prominently implemented in trust region policy optimization (TRPO; \citealt{schulman2015trust}), or proximal policy optimization (PPO; \citealt{schulman2017proximal}). This trust-region idea has been further generalized through the introduction of a Drift functional \citep{kubaMirr}, motivating the term Drift regularizer. The second component is a convex \textit{MDP regularizer} that augments the reward function with a structure-promoting term, an approach motivated by the objectives of enhanced exploration and robustness \citep{ziebart2010modeling, haarnojaRein2017, chow2018path, leeTsal2019}. A prominent example hereof is the negative Shannon-Entropy, a core component in soft Reinforcement Learning \citep{haarnojaRein2017, haarnojaSoft2018}.

While extensive theoretical work has established strong convergence results for PMD, these guarantees largely disappear in approximate settings where the exact value function is inaccessible. On the other hand, only a limited number of numerical experiments have been conducted to validate its practical performance \citep{vieillardLeve2021, tomarMirr2021, alfano2023novel}. Notably, empirical studies have primarily focused on specific cases, such as the (reverse) KL divergence or learning Drift regularizers \citep{luDisc2022, alfanolearning}.
%
%
%
%
%

% \paragraph*{Contribution}
In this work, we complement existing empirical studies by systematically analyzing the impact of different regularization components on algorithmic performance of RL algorithms based on PMD. Our main contributions are:
\begin{itemize}
    \item We analyze the interaction between two distinct regularization techniques in RL, a study that, to the best of our knowledge, has not been conducted before. To this end, we run over 500k training seeds on small RL environments, demonstrating brittleness of these algorithms to regularization temperatures.
    \item We investigate the effects of varying temperature parameters both during training and across different reward scales, providing insights to support the design of algorithms that mitigate fragility to regularization temperatures.
    \item Leveraging recent theoretical advancements, we examine the effects of different regularization combinations, particularly their influence on robustness to temperature tuning. Our findings suggest this aspect may be understudied in the existing literature.
\end{itemize}

The rest of the paper is organized as follows: In Section \ref{sec:Background}, we introduce the necessary background on Policy Mirror Descent. Section \ref{sec::pract.alg} details the methodology of our numerical experiments, followed by a discussion of the results in Section \ref{sec:Experiments}. Finally, we conclude with key findings in Section \ref{sec:conclusion}.

% Study empirically interplay (should not only be stated, but studied)
% connect theory to practical algorithms
% Framing of the contributions: I would stress you 1) present, to the best of your knowledge, the first empirical study of the interplay of these two key RL regularizers, and 2) bridge recent theoretical results on appropriate combinations of both. 
% \begin{itemize}
%     \item We empirically investigate the interplay between two distinct regularization techniques in RL, addressing an open question raised in \citet{vieillardLeve2021}. To the best of our knowledge, this analysis has not been conducted before. To this end, we run over 500k training seeds on small RL environments.
%     \item We bridge recent theoretical insights with empirical analysis to deepen our understanding of regularization effects. Particularly their interacting impact on robustness, suggesting that this aspect may be understudied.
%     \item We examine the effect of varying regularization levels, referred to as temperatures, during training and across different environments, providing insights to aid researchers in their algorithm design.
% \end{itemize}
%
%
%
%
%
%
\section{Background}
\label{sec:Background}

% General Definitions
We assume a standard Markov Decision Process (MDP) defined by a tuple $\mathcal{M} = (\S,\A,p_0,p,\gamma)$. Here, $\S$ is a (discrete) state space, $\A$ is a discrete action space, $p_0 \in \Delta(\S)$\footnote{$\Delta(\mathcal{X})$ denotes the set of probability distributions over a set $\mathcal{X}$} is the initial state distribution, $p(\cdot,\cdot|s,a) \in \Delta(\S\times\R)$ is the probabilistic transition and reward function, and $\gamma \in [0,1)$ is the discount factor. We usually abbreviate the reward by $r(s,a) = \IE_{s',r\sim p(s',r|s,a)}[r]$. %and may often assume that this is determinstic, i.e. p(r|s,a) is deterministic

The behaviour of an agent interacting with this MDP is defined by a so-called policy, that is, a mapping $\pi: \S \to \Delta(\A)$ assigning a distribution over actions to each state. Each policy induces a distribution $p_\pi(\tau)$ on the set of trajectories $\tau = (s_0,a_0,s_1,\dots)$ \citep{agarwalRein}. The (unregularized) value function is defined as $V^\pi(s) = \IE_{\tau \sim p_\pi(\tau)}[\sum_{t=0}^\infty \gamma^t \,r(s_t,a_t)|s_0=s]$, resulting in the (unregularized \& point-wise) MDP objective $\pi^\ast(\cdot|s) = \arg\max_{\pi \in \Pi} V^\pi(s)$.
% \begin{equation}
%     \label{eq::objectiveunreg}
%     \pi^\ast(\cdot|s) = \arg\max_{\pi \in \Pi} V^\pi(s)
% \end{equation}
%where $\Pi = \bigtimes_{s\in\S}\Delta(\A)$ denotes the set of all possible policies. 

\paragraph*{Regularized MDP} A natural extension hereof can be obtained by adding a regularization term on the reward level, also referred to as MDP regularization \citep{ziebart2010modeling, leeTsal2019}. More precisely, given a convex function $h: \Delta(\A) \to \R$ % and some $\alpha>0$, 
the corresponding regularized Q-value function is defined as $Q_\alpha^\pi(s,a) := \IE_{\tau\sim p_\pi(\tau)}\left[\sum_{t=0}^\infty \gamma^t \left\{r(s_t,a_t) - \alpha\,h(\pi(\cdot|s_t))\right\} | s_0=s,a_0=a\right]$ \citep{lanPoli2023}. Similarly we define state-value function $V_\alpha^\pi(s)$. Note, that this definition of the regularized Q-value function differs slightly from how it is defined in \citet{zhan2023policy}, or in \citet{haarnojaSoft2018}. However, both versions of the regularized Q-value functions can be used interchangeably in all schemes presented in this work.
% \begin{align}
%     Q_\alpha^\pi(s,a) &:= %\\
%     \IE_{\tau\sim p_\pi(\tau)}\left[\sum_{t=0}^\infty \gamma^t \left\{r(s_t,a_t) - \alpha\,h(\pi(\cdot|s_t))\right\} | s_0=s,a_0=a\right],\\
%     V_\alpha^\pi(s) &:= %\\
%     \IE_{\tau\sim p_\pi(\tau)}\left[\sum_{t=0}^\infty \gamma^t \left\{r(s_t,a_t) - \alpha\,h(\pi(\cdot|s_t))\right\} | s_0=s\right].
% \end{align}
% Extending the definition to $\alpha=0$ would recover the unregularized value functions. 
The resulting (point-wise) regularized MDP objective can be expressed as
\begin{equation}
    \label{eq::objective}
    \pi^\ast(\cdot|s) = \arg\max_{\pi\in\Pi} V_\alpha^\pi(s).% = \arg\min_{\pi\in\Pi} -V_\tau^\pi(s).
\end{equation}
%
% A notable example is the negative Shannon-Entropy, i.e. $h(p) =  \sum_{i=1}^{|\A|} p_i \log(p_i)$, where we want to restrict the search space towards more exploratory functions \citep{ziebart2010modeling, haarnojaRein2017, haarnojaSoft2018}.
%
%
% \begin{bem}
%     Note, that the above definition of the regularized Q-value function yields the following expression
%     \begin{align}
%         Q_\alpha^\pi(s,a) &= r(s,a) + \alpha h(\pi(\cdot|s)) + \gamma \IE_{s'\sim p(\cdot|s,a)}[V_\alpha^\pi(s')],
%     \end{align}
%     which differs slightly from how the regularized Q-value function is defined in \citet{zhan2023policy} %, Eq. 2.8a 
%     or in \citet{haarnojaSoft2018}%, Eq. 2
%     . However, both versions of the regularized Q-value functions can be used interchangeably in all schemes presented in this work.
% \end{bem}

\subsection{Policy Mirror Descent}

\paragraph*{Mirror Descent} A common method to solve optimisation problems like (\ref{eq::objective}) is the so-called \textit{mirror descent} algorithm \citep{beckMirr2003, beck2017first}. Consider a general differentiable function $f: \mathcal{X}\subset\R^n\to\R$ and the optimisation problem
\begin{equation}
    x^\ast = \arg\min_{x\in\mathcal{X}} f(x).
\end{equation}
Mirror descent gives the iterative update scheme starting from some $x_0 \in \mathcal{X}$
\begin{equation}
    \label{eq::MD}
    x_{k+1} = \arg\min_{x\in\mathcal{X}} \left\{\langle x, \nabla f(x)|_{x=x_k}\rangle + \lambda_k\, B_\omega(x,x_k)\right\}.
\end{equation}
Here $\langle \cdot,\cdot \rangle$ denotes the standard inner product and $B_\omega$ denotes the (generalized) Bregman divergence \citep{lan2011primal} for a convex potential function $\omega:\mathcal{X}\to\R$
\begin{equation}
    \label{eq::BregmanDef}
    B_\omega(x,y) = \omega(x) - \omega(y) - \langle \nabla \omega(y), x-y\rangle,
\end{equation}
where $\nabla \omega(y)$ can be any vector falling within the subdifferential. This scheme generalizes the (projected) gradient descent, which is recovered by choosing the squared Euclidean distance as the Bregman divergence, i.e. $B_\omega(x,y) = \Vert x - y \Vert_2^2$ \citep{beckMirr2003}.

\paragraph*{Mirror Descent inspired Policy updates} By applying the MD scheme (\ref{eq::MD}) to the regularized objective (\ref{eq::objective}) the \textit{Policy Mirror Descent} Update (see  \citep{lanPoli2023}, Algorithm 1) can be obtained
\begin{equation}
    \label{eq::PMD}
    \pi_{k+1}(\cdot|s)= \arg\min_{\pi\in\Pi}\left\{\langle \pi(\cdot|s),-Q_\alpha^{\pi_k}(s,\cdot)\rangle  + \alpha\,h(\pi(\cdot|s)) + \lambda_k\, B_\omega(\pi,\pi_k;s)\right\},
\end{equation}
for a sequence of step sizes $\lambda_k > 0$. For notational convenience we write $B_\omega(\pi,\pi_k;s)$ instead of $B_\omega(\pi(\cdot|s),\pi_k(\cdot|s))$. A popular choice for $h$ is the negative Shannon-Entropy $-\mathcal{H}$,  while the (reverse) Kullback-Leibler (KL) divergence is a common selection for $B_\omega$. Notably, the KL divergence is induced by the negative entropy as its potential function \citep{beckMirr2003}.

\subsection{Regularization in Policy Mirror Descent} 

The policy improvement scheme (\ref{eq::PMD}) consists of two components. The first, $\alpha h(\pi(\cdot|s))$, results from regularizing the MDP with a convex potential function $h$ (MDP regularizer). This term modifies the value function $V^\pi_\alpha$, reshaping the optimization surface by influencing how $V^\pi_\alpha$ changes with $\pi$ and the location of (local) minima. The second component, the Drift regularizer $\lambda_k B_\omega(\pi,\pi_k;s)$, impacts how this landscape is traversed, by penalising large deviations in updating the policy. 

% A shortcoming in using terms like "KL regularized RL" \cite{rudner2021pathologies} 
% When discussing regularization in RL, it is important to not only specify the functional form of the regularizer, i.e. the convex function $h$ or the Bregman Divergence $B_\omega$ respectively, but also specify where this regularization is applied, e.g. as an MDP regularizer or as Drift term. This can indeed be vital, as there might be overlap in valid choices for $h$ and $B_\omega$: A common Bregman-Divergence is the KL-Divergence, associated with the negative Shannon-Entropy as potential function \citep{beckMirr2003}. Moreover, for a fixed distribution $\mu \in \Delta(\A)$, the KL-Divergence $D_\mathrm{KL}(\pi(\cdot|s)||\mu)$ defines a convex function, which could be employed as MDP regularizer, for instance in imitation learning \citep{rudner2021pathologies, tiapkinDemo2024a}. Inserting these regularizers into the above policy optimization scheme would yield
% \begin{equation}
%     \pi_{k+1}(\cdot|s) = \arg\min_{\pi \in \Pi} \left\{  \langle \pi(\cdot|s), -Q_\alpha^{\pi_k}(s,\cdot)\rangle \right. \left. +\alpha\, D_\mathrm{KL}(\pi(\cdot|s)||\mu(\cdot|s)) + \lambda_k \,D_\mathrm{KL}(\pi(\cdot|s)||\pi_k(\cdot|s))\right\}.
% \end{equation}

\paragraph*{Interplay of the regularization terms: Convergence Results} Intuitively, one might expect that reshaping the optimisation landscape with an MDP regularizer $h$ would naturally impact the valid choices of the Drift regularizer, or in other words, that these types of regularization have an effect on each other. In fact, in the absence of an MDP regularizer (i.e. $\alpha = 0$%, or equivalently $h \equiv 0$
), \citet{kubaMirr} have shown that there exists a broad class of valid Drift regularizers, naturally encompassing the set of Bregman-Divergences. More specifically it was shown that for any Drift functional $D_{\tilde{\pi}}(\pi|s)$ (a distance function with only minimal requirements) the scheme
\begin{equation}
    \label{eq::MirrorLearningUpdate}
        \pi_{k+1}(\cdot|s) = \arg\min_{\pi \in \Pi} \left\{  \langle \pi(\cdot|s), -Q^{\pi_k}(s,\cdot)\rangle \right. \left. + \lambda_k\,\mathcal{D}_{\pi_k}(\pi|s)\right\}
\end{equation}
provably converges to the optimal policy $\pi^\ast$ with monotonic improvements of the return. Notably, this scheme encompasses well-known RL algorithms such as PPO \citep{schulman2017proximal}, and Mirror Descent Policy Optimisation (MDPO; \citealt{tomarMirr2021}), highlighting that many RL algorithms naturally emerge from the mirror descent perspective. %For a comprehensive list, see \citep{kubaMirr}.

In contrast, when an MDP regularizer is present (i.e., $\alpha \neq 0$), similar results could only be shown to hold for the class of Bregman Divergences \citep{lanPoli2023}, hence restricting the choice of valid Drift regularizers. Moreover, convergence rates improve from sublinear to linear when the potential function for the Bregman divergence is chosen to be the MDP regularizer, i.e., when $B_\omega = B_h$ \citep{lanPoli2023, zhan2023policy}.

\section{Methodology}
\label{sec::pract.alg}

Building on the guaranteed monotonic improvements from policy updates in (\ref{eq::PMD}) and (\ref{eq::MirrorLearningUpdate}), these updates can be incorporated into a generalized policy iteration (GPI)-like algorithm that alternates between policy evaluation and improvement \citep{sutton2018reinforcement, geistTheo2019, vieillardLeve2021}. While theoretical results assume access to exact value functions, practical actor-critic implementations typically rely on neural networks to approximate $Q^{\pi_k} \approx Q^{\phi_k}$, e.g. by minimizing a Bellman residual loss \citep{haarnojaSoft2018}. Similarly, policies are parameterized as neural networks, $\pi_k \approx \pi_{\theta_k}$, rendering point-wise updates as in (\ref{eq::PMD}) infeasible. Instead, we optimize an expectation over a state distribution $\mathcal{D} \in \Delta(\S)$, leading to a single policy improvement objective:
\begin{equation}
    \label{eq::RegRL_exp}
    % \pi_{k+1} = \arg\min_{\pi \in \Pi} \IE_{s\sim \mathcal{D}}[ \IE_{a\sim\pi(\cdot|s)}[-Q_{\alpha_k}^{\pi_k}(s,a)] + \alpha_k\, h((\pi(\cdot|s)) + \lambda_k\, D(\pi(\cdot|s);\pi_k(\cdot|s)) ].
    \theta_{k+1} = \arg\min_{\theta \in \Theta}\, \IE_{s\sim \mathcal{D}}\left[ \IE_{a\sim\pi_\theta(\cdot|s)}[-Q_{\textcolor{royalpurple}{\alpha_k}}^{\phi_k}(s,a)] + \textcolor{royalpurple}{\alpha_k}\, \textcolor{deepred}{h((\pi_\theta(\cdot|s))} + \textcolor{royalpurple}{\lambda_k}\, \textcolor{forestgreen}{D(\pi_\theta;\pi_{\theta_k}|s)} \right]
    %\textcolor{forestgreen}{D(\pi_\theta(\cdot|s);\pi_{\theta_k}(\cdot|s))} \right].
\end{equation}

\paragraph*{Objects of study} Due to the above assumptions, the theoretical guarantees do not translate into practice. To bridge this gap, we empirically analyze the effects of both commonly used and less conventional regularizers.
\begin{itemize}
    \item[\textcolor{deepred}{\scalebox{1.35}{$\bullet$}}] \textbf{MDP Regularizers} A common choice for this level of regularization are entropy-like functions, such as the (negative) Shannon-Entropy $-\mathcal{H}$ \citep{ziebart2010modeling, haarnojaRein2017, haarnojaSoft2018, haarnojaSoft2019}, or, as a generalization hereof, the (negative) Tsallis Entropy $-\mathcal{H}_m$ \citep{leeTsal2019, chow2018path}. Additionally, as less common choices $h=||\cdot||_2^2$ and the non-smooth $\max$ function will be tested. We refer to Appendix \ref{sec::apppendix-h&B} for details.
    \item[\textcolor{forestgreen}{\scalebox{1.35}{$\bullet$}}] \textbf{Drift Regularizers} A common choice as a Drift regularizer is the (reverse) KL-Divergence $D_{\mathrm{KL}}(\pi;\pi_k|s)$, a core component in MDPO \citep{geistTheo2019, tomarMirr2021}, and Uniform TRPO \citep{shani2020adaptive}, but was also studied in \citep{liu2019neural, vieillardLeve2021}. %, and moreover, arises from the natural gradient algorithm \citep{amari1998natural, kakade2001natural}. 
    Moreover, the Bregman-Divergences corresponding to the Tsallis-Entropy, $\max$ function and squared norm, respectively, will be studied in this work. We refer to Appendix \ref{sec::apppendix-h&B} for details.    
    \item[\textcolor{royalpurple}{\scalebox{1.35}{$\bullet$}}] \textbf{Temperature Parameters} In addition to the choice of regularizers, their weighting coefficients, $\alpha_k$ and $\lambda_k$, referred to as temperature parameters \citep{haarnojaSoft2018}, significantly influence algorithm performance. In many RL tasks, MDP regularization acts as an auxiliary reward to improve exploration and stability. This makes choosing a fixed $\alpha_k = \alpha$ challenging, as effective regularization should promote exploration early in training and decrease over time, motivating a linear annealing schedule.
    To simplify the selection of $\alpha_k$, \citet{haarnojaSoft2019} proposed an adaptive adjustment strategy that maintains a desired level of exploration. Specifically, $\alpha$ is updated via gradient descent on the objective $J(\alpha) = \IE_{s \sim \mathcal{D}} [\IE_{a \sim \pi_k}[-\alpha \log(\pi_k(a|s)) - \alpha \Bar{\mathcal{H}}]], $
    % \begin{equation} 
    %     J(\alpha) = \IE_{s \sim \mathcal{D}} [\IE_{a \sim \pi_k}[-\alpha \log(\pi_k(a|s)) - \alpha \Bar{\mathcal{H}}]], 
    % \end{equation}
    where $\Bar{\mathcal{H}}$ represents the expected minimum entropy of the policy. This approach extends naturally to general MDP regularizers $h$, leading to
    \begin{equation} 
        \label{alphaLossh} 
        J(\alpha) = \IE_{s \sim \mathcal{D}} [-\alpha\, h(\pi_k(\cdot|s)) + \alpha\, \Bar{h}], 
    \end{equation}
    where $\Bar{h}$ can be linearly annealed to reduce the influence of MDP regularization over training.
    Regarding $\lambda_k$, theoretical results \citep{lanPoli2023, zhan2023policy} suggest keeping it constant, i.e., $\lambda_k = \lambda$. However, empirical findings by \citet{tomarMirr2021} indicate that linearly annealing $\lambda$ over training may be more effective. %page 4. right bottom
    % Also in Schulman's initial work he proposed an adaptive scheme for fixing these parameters
\end{itemize}

\paragraph*{Algorithm} Setting $h=-\mathcal{H}$ and $D(p,q) = D_\mathrm{KL}(p,q)$ recovers the (soft) MDPO algorithm \citep{tomarMirr2021}. As noted in their work, MDPO can be implemented in either an off-policy or on-policy manner%, depending on design choices
. In this study, we adopt an off-policy implementation and refer to the resulting algorithm as MDPO($h$, $D$), emphasizing the choice of regularization%. Notably, in our implementation, the minimum in Eq. \ref{eq::RegRL_exp} is approximated by a single gradient step
; see Appendix \ref{app::sec::Alg} for details.

\section{Experiments}
\label{sec:Experiments}

% In the following sections, we empirically evaluate the impact of design choices, including the selection of $h$ and $D$, temperature levels, and their adaptation during training.

% \paragraph*{Environment Suite} 
Regularization is a key component in many deep RL algorithms but is rarely the sole factor. The goal of this work is to study the core effect of RL specific regularization. To hence keep the influence of other regularization techniques \citep{engstrom2020implementation, andrychowiWhat2020} as minimal as possible we focus on small environments with a finite action space.  This choice also enables more training seeds per experiment, improving statistical reliability \citep{agarwalDeep2021}. Experiments were conducted on the gymnax implementations \citep{gymnax2022github} of Cartpole, Acrobot \citep{DBLP:journals/corr/BrockmanCPSSTZ16}, Catch, and DeepSea \citep{osband2019behaviour}, representing a diverse set of tasks.

% \paragraph*{Performance Metric} 
To evaluate a fixed algorithm $	A$, i.e. an instance of MDPO($h,D$) with specified temperatures and regularizers, on a metric $d$, we compute the mean across environments ($e=1,\dots,E$), training seeds ($n=1,\dots,N$), and post-training evaluations ($m=1,\dots,M$) \citep{agarwalDeep2021}
\begin{equation}
    \label{eq::metric}
    d(A) = \frac{1}{E\,N\,M} \sum_{e=1}^E  \sum_{n=1}^N  \sum_{m=1}^M d_{e,n,m}(A),
\end{equation}
where $d_{e,n,m}(A)$ is the metric value for $A$ after training on environment $e$ with seed $n$. The primary metric is the normalized return after training; although no reward normalization was applied during training, the obtained returns were linearly rescaled to the unit interval $[0,1]$ to ensure comparability across environments. Further details are provided in Appendix~\ref{app::sec::ExpDetails}.

\subsection{Interplay of MDP \& Drift Regularizers}

\begin{figure}[t]
    \centering
    \begin{subfigure}[b]{0.32\textwidth}
        \centering
        \includegraphics[width=\textwidth]{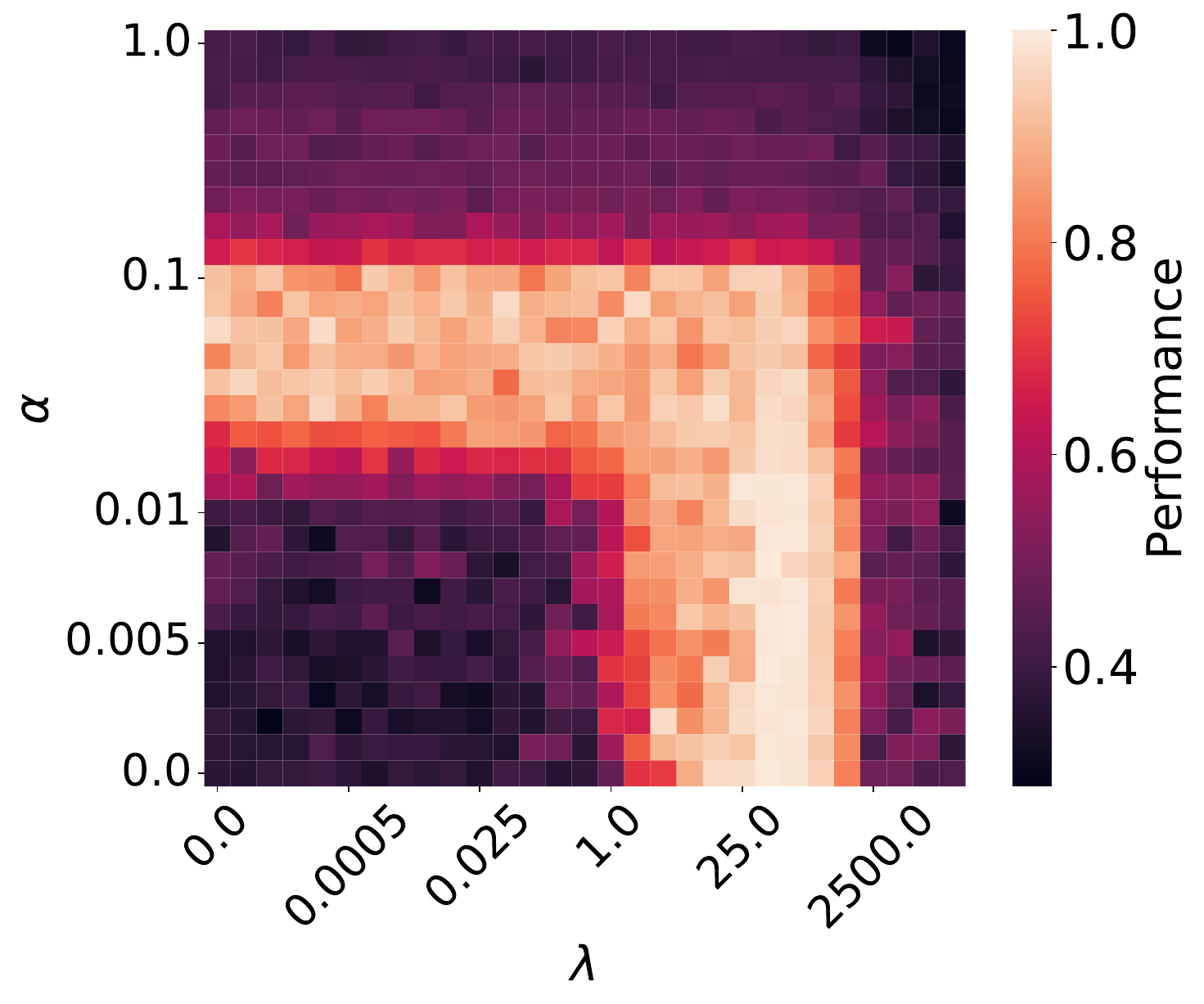}
        \caption{On standard environment suite}
        \label{fig:1a}
    \end{subfigure}
    \hfill
    \begin{subfigure}[b]{0.32\textwidth}
        \centering
        \includegraphics[width=\textwidth]{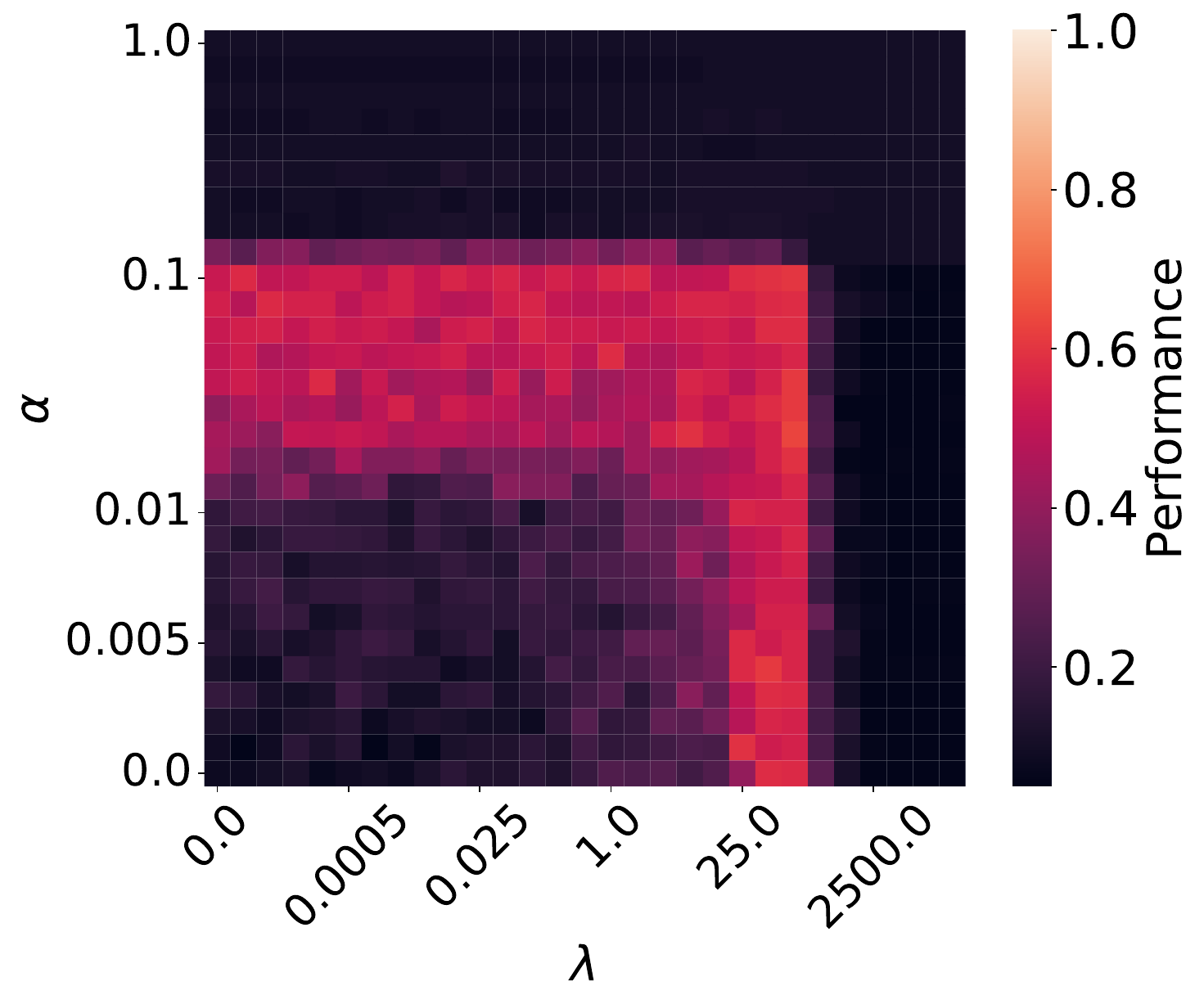}
        \caption{On upscaled versions of Catch}
        \label{fig:1b}
    \end{subfigure}
    \hfill
    \begin{subfigure}[b]{0.32\textwidth}
        \centering
        \includegraphics[width=\textwidth]{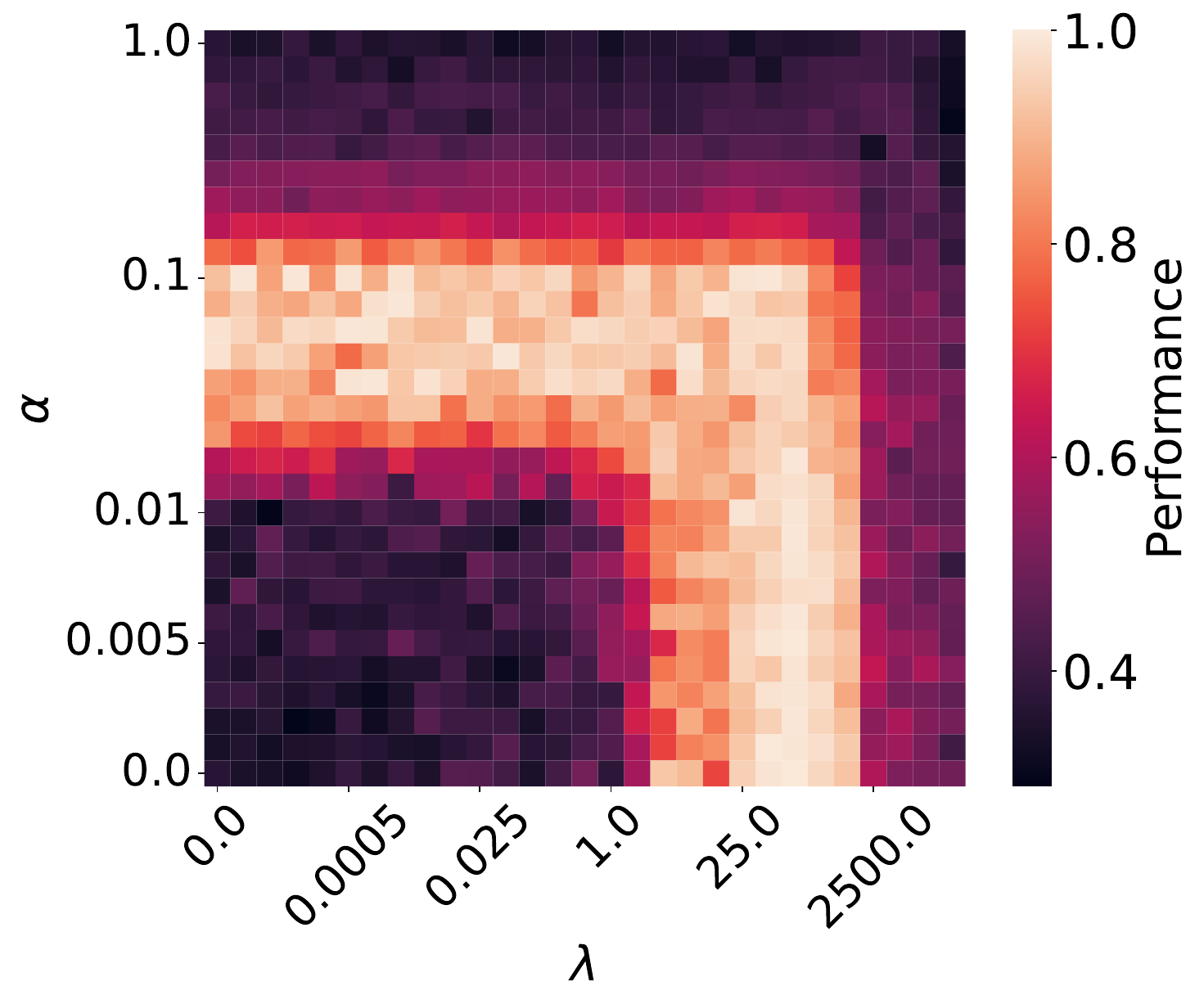}
        \caption{With APMD policy updates}
        \label{fig:1c}
    \end{subfigure}
    \caption{Mean normalized return after training of MDPO($-\mathcal{H},D_\mathrm{KL}$) for different temperature levels. Each cell represents the normalized return averaged over environments, seeds and evaluations as described in (\ref{eq::metric}). All heat maps exhibit the same principal "L" structure, highlighting regions where performance mainly depends on either of the regularization terms.}
    \label{fig:01}
\end{figure}

\paragraph*{Baseline} As a baseline we select the combination of the two well-studied regularizers $h=-\mathcal{H}$ and $D=D_\mathrm{KL}$, using constant temperature parameters during learning and across all environments. We performed minor hyperparameter tuning to ensure learnability with fixed temperature values.

Figure \ref{fig:1a} shows results for varying values of $\alpha$ and $\lambda$. Despite differences in reward structures, a joint region of well-performing temperature values emerges across environments, highlighting an overlap in optimal temperature values. Moreover, the "L" structure of this region indicates that Entropy and KL regularization may be substitutable: for sufficiently low $\lambda$, performance depends primarily on $\alpha$, and vice versa; a pattern that persists when scaling up the Catch environment (Figure \ref{fig:1b}). However, some degree of regularization remains necessary in all cases to successfully solve the given tasks.

Since the goal is to optimize the unregularized value function $V^\pi$, MDP regularization may be treated as a means to an end. This raises the question of whether policy updates should use the regularized value function (\ref{eq::RegRL_exp}). Homotopic PMD \citep{li2023homotopic}, a special case of Approximate PMD (APMD; \citealt{lanPoli2023}), provides theoretical justification for replacing the regularized Q-value function $Q^\pi_\alpha$ with the unregularized one $Q^\pi$. Figure \ref{fig:1c} shows that, at least for the environments studied, this substitution has no noticeable effect.

\paragraph*{Varying MDP regularizer and Drift} How does the performance landscape evolve with different MDP regularizers $h$? Furthermore, is it beneficial to pair the drift term $D$ with its corresponding Bregman divergence $D = B_h$, as suggested in \cite{zhan2023policy}? To explore these questions, we replicated the previous experiments using a range of choices for $h$ and $D$.

\begin{figure}[t]
    \centering
    \begin{minipage}{0.45\textwidth}
        \centering
        \includegraphics[width=\linewidth]{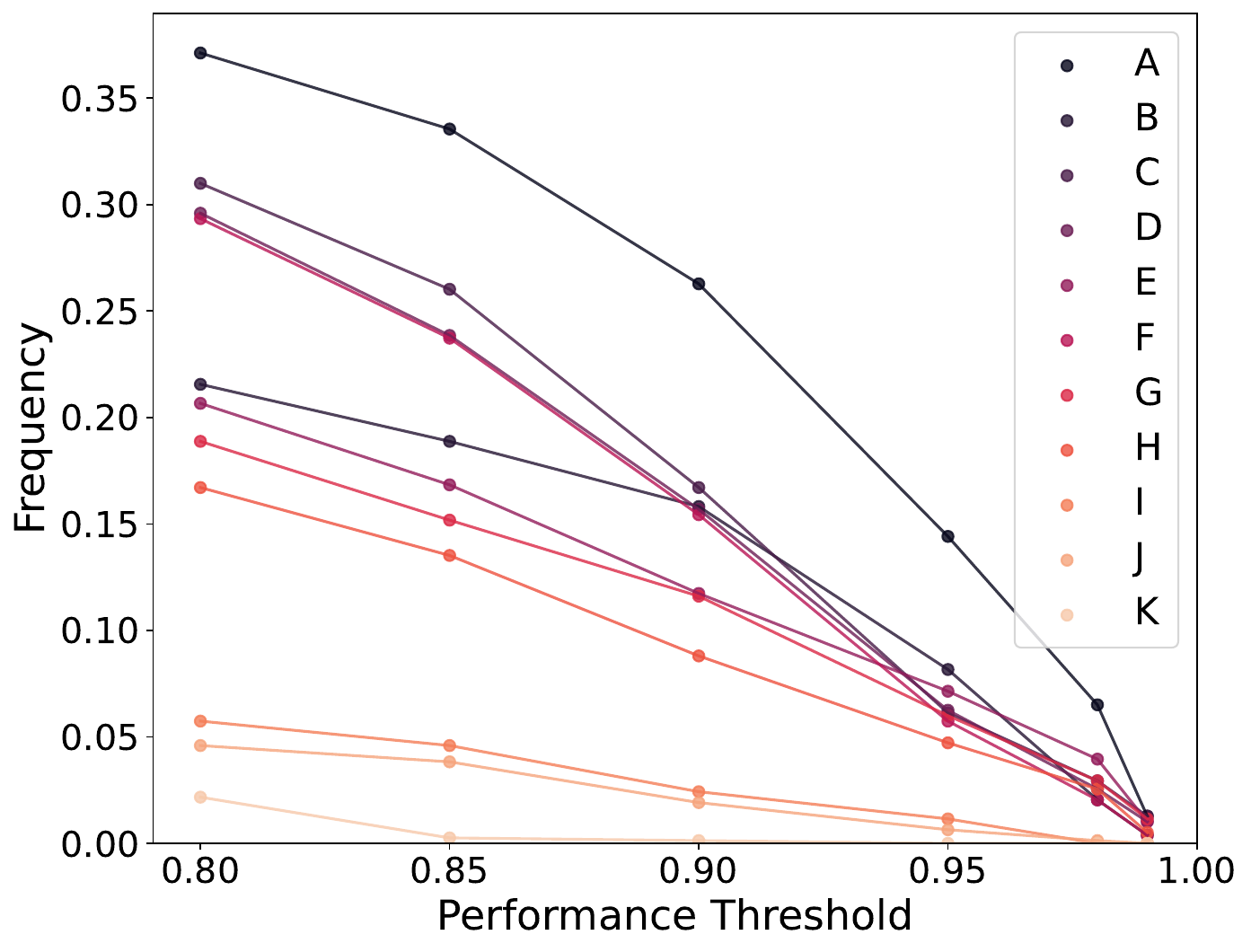} 
        \caption{Performance frequency curves for MDPO($h,D$) instances, showing the proportion of $784$ temperature configurations reaching each performance level. While high-performance regions ($\geq 0.98$) narrow similarly, regularizers $h$ and $D$ significantly affect the frequency of achieving $\geq$ 90\% performance. Labels correspond to table \ref{tab:01}.}
        \label{fig:02}
    \end{minipage}
    \hfill
    \begin{minipage}{0.5\textwidth}
        \centering
        \captionof{table}{Robustness Measure of MDPO($h,D$) for different pairs ($h,D$) in descending order, representing the (normalized) value under the curves in Figure \ref{fig:02}. Robustness appears to depend on the pair of regularizers rather than on the influence of one alone.}
        \begin{tabular}{c|c|c|c|c}
            \# & $h$ & $D$ & $\mathrm{Rbst}_{0.9}$ & $\mathrm{Rbst}_{0.95}$ \\ 
            \hline
            A & $-\mathcal{H}$ & $B_\mathrm{max}$ & 0.1403 & 0.0728 \\
            B & $\max$ & $B_\mathrm{max}$ & 0.0801 & 0.0339 \\
            C & $-\mathcal{H}$ & $D_\mathrm{KL}$  & 0.0747 & 0.0327 \\
            D & $-\mathcal{H}$ & $B_{-\mathcal{H}_{1.5}}$ & 0.0698 & 0.0313 \\    
            E & $||\cdot||_2^2$ & $D_\mathrm{KL}$  & 0.0680 & 0.0390 \\  
            F & $-\mathcal{H}$ & $B_{-\mathcal{H}_{0.5}}$ & 0.0628 & 0.0261 \\  
            G & $||\cdot||_2^2$ & $B_{||\cdot||_2^2}$ & 0.0598 & 0.0310 \\  
            H & $\max$ & $D_\mathrm{KL}$  & 0.0457 & 0.0231 \\
            I & $-\mathcal{H}_{0.5}$ & $B_{-\mathcal{H}_{0.5}}$ & 0.0101 & 0.0032 \\  
            J & $-\mathcal{H}$ & $B_{||\cdot||_2^2}$ & 0.0078 & 0.0027 \\   
            K & $-\mathcal{H}_{0.5}$ & $D_\mathrm{KL}$  & 0.0001 & 0.0000 \\ 
        \end{tabular}
        \label{tab:01}
    \end{minipage}
\end{figure}

% \begin{tabular}{c|c|c|c}
%     $h$ & $D$ & $\mathrm{Rbst}_{0.9}$ & $\mathrm{Rbst}_{0.95}$ \\ 
%     \hline
%     $-\mathcal{H}$ & $B_\mathrm{max}$ & 0.1403 & 0.0728 \\
%     $\max$ & $B_\mathrm{max}$ & 0.0801 & 0.0339 \\
%     $-\mathcal{H}$& $D_\mathrm{KL}$  & 0.0747 & 0.0327 \\
%     $-\mathcal{H}$  & $B_{-\mathcal{H}_{1.5}}$ & 0.0698 & 0.0313 \\    
%     $||\cdot||_2^2$ & $D_\mathrm{KL}$  & 0.0680 & 0.0390 \\  
%     $-\mathcal{H}$ & $B_{-\mathcal{H}_{0.5}}$ & 0.0628 & 0.0261 \\  
%     $||\cdot||_2^2$ & $B_{||\cdot||_2^2}$ & 0.0598 & 0.0310 \\  
%     $\max$ & $D_\mathrm{KL}$  & 0.0457 & 0.0231 \\
%     $-\mathcal{H}_{0.5}$ & $B_{-\mathcal{H}_{0.5}}$ & 0.0101 & 0.0032 \\  
%     $-\mathcal{H}$& $B_{||\cdot||_2^2}$ & 0.0078 & 0.0027 \\   
%     $-\mathcal{H}_{0.5}$& $D_\mathrm{KL}$  & 0.0001 & 0.0000 \\ 
% \end{tabular}

The results are summarized in Figure \ref{fig:02}, showing the proportion of tested temperature choices, that lead to a performance above a certain threshold, denoted as performance frequency. Notably, for any combination, we could still find temperature values for $\alpha$ and $\lambda$ yielding an average performance $\geq 0.95$, confirming the validity of all these theoretically grounded choices. 

While the heat maps (obtained similar to the baseline; see supplementary material) exhibit similar structures to the baseline case, with distinct $\lambda$- and $\alpha$-dominant regions, the size of well-performing regions varies, indicating an effect on the robustness w.r.t. temperature selection. To quantify this intuition of robustness, we employ a simple metric $\mathrm{Rbst}_\mathcal{T}(\mathcal{A};\Phi)$, that estimates the probability of a randomly chosen hyperparameter configuration $\phi \in \Phi$ achieving a performance of at least $\mathcal{T}$ across all evaluated environments. Further details can be found in Appendix \ref{sec:appendixRobustness}. 

Table \ref{tab:01} presents robustness measures for different $(h, D)$ pairs in descending order; the definitions of $h$ and $D$ are provided in Appendix \ref{sec::apppendix-h&B}. Interestingly, the Bregman Divergence derived from the $\max$ function seems to exhibit the highest robustness, despite being neither smooth nor commonly used. However, robustness appears to depend more on the combination of regularizers than on any single component. For instance, pairing entropy with different Drift regularizers results in both very high and very low robustness values, respectively.

\subsection{Temperature handling during Training}

We now extend our analysis beyond fixed temperature parameters and test whether the adaptive strategies introduced in Section \ref{sec::pract.alg} lead to improved performance. For the temperature $\lambda$ of the Drift regularizer we additionally test linear annealing, paired with either linear annealing or a constant MDP regularization temperature $\alpha$. Additionally, we allow $\alpha$ to be learned via the loss in (\ref{alphaLossh}), using either a linearly decaying $\Bar{h}$ ("learned lin. anneal") or a constant $\Bar{h}$ ("learned constant"). For linear annealing, the initial temperature was gradually reduced to zero, while for both versions of learned $\alpha$, the target $\Bar{h}$ was varied.

\begin{figure}[t]
    \centering
    \includegraphics[width=0.99\linewidth]{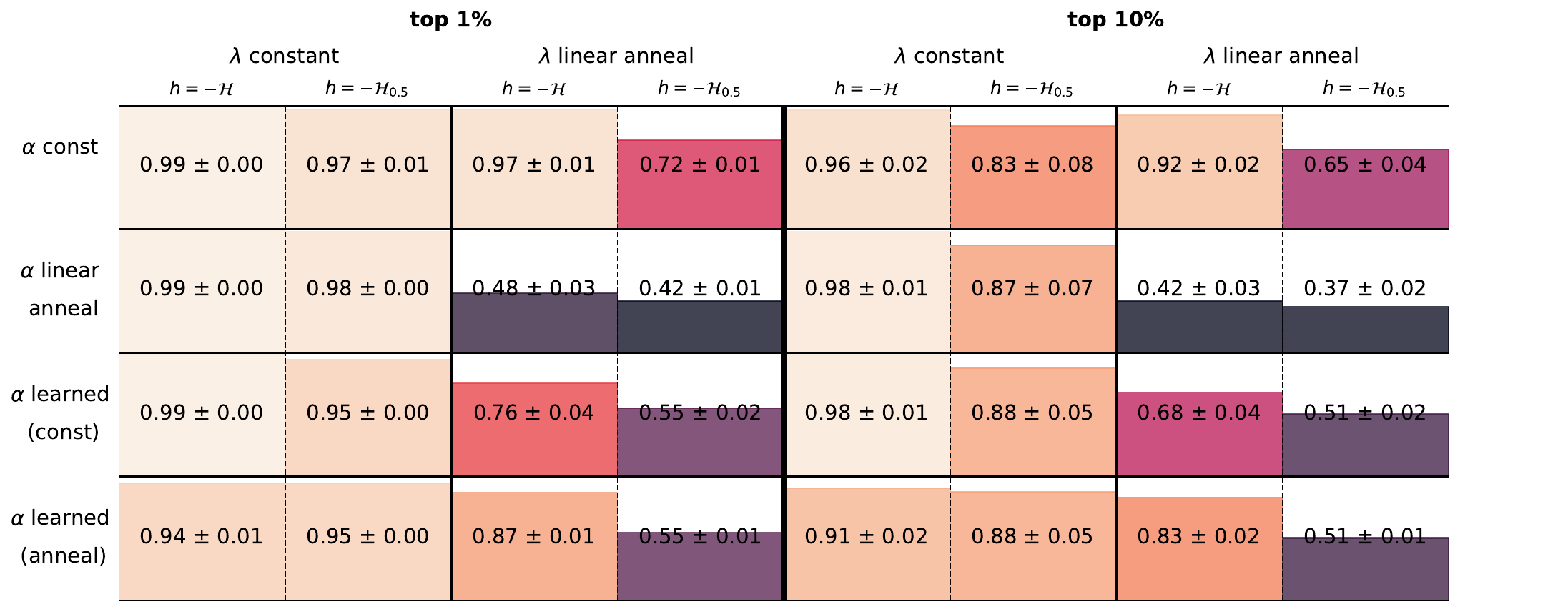}
    \caption{Mean \& standard deviations of Top 1\% and Top 10\% performing hyperparameter configurations (out of 841) for MDPO($-\mathcal{H},D_\mathrm{KL}$) and MDPO($-\mathcal{H}_{0.5},D_{-\mathcal{H}_{0.5}}$) for different temperature adaption schemes. For both algorithms keeping $\lambda$ constant outperforms the linear annealing variant.}
    \label{fig:03}
\end{figure}

To evaluate these strategies, we repeated these experiments for two pairs of $h$ and $D$. The right part of Figure \ref{fig:03} shows the average performance of the top 1\% of temperature configurations (out of 784) on regions where both regularizations apply ($\alpha, \lambda >0$). The left and right of each pair of columns shows the results for MDPO($-\mathcal{H},D_\mathrm{KL}$) and MDPO($-\mathcal{H}_{0.5},D_{-\mathcal{H}_{0.5}}$), respectively. Similarly, the right block shows the same results for the top 10\% temperature configurations. These specific percentiles were selected to reflect the typical effort researchers might invest in hyperparameter tuning.

The results support theoretical predictions that a constant $\lambda$ performs well, while linearly annealing $\lambda$ significantly degrades performance at both the 1\% and 10\% levels. This contrasts with \citet{tomarMirr2021}, where a decaying $\lambda$ was found beneficial; however, their results were obtained in the absence of MDP regularization. In contrast, $\alpha$ handling has little impact on performance, possibly due to its smaller scale, making it more robust to tuning. Notably, these trends preserve even when changing both the MDP and Drift regularizer.
% Somthing about the regions?
% Also topology of the heatmaps changes, but this is probably not tooo important to report.

\subsection{Temperature handling between different environments}

In practice, researchers also need to find an appropriate range to tune their parameters in, which can be challenging and require much intuition. However, from (\ref{eq::RegRL_exp}), we hypothesize that the preferred range of $\alpha$ is related to the absolute range of returns, rendering this choice heavily task dependent. To study this effect, we reran the experiments for MDPO($-\mathcal{H},D_\mathrm{KL}$) for $841$ different constant temperature pairs each on a set of multiplicatively rescaled versions of CartPole, spanning a maximum return range from $5$ to $1000$.

\begin{figure}[t]
    \centering
    \begin{subfigure}[b]{0.45\textwidth}
        \centering
        \includegraphics[width=\textwidth]{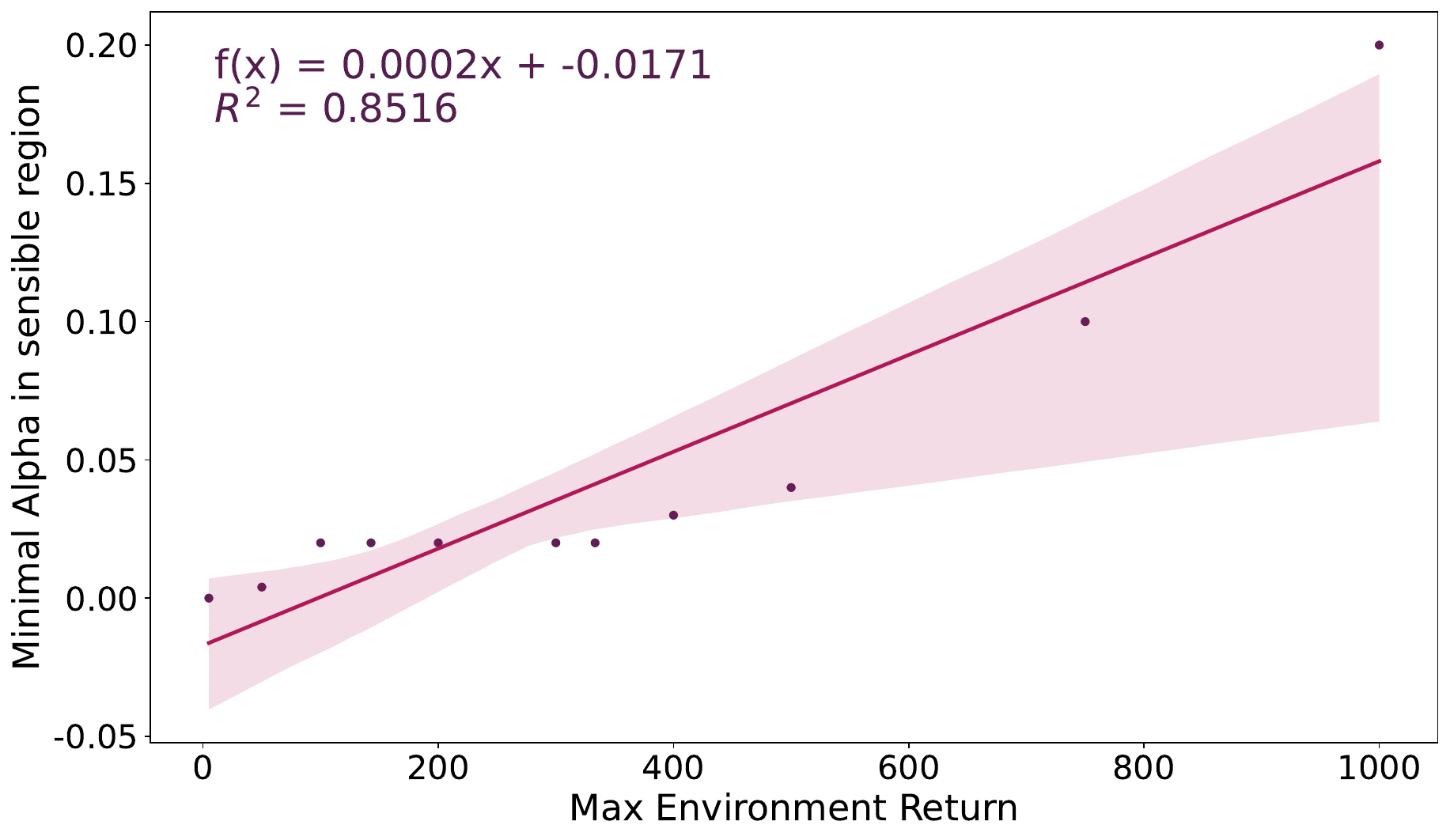}
        \caption{MDP Regularizer temperature $\alpha$}
        \label{fig:4a}
    \end{subfigure}
    \hfill
    \begin{subfigure}[b]{0.45\textwidth}
        \centering
        \includegraphics[width=\textwidth]{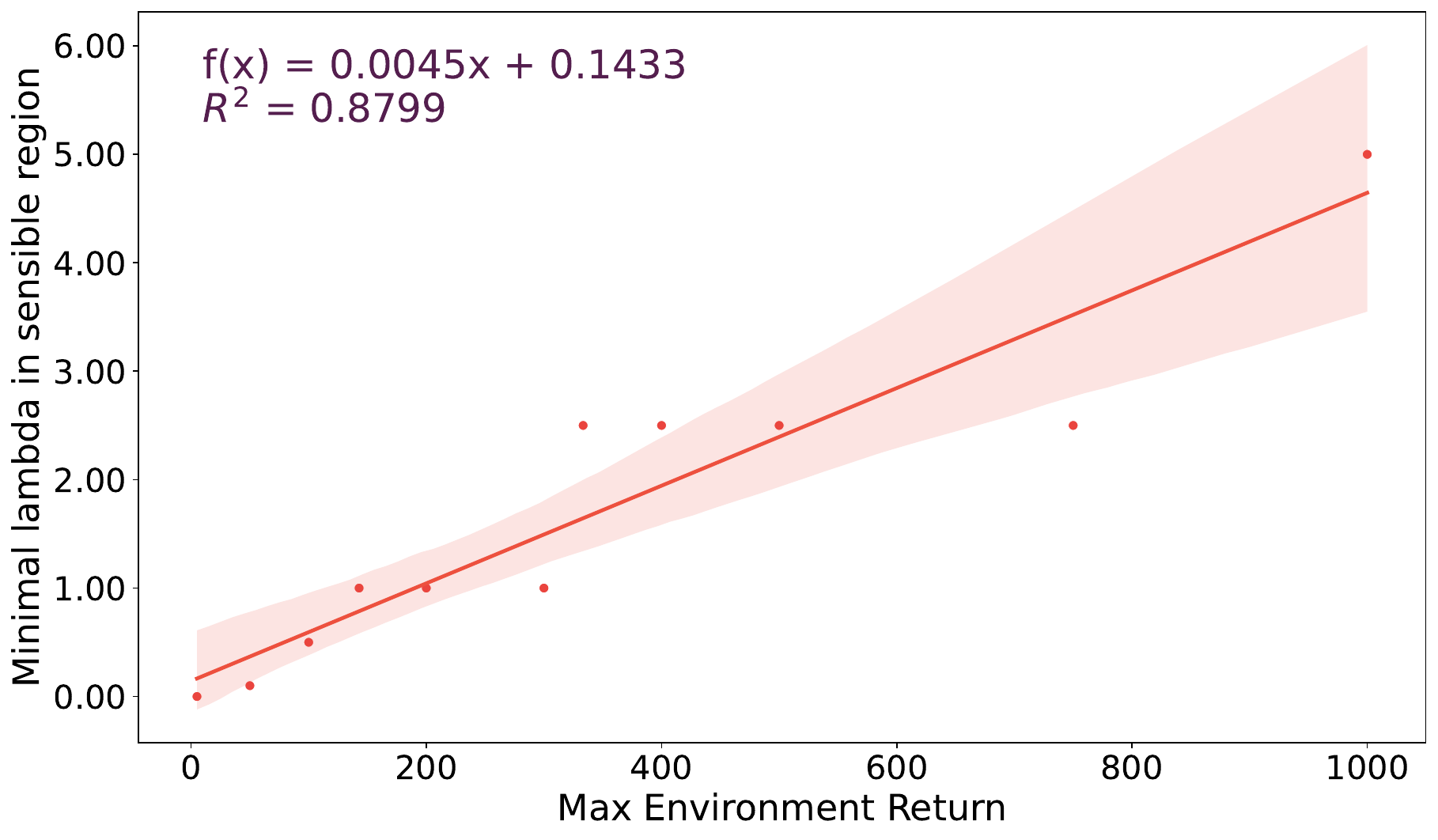}
        \caption{Drift Regularizer temperature $\lambda$}
        \label{fig:4b}
    \end{subfigure}
    \caption{Minimal required temperature for successful learning (normalized return $\geq 0.75$) as a function of maximum return in a rescaled CartPole environment. Points represent minimum temperature values; lines depict linear regressions, and shaded areas indicate 95\% confidence intervals. The minimal temperatures exhibit a monotonic trend in the maximum environment return}
    \label{fig:04}
\end{figure}

Figure \ref{fig:04} illustrates the minimum temperature required for successful learning (normalized return $\geq 0.85$) as a function of the environment’s maximum return. Extracting this value was restricted to the region, where this choice was meaningful, that is, the choice for $\alpha$ was done in regions for sufficiently low $\lambda$ and vice versa. Full heat maps are provided in the supplementary material.

These experiments indeed confirm empirically that both the temperature for the MDP regularizer $\alpha$ as well as the temperature for the Drift regularizer $\lambda$ grow linearly with the maximum return obtainable. This can help RL researchers with judging the empirical range they need to test in for setting optimal temperature values, aiding in speeding up the hyperparameter optimisation. %

\section{Related work}

% We have focused on a very special kind of Regularization for Reinforcement Learning. We try to provide an overview over those of these approaches, that we think are closely related to this work.

\paragraph*{Unified View on Reinforcement Learning Algorithms} Studying regularization is partially inspired by attempts to provide a unified view on RL algorithms. Key milestones include the introduction of MDPO in \citep{geistTheo2019}, later explored in \citep{tomarMirr2021, vieillardLeve2021}, and the Mirror Learning framework in \citep{kubaMirr}, which extends regularization beyond Mirror Descent. Recent work on Policy Mirror Descent \citep{lanPoli2023, xiaoConv, vaswaniGene2023, alfano2023novel, zhan2023policy} has further detailed the connection between model-free algorithms and mirror descent.

% Studying regularization is partially inspired by attempts to provide a unified view on RL algorithms. Notable milestones in this long line of work are \citep{geistTheo2019}, where the term \textit{Mirror Descent Policy optimization} (MDPO) was coined, which was picked up in \citep{tomarMirr2021, vieillardLeve2021}. In \citep{kubaMirr} a framework called Mirror Learning was introduced, providing a theoretically grounded class of regularizers for RL algorithms, even beyond Mirror Descent. In the line of work around Policy Mirror Descent \citep{lanPoli2023, xiaoConv, vaswaniGene2023, alfano2023novel, zhan2023policy} the connection between a big class of state-of-the-art model-free algorithms and mirror descent was laid out in detail.% and studied to obtain (optimal) convergence rates. Further studies on PMD include \citep{li2023homotopic, johnsonOpti2023, protopapasPoli2024}.

\paragraph*{Regularization in Policy Mirror Descent like Algorithms} 
% Regularization in RL
While RL algorithms can be regularized using techniques from supervised learning, e.g., weight decay, layer normalization \citep{naumanOver2024, lee2024simba, nauman2025bigger}, this work focuses on RL-specific regularization. From a Generalized Policy Iteration (GPI) perspective, regularization can be applied to three principal components: (i) the MDP itself, including entropy-based and KL-regularized RL \citep{ziebart2010modeling, haarnojaRein2017, leeTsal2019, shani2020adaptive, rudner2021pathologies, tiapkinDemo2024a}, (ii) policy evaluation (critic regularization) \citep{fujimoto2018addressing, nachumAlga2019, eysenbachConn2023, cetinLear2023a}, and (iii) policy improvement \citep{kubaMirr, lanPoli2023, xiaoConv, vaswaniGene2023, alfano2023novel, zhan2023policy}. Unlike prior work on learning optimal Drift regularizers \citep{luDisc2022, alfanolearning}, this study examines fixed regularizers for policy improvement.

%(i) The MDP itself, where part from Entropy-like regularization \cite{ziebart2010modeling, haarnojaRein2017, leeTsal2019, shani2020adaptive} a lot of work has been done studying so called KL-Regularized RL \citep{rudner2021pathologies}, for instance applied for learning from demonstrations \citep{tiapkinDemo2024a}. (ii) The policy evaluation step, also referred to as critic regularization \citep{fujimoto2018addressing, nachumAlga2019, eysenbachConn2023, cetinLear2023a}. (iii) The policy improvement step \citep{kubaMirr, lanPoli2023, xiaoConv, vaswaniGene2023, alfano2023novel, zhan2023policy}. In contrast to this work, where fixed regularizers for the latter one were studied, prior work has focused on learning optimal Drift regularizers \citep{luDisc2022, alfanolearning}.

\section{Discussion \& Conclusion}
\label{sec:conclusion}

In this work, we empirically analyzed the interplay between the two regularization components in PMD: Drift and MDP regularization. Across 500k training seeds, we systematically examined a broad range of regularizers and their temperature parameters.

To ensure comprehensive coverage of algorithm configurations, our study focused on a small set of environments with a limited number of environment steps, meaning our findings reflect performance within this training horizon. Future research could extend this analysis to longer training horizons and larger-scale environments to assess whether the observed trends persist. 

Our results indicate that, for most combinations, Drift and MDP regularization can act as substitutes, as reflected in the “L”-shaped regions of well-performing temperature values. Selecting well-suited temperature values is vital, as our results demonstrate that even in small environments, poorly chosen parameters can severely degrade performance. To address this brittleness, we examined the impact of temperature scales both during training and across environments, confirming a linear relationship between temperature and reward scale for both regularizers. Additionally, we evaluated dynamic adaptation strategies for temperature parameters and found that maintaining constant values is often more effective, particularly when moving beyond the conventional Entropy-KL regularization pair. This raises the question of whether existing adaptation strategies are overly tailored to well-established regularizers. Another approach to mitigating this brittleness is through improved robustness with respect to temperature selection. Our findings suggest that the choice of regularizers plays an important role in robustness, particularly for less commonly used ones, highlighting an important direction for future research. Overall, our study underscores the need for further investigation into regularization in RL to develop not only more performant but also more robust algorithms.

% \newpage

% \subsubsection*{Broader Impact Statement}
% \label{sec:broaderImpact}
% In this optional section, RLJ/RLC encourages authors to discuss possible repercussions of their work, notably any potential negative impact that a user of this research should be aware of. 

%%%%%%%%%%%%%%%%%%%%%%%%%%%%%%%%%%%%%%%%%%%%%%%%%%%%%%%%%%%%%%%%
%% Appendices
%%%%%%%%%%%%%%%%%%%%%%%%%%%%%%%%%%%%%%%%%%%%%%%%%%%%%%%%%%%%%%%%

%Baseline summary for 

% \label{sec:ack}
% Use unnumbered third level headings for the acknowledgments. All acknowledgments, including those to funding agencies, go at the end of the paper. Only add this information once your submission is accepted and deanonymized. The acknowledgments do not count towards the 8--12 page limit.

%%%%%%%%%%%%%%%%%%%%%%%%%%%%%%%%%%%%%%%%%%%%%%%%%%%%%%%%%%%%%%%%
%% NOTE: THIS MARKS THE END OF THE "MAIN TEXT"
%%%%%%%%%%%%%%%%%%%%%%%%%%%%%%%%%%%%%%%%%%%%%%%%%%%%%%%%%%%%%%%%

%%%%%%%%%%%%%%%%%%%%%%%%%%%%%%%%%%%%%%%%%%%%%%%%%%%%%%%%%%%%%%%%
%% Bibliography
%%%%%%%%%%%%%%%%%%%%%%%%%%%%%%%%%%%%%%%%%%%%%%%%%%%%%%%%%%%%%%%%
% \newpage
\bibliography{_arxiv}
% \bibliographystyle{rlj}

%%%%%%%%%%%%%%%%%%%%%%%%%%%%%%%%%%%%%%%%%%%%%%%%%%%%%%%%%%%%%%%%
% AUTHOR: If your paper has no supplementary materials, you may 
%         comment out the line below, which creates the title for
%         the supplementary materials.
%%%%%%%%%%%%%%%%%%%%%%%%%%%%%%%%%%%%%%%%%%%%%%%%%%%%%%%%%%%%%%%%
\newpage
\appendix
\section{Convex Functions and their Bregman Divergences}
\label{sec::apppendix-h&B}

\subsection{Entropy}

The Shannon entropy of a (finite) probability distribution $p \in \Delta(\mathcal{X})$ is defined as 
\begin{equation}
    \mathcal{H}(p) = \sum_{x\in\mathcal{X}} -p(x) \, \log(p(x)) = \IE_{x \sim p}[-\log(p(x))].
\end{equation}
The Tsallis Entropy for $m \neq 1$ is defined as
\begin{equation}
    \mathcal{H}_m(p) = \frac{1}{m-1} \sum_{x\in \mathcal{X}} p(x) - p(x)^m.
\end{equation}
For $m > 0$ $h(p) = -\mathcal{H}_m(p)$ is a convex function. Moreover, the Tsallis Entropy generalizes the Shannon Entropy in the sense, that $\lim_{m\to1}\mathcal{H}_m = \mathcal{H}$.

The Bregman Divergence for a convex potential function is defined as 
\begin{equation}
    B_h(p,q) = h(p) - h(q) - \langle \nabla h(q), p-q\rangle,
\end{equation}
where $h(q)$ is any vector within the subdifferential and $\langle\cdot,\cdot\rangle$ denotes the standard inner product on $\R^{|\mathcal{X}|}$. A straight forward calculation yields
\begin{equation}
    B_{-\mathcal{H}_m}(p,q) = \frac{1}{m-1} \sum_{x \in \mathcal{X}} p(x)^m - m p(x) q(x)^{m-1} - (1-m) q(x)^m.
\end{equation}

While $|\mathcal{H}(p)| \leq \log(|\mathcal{X}|)$, the Tsallis Entropy for $m\neq 1$ can be bounded on $\Delta(\mathcal{X})$ by 
\begin{equation}
    |\mathcal{H}_m(p)| \leq 
    \begin{cases}
        1/(m-1) & m >1 \\
        |\mathcal{X}|/(1-m) \max_{y \in [0,1]} |y-y^m| & 0 < m < 1.
    \end{cases}
\end{equation}

This can be used to provide a reasonable guess for setting $\Bar{h}$ in Eq. (\ref{alphaLossh}).

\subsection{$L_p$ Norm}

Let $\mathcal{X}$ be a finite set, $q \in \Delta(\mathcal{X})$. Then 
\begin{align}
    ||\cdot||_p^p: \Delta(\mathcal{X}) &\to \R,\\
    q &\mapsto  ||q||_p^p:= \sum_{x \in \mathcal{X}} |q(x)|^p
\end{align}
is a convex function for $p\geq 1$. % as ()^p convex and monotonic, and ||.||_p is convex
The corresponding Bregman Divergence for $q, q' \in \Delta(\mathcal{X})$ is given by
\begin{equation}
    B_{||\cdot||_p^p}(q,q') = \sum_{x\in\mathcal{X}} q(x)^p - q'(x)^p - p\,q(x)\,q'(x)^{p-1} + p \,q'(x)^p
\end{equation}

% Reason is the gradient is 
% \begin{equation}
%     \frac{\partial ||x||_p^p}{\partial x_j} = p \, x_j ^{p-1} 
% \end{equation}

\subsection{Max Function}

Let $\mathcal{X}$ be a finite set, $q \in \Delta(\mathcal{X})$. Then 
\begin{align}
    \max: \Delta(\mathcal{X}) &\to \R,\\
    q &\mapsto \max_{x \in \mathcal{X}} |q(x)| = \max_{x \in \mathcal{X}} q(x),
\end{align}
is a convex function. This function is not smooth, however, it is convex with subdifferential 
\begin{equation}
    \partial \max (q) = \left\{ \sum_{j \in J(x)} \alpha_j e_j : \sum_{j \in J(x)} \alpha_j = 1, J(x) = \arg\max_{x\in\mathcal{X}}q(x) \right\},
\end{equation}
where $e_j$ denotes the $j$-th unit vector in $\R^{|\mathcal{X}|}$. The corresponding Bregman Divergence for $q, q' \in \Delta(\mathcal{X})$ can hence be expressed as
\begin{equation}
    B_{\max}(q,q') = \max(q) - \max(q') - \langle \nabla \max(q'), q-q'\rangle,
\end{equation}
where we can canonically select $\nabla \max(q') = e_j$ for some $j \in \arg\max_{x\in\mathcal{X}}q'(x) $.

\section{Robustness measure of an algorithm} 
\label{sec:appendixRobustness}

To quantify robustness of an algorithm $\A_\phi$ (in this work an instance of MDPO($h,D$) w.r.t. a set of hyperparameters $\Phi$ (in this work the temperature levels), we define the performance frequency as the proportion of hyperparameter configurations achieving at least a given performance $\tau$:
\begin{equation} 
    \mathrm{freq}(\tau; \mathcal{A}, \Phi) = \frac{\#\{\phi \in \Phi| d(\A_\phi) \geq \tau\}}{|\Phi|}.
\end{equation} 
Intuitively, the frequency to a given performance level $\tau$ may be seen as the probability of achieving at least this performance level for a random selection of hyperparameters (within $\Phi$). In this sense, the normalized area under this curve would provide a measure of robustness: the likelihood of a random configuration achieving at least a certain performance
\begin{equation}
    \label{eq::Robstmeasure}
    \mathrm{Rbst}_\mathcal{T}(\A;\Phi) := \frac{1}{1-\mathcal{T}}\int_{\mathcal{T}}^1 \mathrm{freq}(\tau; \mathcal{A}, \Phi)\,d\tau.
\end{equation}

\section{Experiment Details}
\label{app::sec::ExpDetails}

% Unless stated otherwise, all algorithms were trained for an equal number of environment steps across four environments—CartPole, Acrobot, Catch, and DeepSea—using their standard implementations from Gymnax \citep{gymnax2022github}. In all environments except Acrobot, the maximum return is attainable and was therefore used for return normalization. For Acrobot, we considered the task solved at a return of $-75$, a threshold slightly exceeding the provided PPO baseline in Gymnax \citep{gymnax2022github}, and used this value for normalization.

% Additionally, experiments were conducted on a scaled-up version of Catch (Figure \ref{fig:1b}), where the number of rows and columns was increased by factors of two and three, respectively, compared to the default configuration. % scaled up CartPole

% For each algorithm configuration, defined by a specific set of temperature parameters for a given MDP regularizer $h$ and Drift regularizer $D$, we ran $N=5$ train seeds per environment, followed by $M=10$ evaluations per trained model. If not specified otherwise, a total of $29^2 = 841$ temperature configurations were tested for each instance of MDPO($h,D$). 

Unless stated otherwise, all algorithms were trained for an equal number of environment steps across four environments---CartPole, Acrobot, Catch, and DeepSea---using their standard implementations from Gymnax~\citep{gymnax2022github}. No reward normalization was applied during training. However, to enable averaging of performance across environments, returns were linearly rescaled to the unit interval $[0,1]$ using
\[
\mathrm{Return}_\mathrm{rescaled} = \frac{\mathrm{Return} - R_\mathrm{min}}{R_\mathrm{max} - R_\mathrm{min}},
\]
where $R_\mathrm{max}$ and $R_\mathrm{min}$ denote the maximum and minimum attainable returns for each environment, respectively. By design, these were $\{500, 0\}$ for \textbf{CartPole}, $\{1, -1\}$ for \textbf{Catch}, and $\{1, 0\}$ for \textbf{DeepSea}. For \textbf{Acrobot}, we considered the task solved at a return of $-75$, slightly outperforming the PPO baseline in Gymnax, and used $\{R_\mathrm{max}, R_\mathrm{min}\} = \{-75, -500\}$ for normalization.

% Unless stated otherwise, all algorithms were trained for an equal number of environment steps across four environments—CartPole, Acrobot, Catch, and DeepSea—using their standard implementations from Gymnax \citep{gymnax2022github}. The algorithms were trained on all environments without any reward normalization. To allow, however, for an average of performances over different environments, the returns were linearly rescaled to the unit interval $[0,1]$, via $\mathrm{Return}_\mathrm{rescaled} = \frac{\mathrm{Return} - R_\mathrm{min}}{R_\mathrm{max} - R_\mathrm{min}}$, where $R_\mathrm{max}$ and $R_\mathrm{min}$ are the maximally and minimally obtainable returns respectively in the chosen environment. By design, the return scales for Cartpole are $\{R_\mathrm{max},R_\mathrm{min}\} = \{500, 0\}$, for Catch are $\{R_\mathrm{max},R_\mathrm{min}\} = \{1, -1\}$ and for DeepSea $\{R_\mathrm{max},R_\mathrm{min}\} = \{1, 0\}$. For Acrobot, we considered the task solved at a return of $-75$, a threshold slightly exceeding the provided PPO baseline in Gymnax \citep{gymnax2022github}, and used values of $\{R_\mathrm{max},R_\mathrm{min}\} = \{-500, -75\}$ for normalization.

For each algorithm configuration, defined by a specific set of temperature parameters for a given MDP regularizer $h$ and Drift regularizer $D$, we ran $N=5$ training seeds per environment, followed by $M=10$ evaluations per trained model. In all experiments, a total of $29^2 = 841$ temperature configurations were tested for each instance of $\text{MDPO}(h, D)$. For constant temperature settings, the specified value refers directly to the temperature parameter; for linearly annealed settings, it refers to the initial value.

\newpage
For the Drift regularizer temperature $\lambda$, we tested the following values:
\begin{equation*}
\begin{aligned}
    \lambda \in \{&0.0,\ 5{\cdot}10^{-5},\ 7.5{\cdot}10^{-5},\ 10^{-4},\ 2.5{\cdot}10^{-4},\ 5{\cdot}10^{-4},\ 7.5{\cdot}10^{-4},\ 10^{-3},\ 5{\cdot}10^{-3},\ 10^{-2}, \\
    &2.5{\cdot}10^{-2},\ 5{\cdot}10^{-2},\ 10^{-1},\ 2.5{\cdot}10^{-1},\ 5{\cdot}10^{-1},\ 1,\ 2.5,\ 5,\ 7.5,\ 10, \\
    &25,\ 50,\ 10^2,\ 5{\cdot}10^2,\ 10^3,\ 2.5{\cdot}10^3,\ 5{\cdot}10^3,\ 10^4,\ 5{\cdot}10^4\}.
\end{aligned}
\end{equation*}

For the MDP regularizer temperature $\alpha$, we used:
\begin{equation*}
    \alpha \in \{0.0,\ 0.001,\ 0.002,\ 0.003,\ \ldots,\ 0.009,\ 0.01,\ 0.02,\ 0.03,\ \ldots,\ 0.1,\ 0.2,\ 0.3,\ \ldots,\ 1.0\},
\end{equation*}
where values increment in steps of $0.001$ from $0.001$ to $0.009$, then in coarser steps up to $1.0$ (29 values in total).

In the case of a learned MDP regularizer temperature $\alpha$ with a constant target, the target was defined as $\bar{h} = w \cdot \bar{h}_0$, where $\bar{h}_0$ is an environment-specific reference value and $w$ was varied over the same 29 values used for $\alpha$. For learned parameters with linearly annealed targets of the form $\bar{h}_k = w_k \cdot \bar{h}_0$, the initial weight $w_0$ was similarly varied over the same set.

On each environment, we ran $841$ temperature configurations with $5$ seeds each, totaling $4205$ runs per environment and algorithm. In addition to the baseline setup (4 environments), we evaluated 10 variations of $\text{MDPO}(h,D)$ on four environments each (40 additional environments). The baseline algorithm was also run on a scaled-up version of Catch (Figure~\ref{fig:1b}), where the number of rows and columns was increased by factors of two and three, respectively, compared to the default configuration (2 additional environments). Furthermore, an AMPD-inspired variant was evaluated on a set of 4 environments (Figure~\ref{fig:1c}; 4 additional environments). To study the influence of temperature scaling, 15 additional algorithm configurations were tested on four environments each (Figure~\ref{fig:03}; 60 additional environments). Finally, we ran the baseline algorithm on 10 rescaled versions of CartPole (10 additional environments), resulting in a total of $120 \times 4205 = 504{,}600$ training seeds.

% On each environment we ran $841$ temperature configurations, with each $5$ seeds, equalling $4205$ per environment and algorithm. Additional to the baseline (4 envs), we ran 10 variations of MDPO($h,D$) on four environments each (10 additional environments). Moreover, the baseline algorithm was run on scaled-up version of Catch (Figure \ref{fig:1b}), where the number of rows and columns was increased by factors of two and three, respectively, compared to the default configuration (2 additional environments). Additionally an AMPD insipired variant was evaluated on a set of 4 environments (Figure \ref{fig:1c}; 4 additional environments). To study the influence of temperature scaling an additional 15 algorithm configurations were tested on a set of four environments each (Figure \ref{fig:03}; 60 additional environments). Lastly, we ran the baseline algorithm on 10 additional rescaled versions of CartPole (10 additional environments), resulting in a total of $120 * 4205 = 504.600$ seeds.

\section{Algorithm Details}
\label{app::sec::Alg}

\begin{table}[htbp]
\centering
\caption{Fixed hyperparameters for all MDPO($h,D$) instances}
\begin{tabular}{l|l}
 Parameter & Value  \\ 
\hline
 Number of environments & $16$  \\
Max grad norm &  $1.0$ \\
Gamma & $0.99$  \\
 Replay buffer size &  $10^5$ \\
Environment steps per update &  $256$  \\
Train batch size & $512$  \\
Critic update epochs & $1$ \\
Actor update epochs & $2$  \\
Tau & $0.95$  \\
% Number of minibatches &  $1$ \\
Learning rates & $0.0025$  \\
Total Environment steps & $10^6$ \\
\end{tabular}
\end{table}

\begin{algorithm}[H]
\caption{Off-policy MDPO($h, D$) Algorithm}
\label{alg:naiveRL}
    Init networks $(Q_{\phi_0, i})_{i=1,2}, \pi_{\theta_0}$, data buffer $D = \emptyset$, target networks $\phi_\mathrm{target,\,i} = \phi_{0,i}$\\
    \For{budget}{ %\tcc*{i.e. total timesteps}
        \For{environment steps per update}{
            \tcc{sample data}
            sample $(s,a,r,s',d) \sim \pi_{\theta_k}, p$,\\
            $D\leftarrow D \cup \{(s,a,r,s',d)\}$\\ 
            }
        Sample train batch $\mathcal{D} \sim D$\\
         \text{ }\\
        \tcc{Policy Improvement: update Actor network, cf. (\ref{eq::RegRL_exp})}
        $L(\theta,\theta_k) = \IE_{s\sim \mathcal{D}}\left[ \IE_{a\sim\pi_\theta(\cdot|s)}[-Q_{\alpha_k}^{\phi_k}(s,a)] + \alpha_k\, h((\pi_\theta(\cdot|s)) + \lambda_k\, D(\pi_\theta;\pi_{\theta_k}|s) \right]$ \\
        \For{$i=0,\dots,\mathrm{actor\,epochs}-1$}{
            $\theta_{k}^{(i+1)} \leftarrow \theta_k^{(i)} - \eta_\theta \nabla_\theta L(\theta, \theta_k)|_{\theta=\theta_k^{(i)}}$
        }
        $\theta_{k+1} = \theta_k^\mathrm{actor\,epochs}$\\
        \text{ }\\
        \tcc{Policy Evaluation}
        $L_Q(\phi) = \IE_{(s,a,r,s')\sim\mathcal{D}}[(Q_\phi(s,a) - \{r + \gamma \,\IE_{a'\sim \pi_{\theta_{k+1}}}[\min Q_{\phi_\mathrm{target, i}}(s',a')] - \alpha_k\,h(\pi_{\theta_{k+1}}(\cdot|s')\})^2]$
        \For{$j=0,\dots,\mathrm{critic\,epochs}-1$}{
            $\phi_{k}^{(j+1)} \leftarrow \phi_k^{(j)} - \eta_\phi \nabla_\phi L_Q(\phi)|_{\phi=\phi_k^{(i)}}$
        }
        $\phi_{k+1} = \phi_k^\mathrm{critic\,epochs}$\\
        \text{ }\\
        $\phi_\mathrm{target,\,k+1} = \tau \, \phi_\mathrm{target,\,k} + (1-\tau)\, \phi_{k+1}$
    }
\end{algorithm}

\section{Additional Plots}

\subsection*{Additional Heat maps}

\begin{figure}[htbp]
    \centering
    \begin{subfigure}[b]{0.32\textwidth}
        \centering
        \includegraphics[width=\textwidth]{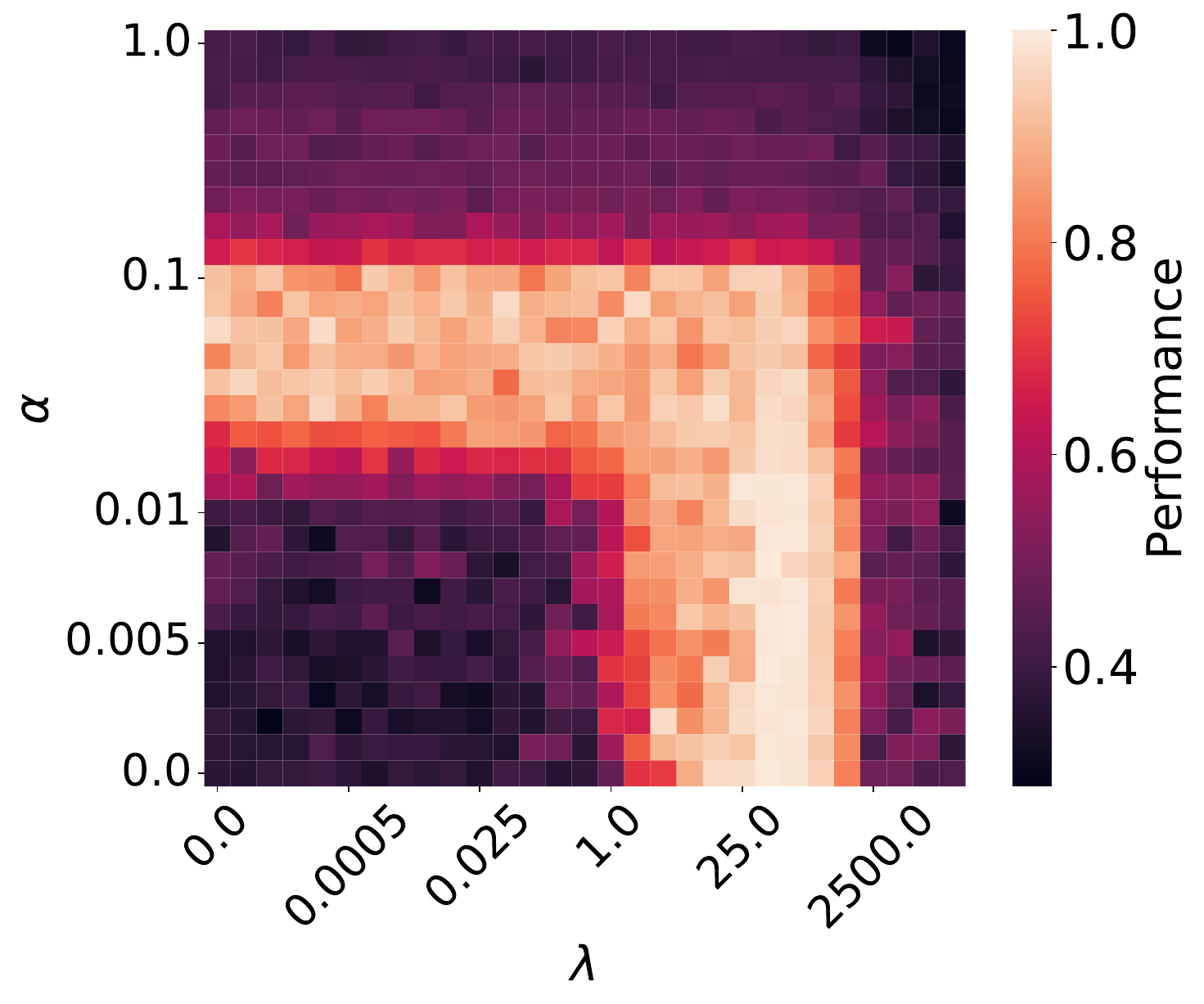}
        \caption{$h=-\mathcal{H}, D=D_\mathrm{KL}$}
    \end{subfigure}
    \hfill
    \begin{subfigure}[b]{0.32\textwidth}
        \centering
        \includegraphics[width=\textwidth]{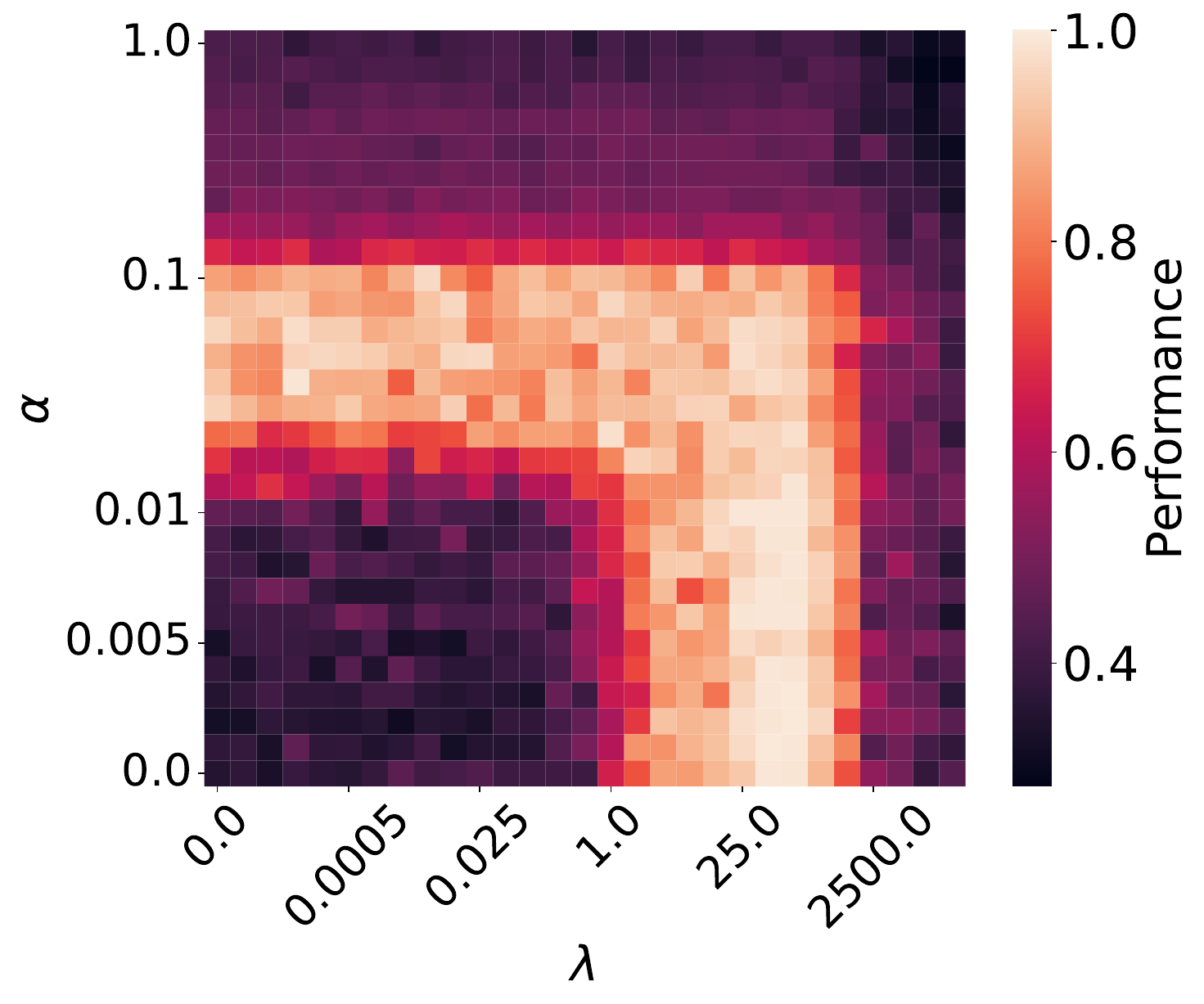}
        \caption{$h=-\mathcal{H}, D=D_\mathrm{fKL}$}
    \end{subfigure}
    \hfill
    \begin{subfigure}[b]{0.32\textwidth}
        \centering
        \includegraphics[width=\textwidth]{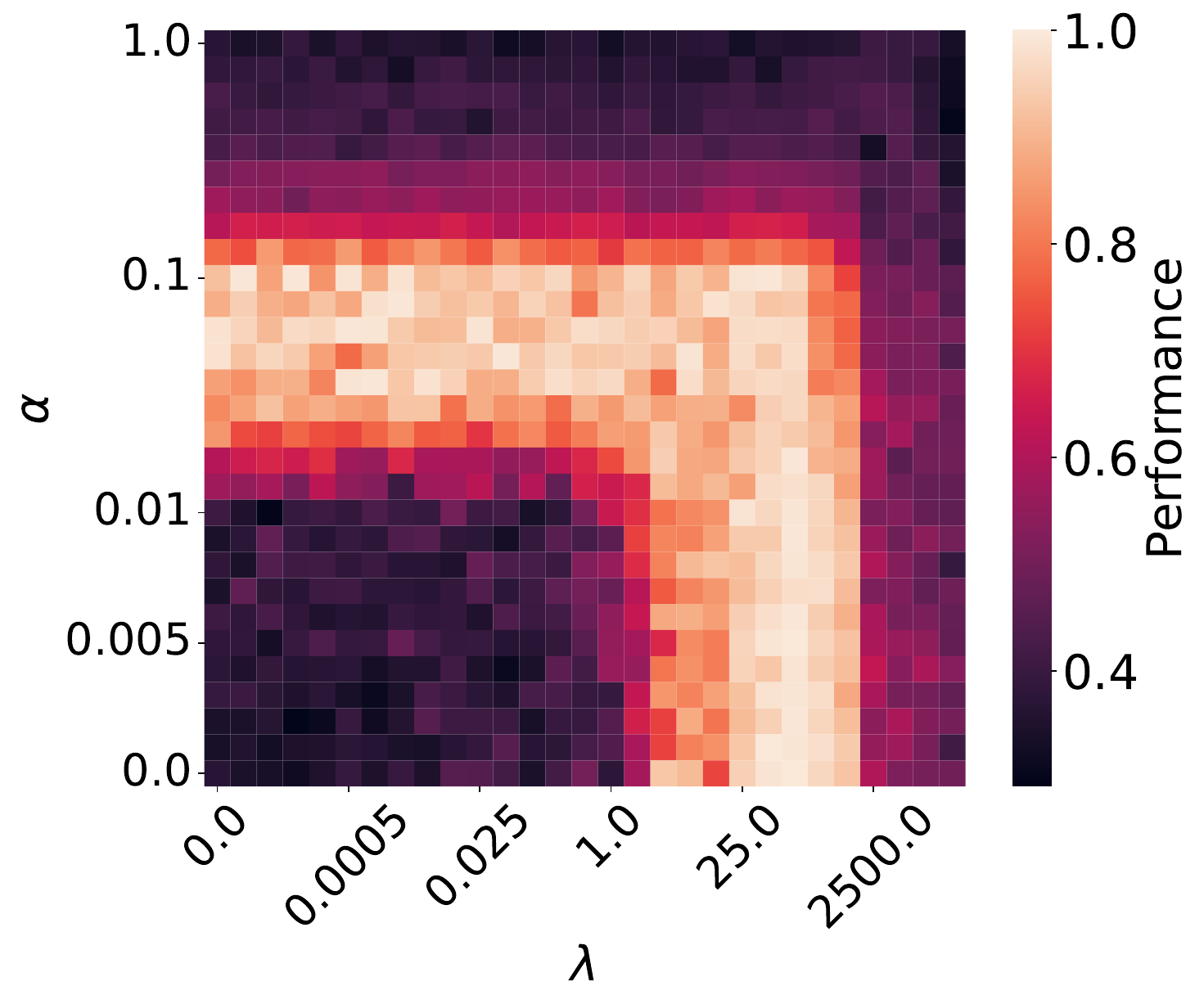}
        \caption{$h=-\mathcal{H}, D=D_\mathrm{KL}$; AMPD}
    \end{subfigure}
    \caption{Different MDPO($h,D$) configurations with constant temperatures}
    \label{fig:App01}
\end{figure}

\begin{figure}[htbp]
    \centering
    \begin{subfigure}[b]{0.24\textwidth}
        \centering
        \includegraphics[width=\textwidth]{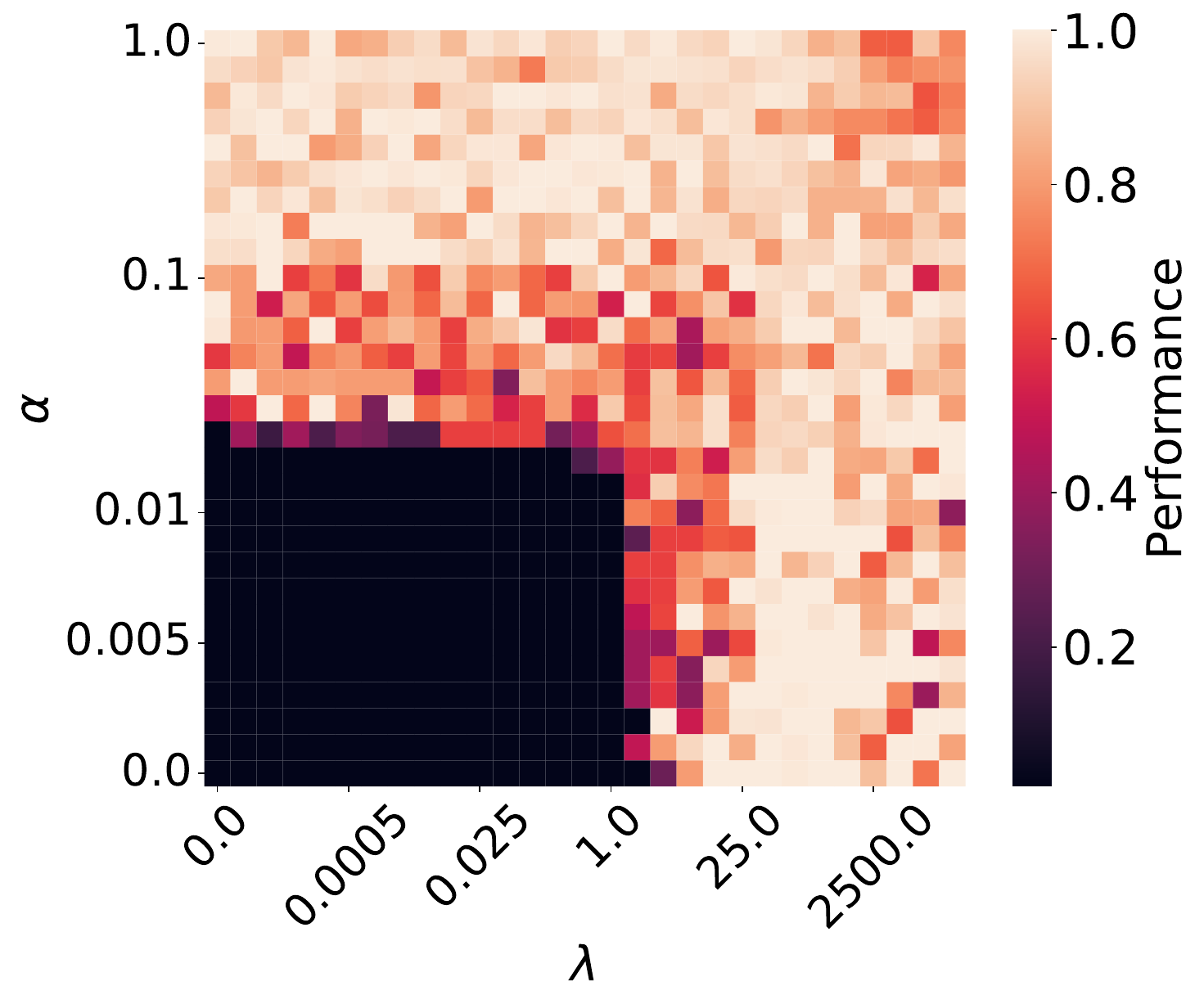}
        \caption{CartPole}
    \end{subfigure}
    \hfill
    \begin{subfigure}[b]{0.24\textwidth}
        \centering
        \includegraphics[width=\textwidth]{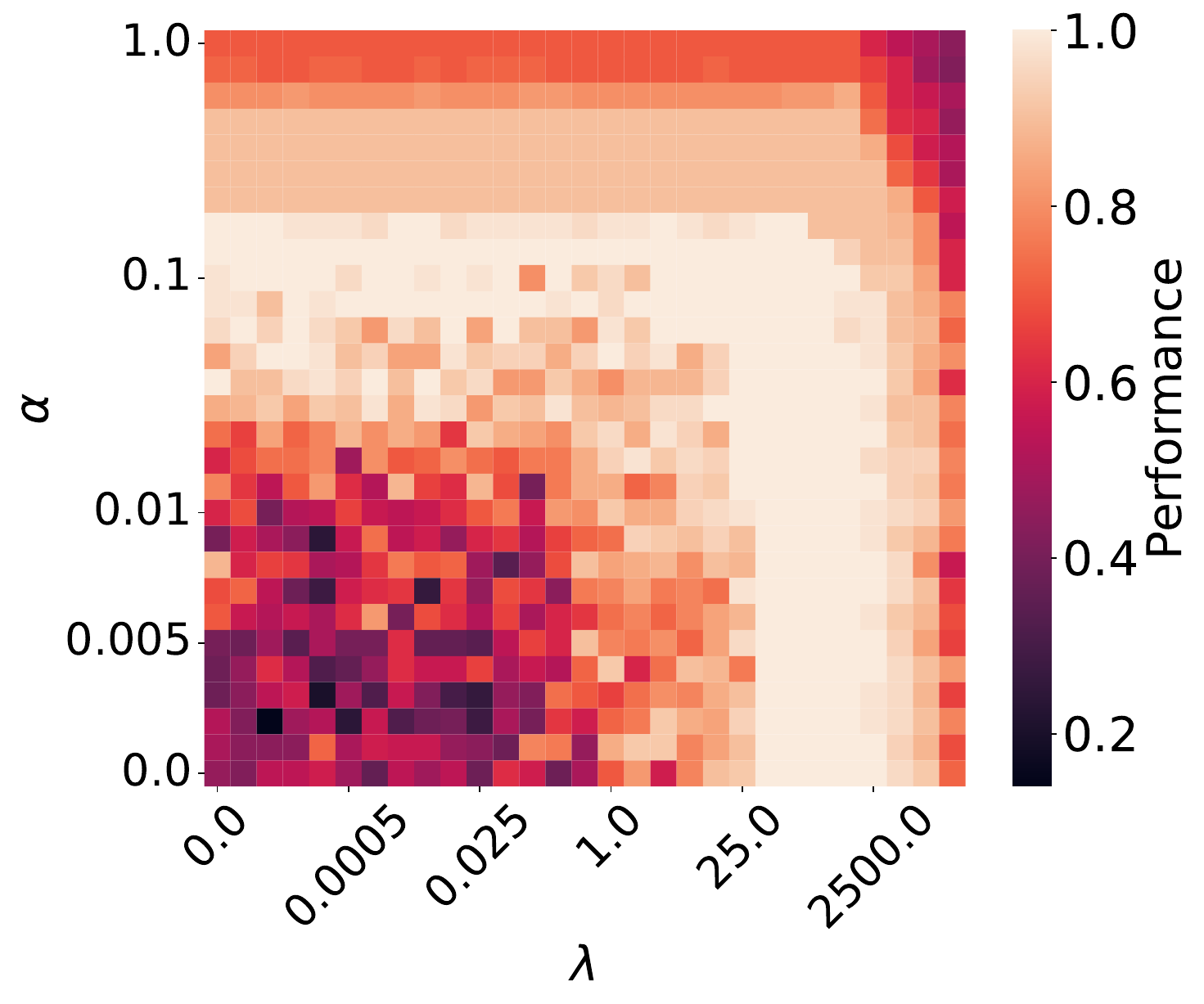}
        \caption{Acrobot}
    \end{subfigure}
    \hfill
    \begin{subfigure}[b]{0.24\textwidth}
        \centering
        \includegraphics[width=\textwidth]{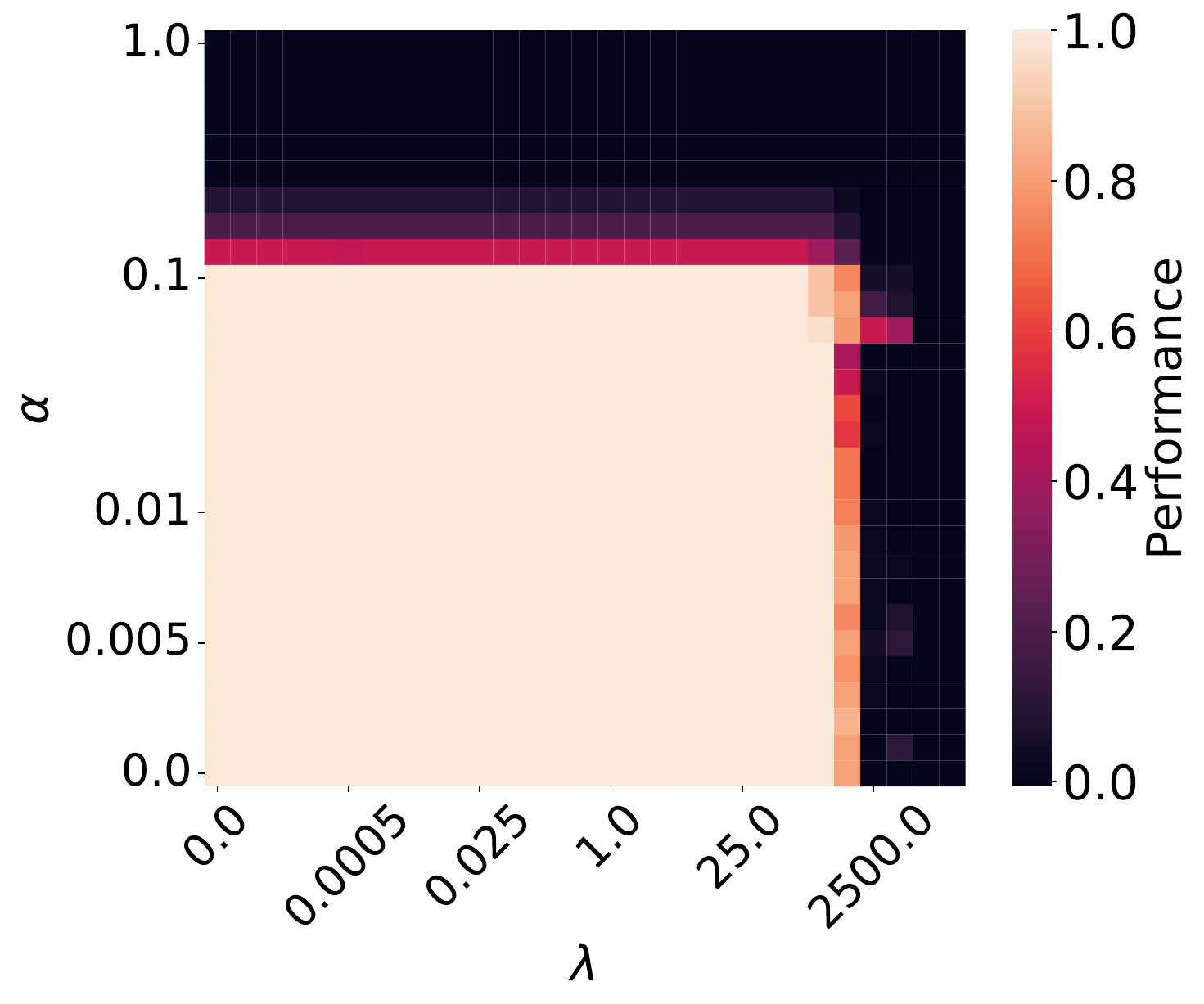}
        \caption{Catch}
    \end{subfigure}
    \hfill
    \begin{subfigure}[b]{0.24\textwidth}
        \centering
        \includegraphics[width=\textwidth]{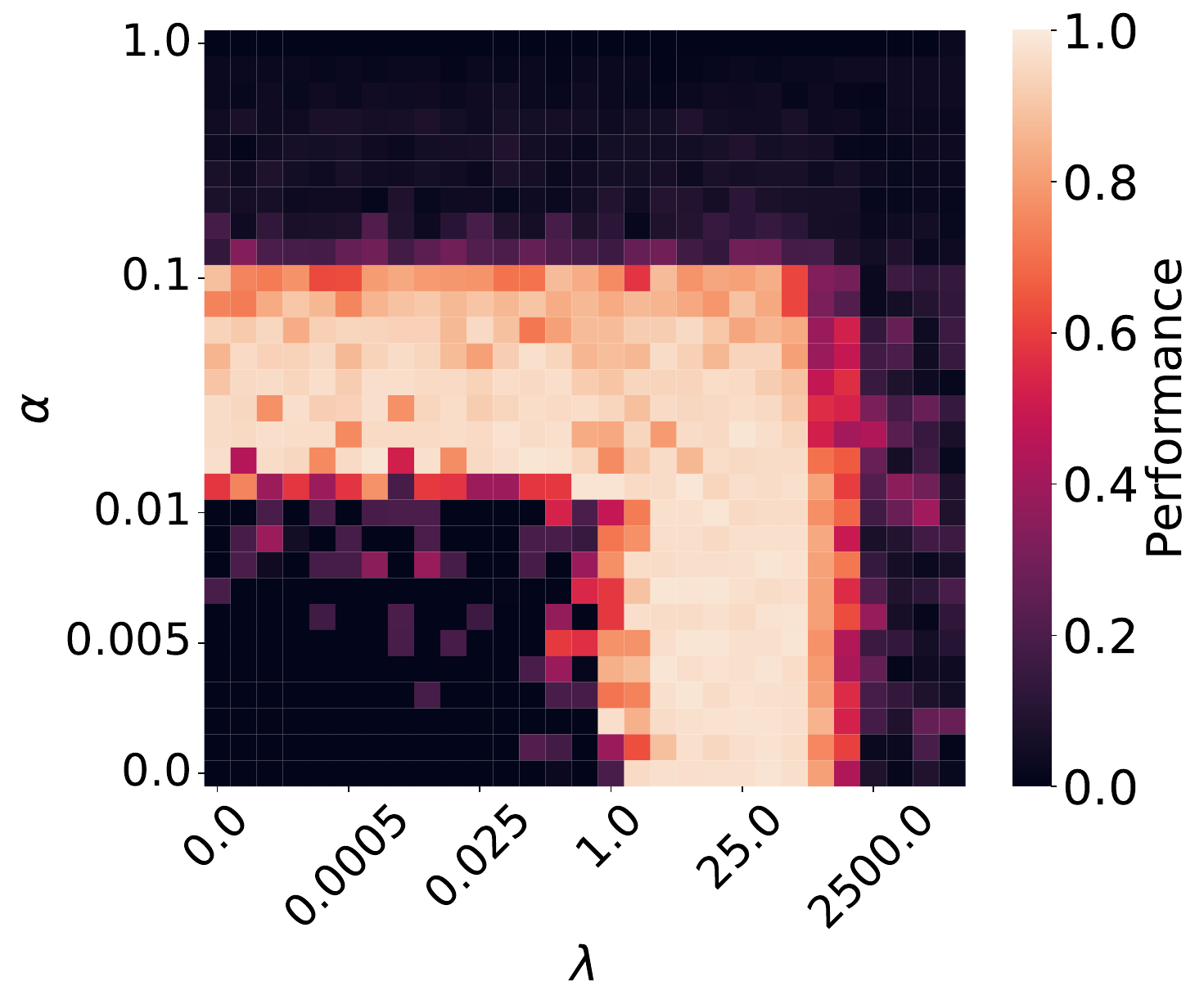}
        \caption{DeepSea}
    \end{subfigure}
    \caption{MDPO($-\mathcal{H}, D_\mathrm{KL}$) with constant temperatures on different environments}
    \label{fig:App02}
\end{figure}

\begin{figure}[htbp]
    \centering
    \begin{subfigure}[b]{0.24\textwidth}
        \centering
        \includegraphics[width=\textwidth]{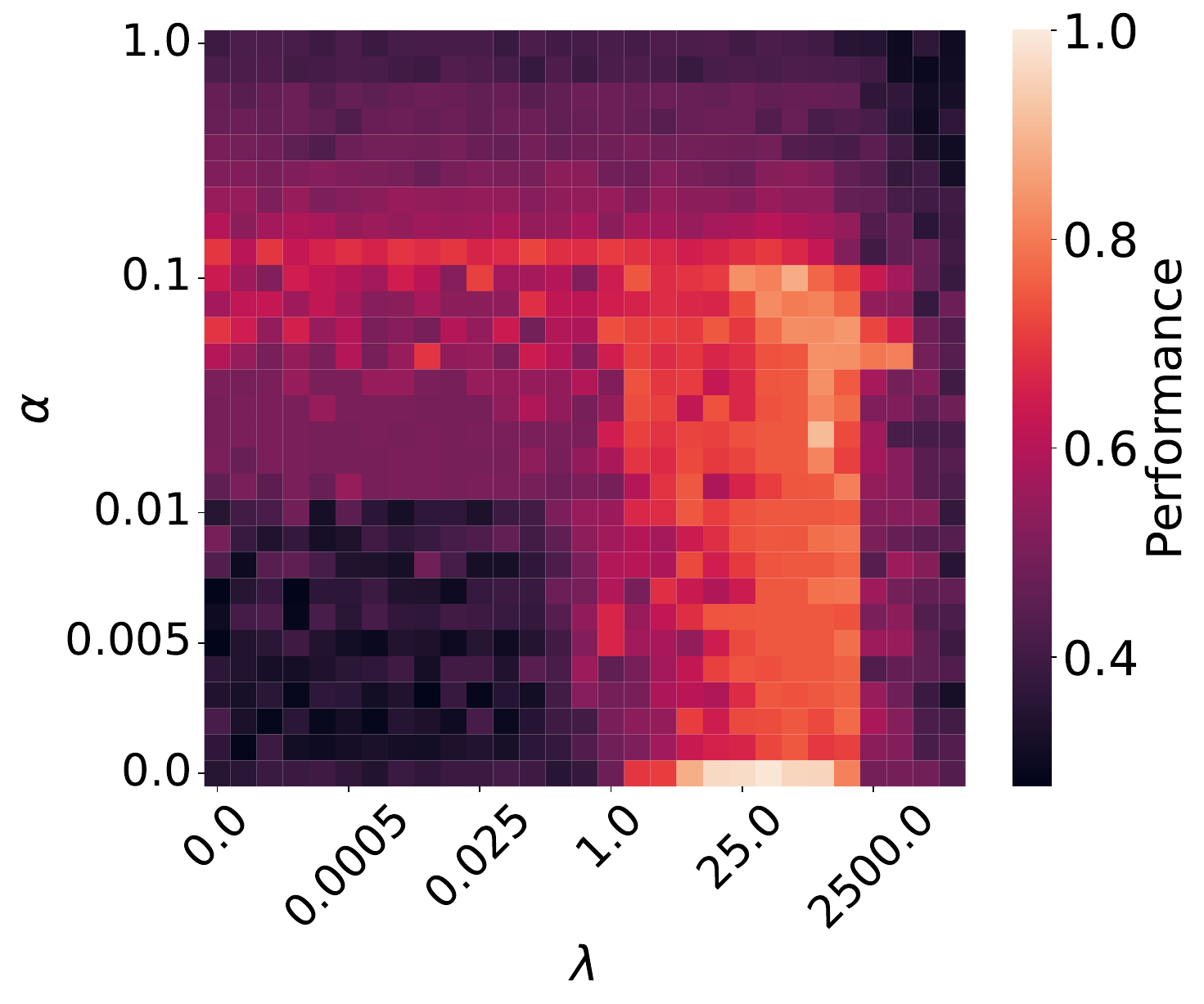}
        \caption{$h=-\mathcal{H}_{0.5}$ \& \\$D=D_\mathrm{KL}$}
    \end{subfigure}
    \hfill
    \begin{subfigure}[b]{0.24\textwidth}
        \centering
        \includegraphics[width=\textwidth]{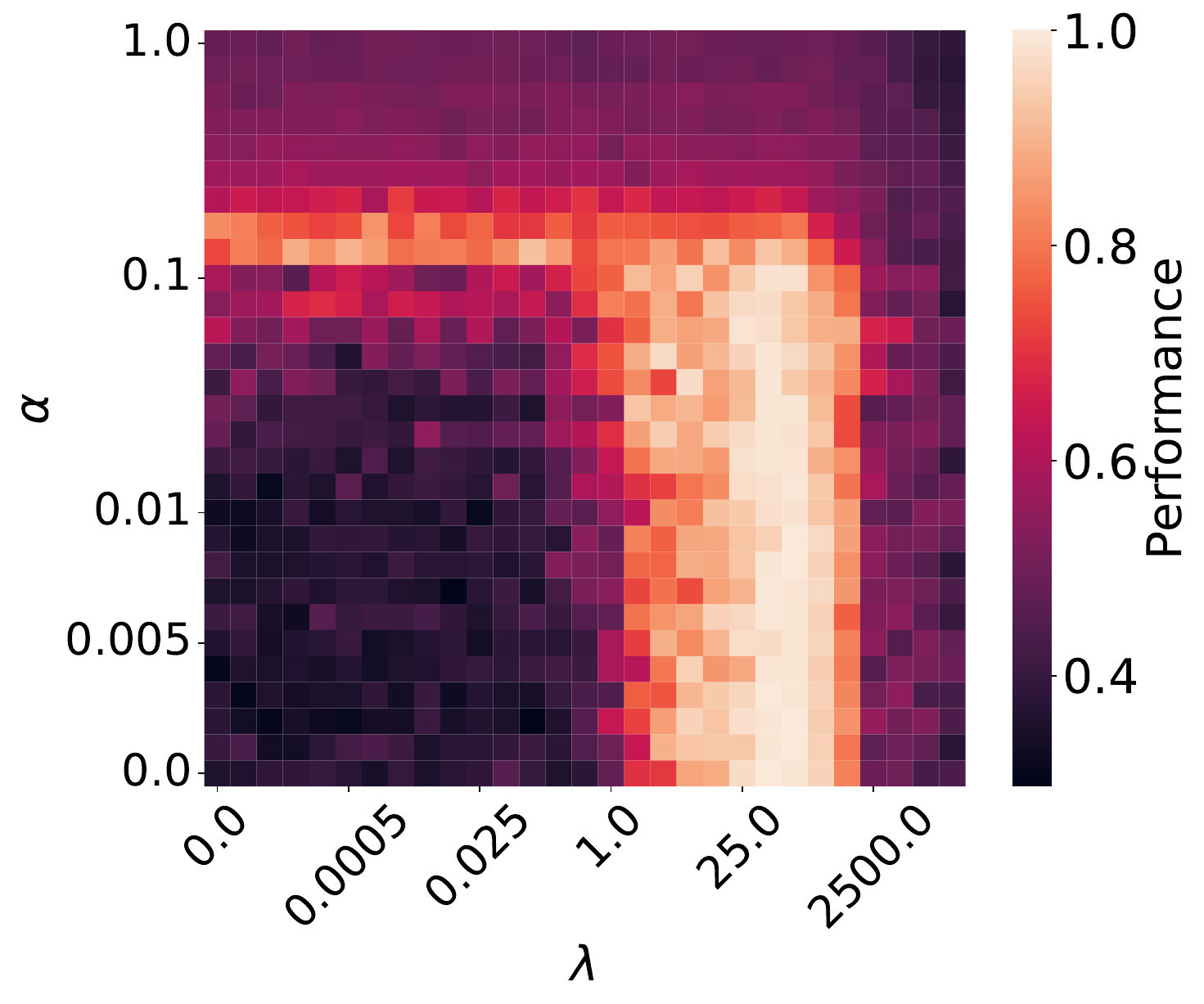}
        \caption{$h=||\cdot||_2^2$ \&\\ $D=D_\mathrm{KL}$}
    \end{subfigure}
    \hfill
    \begin{subfigure}[b]{0.24\textwidth}
        \centering
        \includegraphics[width=\textwidth]{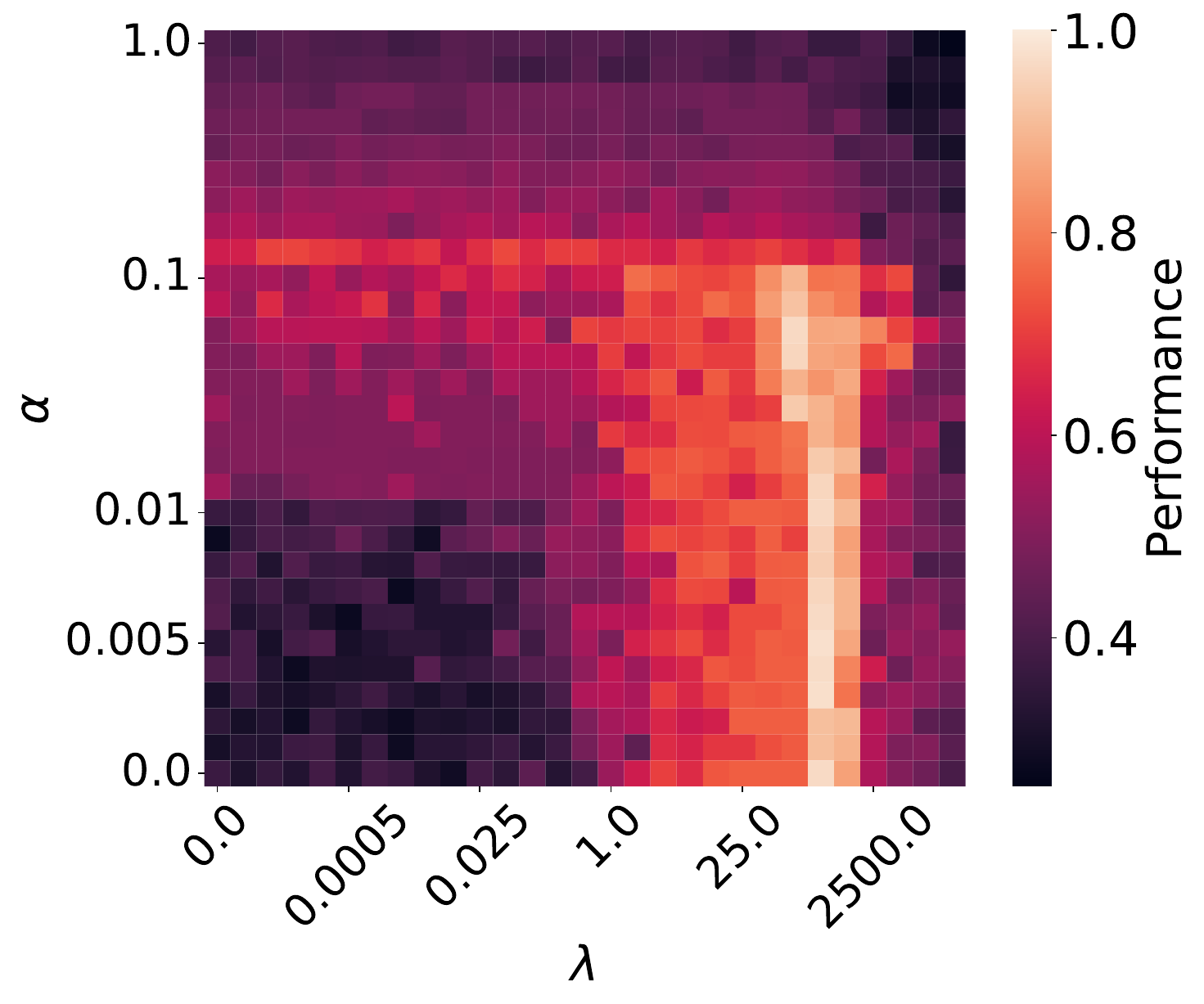}
        \caption{$h=-\mathcal{H}_{0.5}$\& \\$D=B_{-\mathcal{H}_{0.5}}$}
    \end{subfigure}
    \hfill
    \begin{subfigure}[b]{0.24\textwidth}
        \centering
        \includegraphics[width=\textwidth]{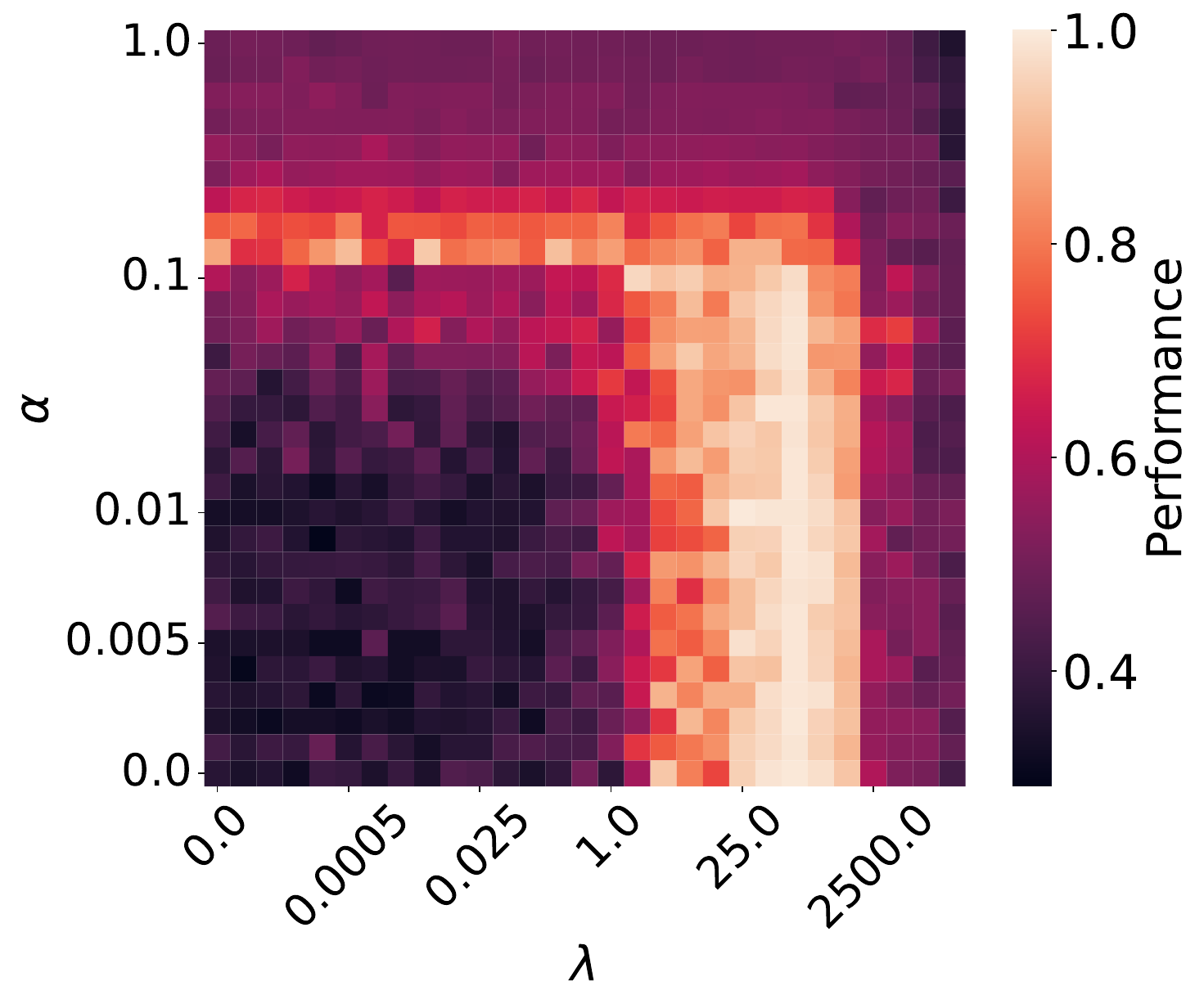}
        \caption{$h=||\cdot||_2^2$  \& \\ $D=B_{L2}$}
    \end{subfigure}
    \caption{MDPO($h, D$) for different $h, D$ pairs}
    \label{fig:App03}
\end{figure}

\begin{figure}[htbp]
    \centering
    \begin{subfigure}[b]{0.24\textwidth}
        \centering
        \includegraphics[width=\textwidth]{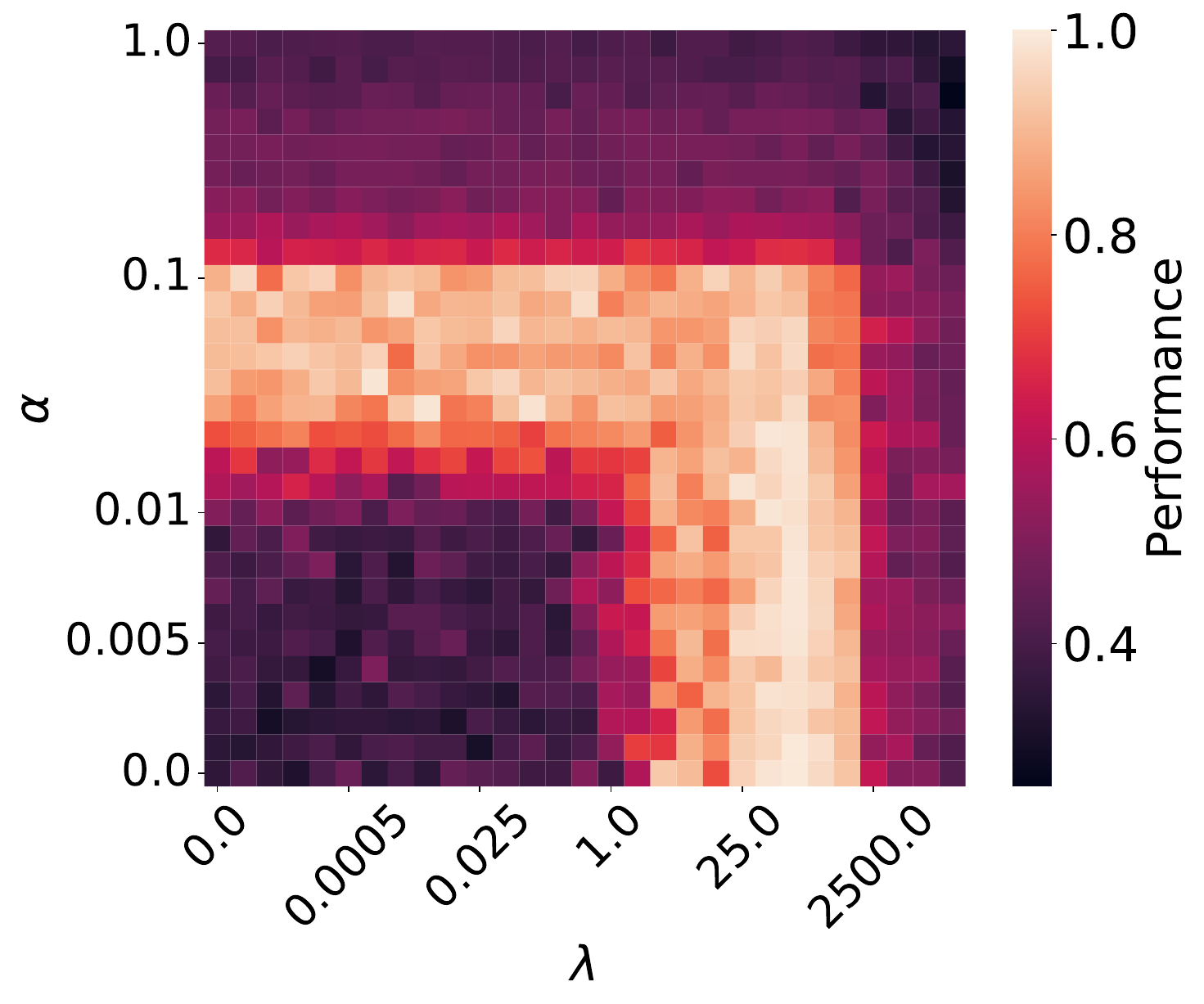}
        \caption{$h=-\mathcal{H}$ \& \\$D=B_{-\mathcal{H}_{0.5}}$}
    \end{subfigure}
    \hfill
    \begin{subfigure}[b]{0.24\textwidth}
        \centering
        \includegraphics[width=\textwidth]{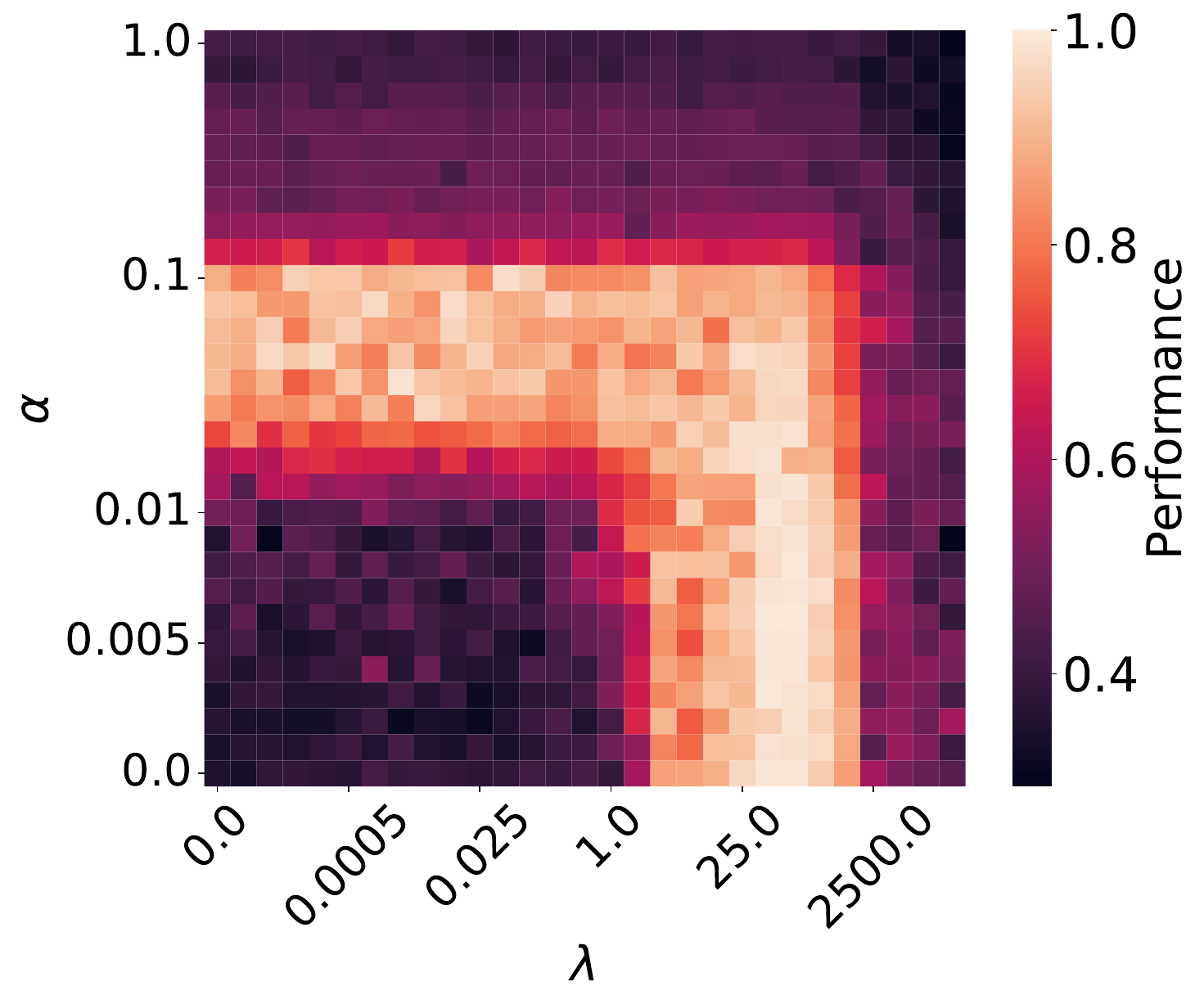}
        \caption{$h=-\mathcal{H}$ \& \\$D=B_{-\mathcal{H}_{1.5}}$}
    \end{subfigure}
    \hfill
    \begin{subfigure}[b]{0.24\textwidth}
        \centering
        \includegraphics[width=\textwidth]{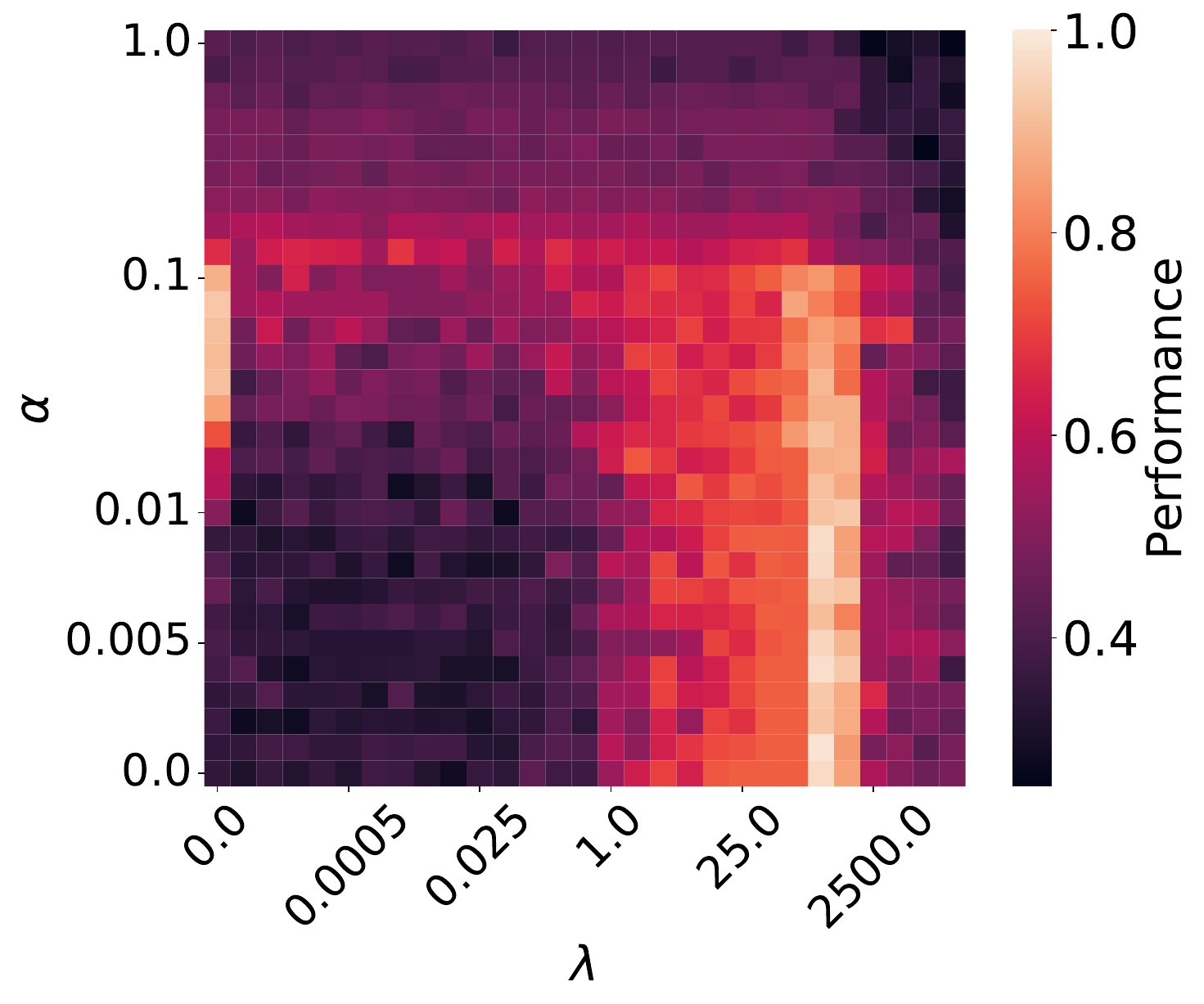}
        \caption{$h=-\mathcal{H}$ \& \\$D=B_{L2}$}
    \end{subfigure}
    \hfill
    \begin{subfigure}[b]{0.24\textwidth}
        \centering
        \includegraphics[width=\textwidth]{images_supplemental/Heatmap_AE200a.pdf}
        \caption{$h=-\mathcal{H}$ \& \\$D=D_\mathrm{KL}$}
    \end{subfigure}
    \caption{MDPO($h, D$) for different $h, D$ pairs}
    \label{fig:App04}
\end{figure}

\begin{figure}[htbp]
    \centering
    \begin{subfigure}[b]{0.32\textwidth}
        \centering
        \includegraphics[width=\textwidth]{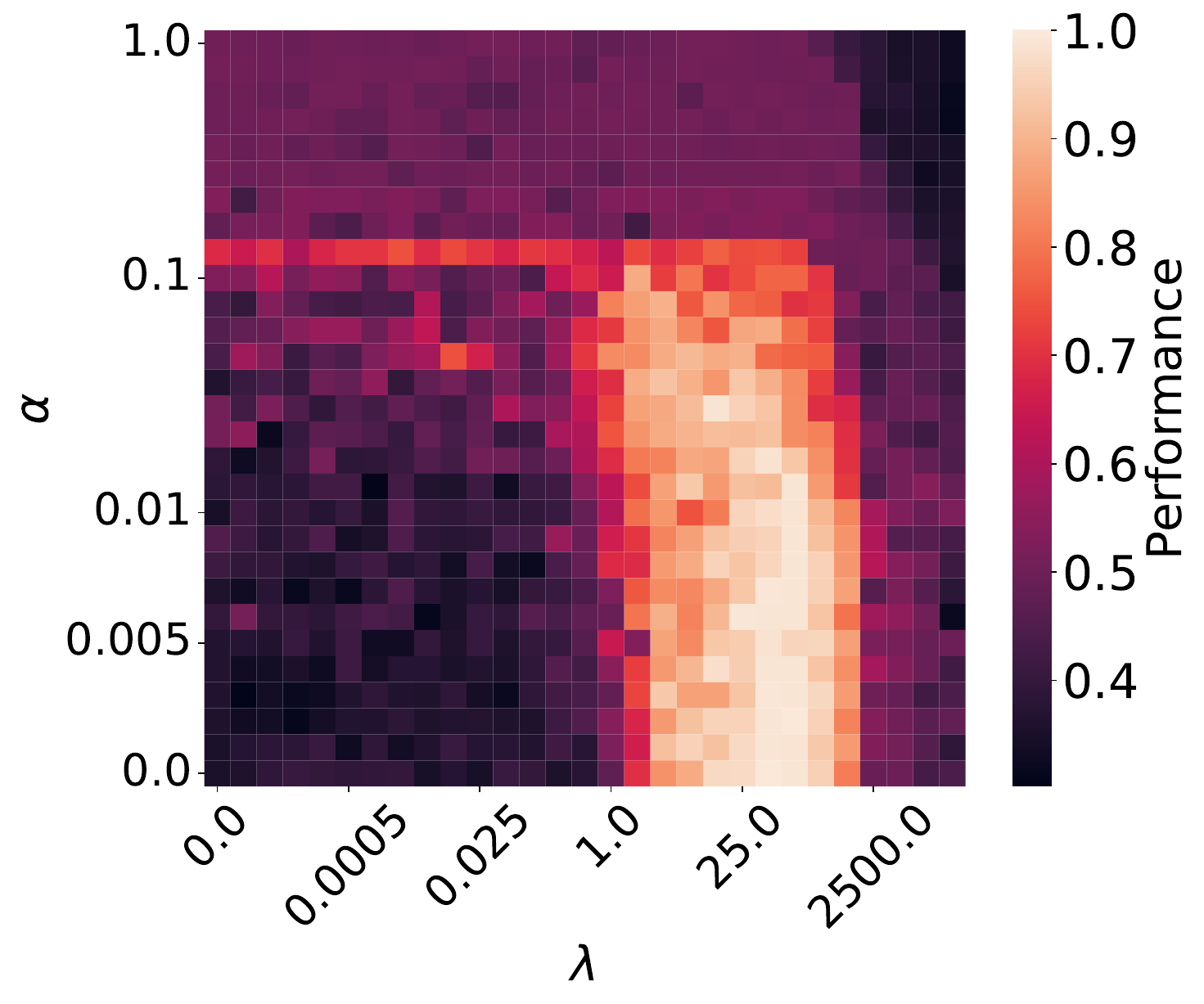}
        \caption{$h=\max$ \& $D=D_\mathrm{KL}$}
    \end{subfigure}
    \hfill
    \begin{subfigure}[b]{0.32\textwidth}
        \centering
        \includegraphics[width=\textwidth]{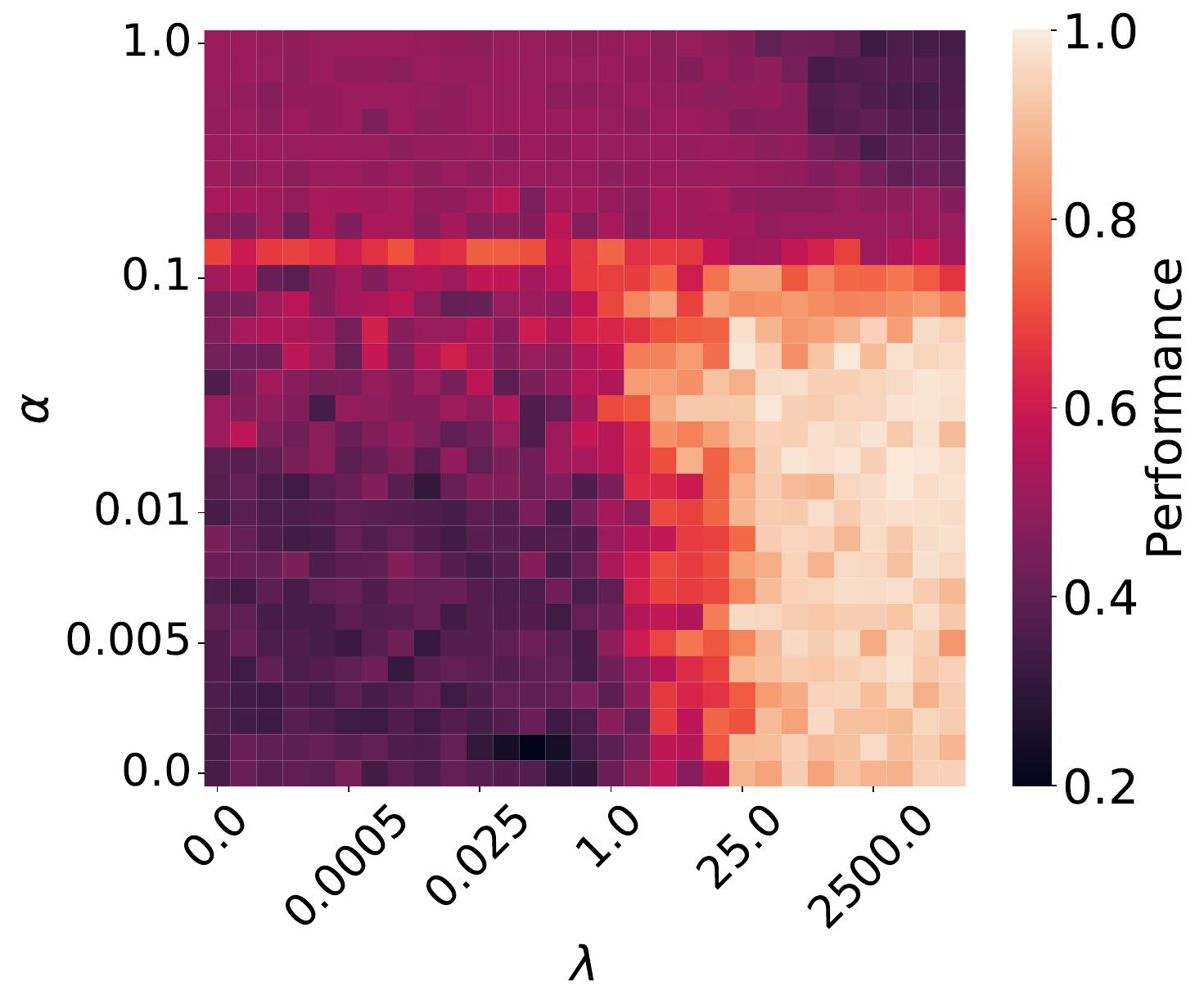}
        \caption{$h=\max$ \& $D=B_\mathrm{max}$}
    \end{subfigure}
    \hfill
    \begin{subfigure}[b]{0.32\textwidth}
        \centering
        \includegraphics[width=\textwidth]{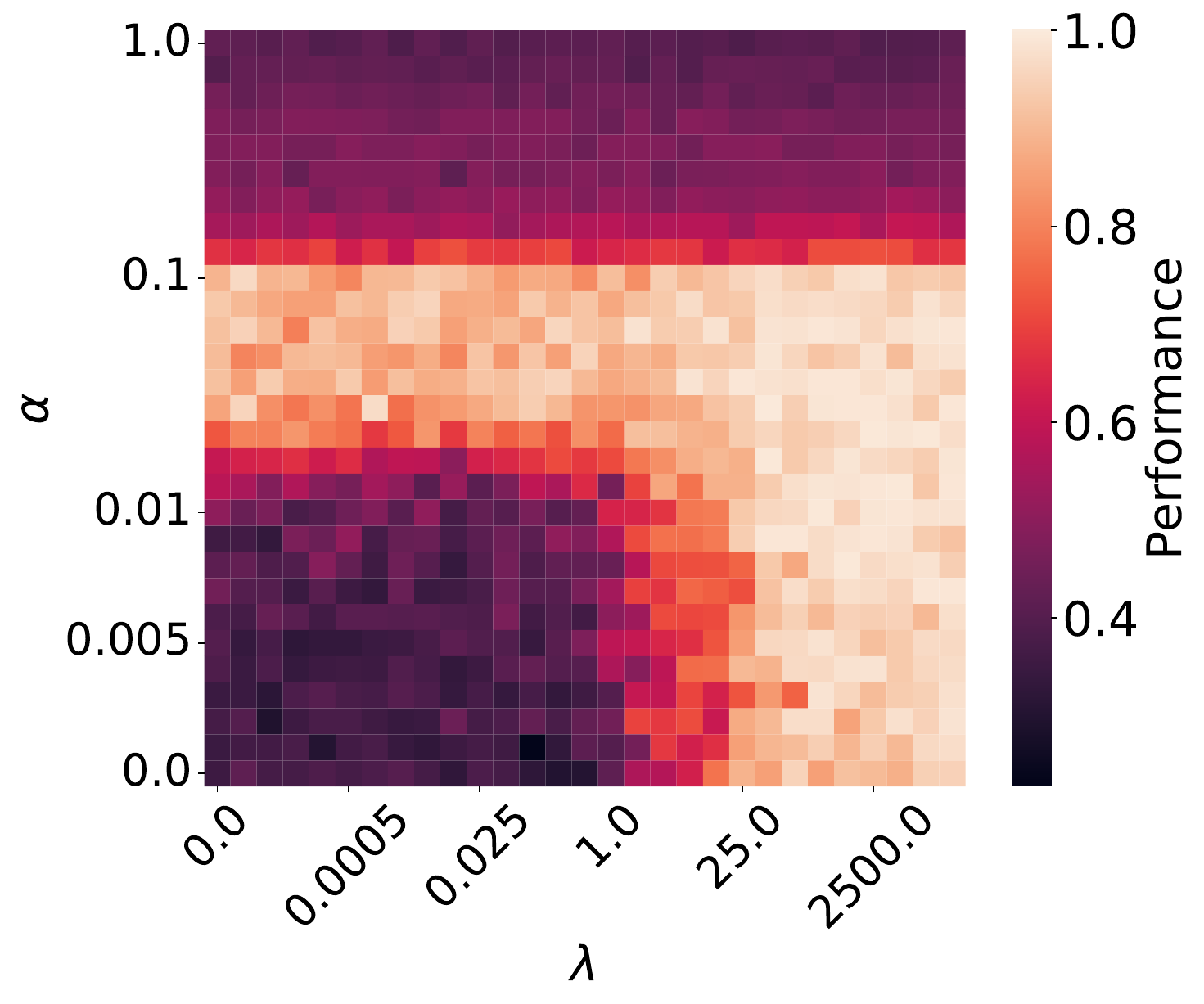}
        \caption{$h=-\mathcal{H}$ \& $D=B_\mathrm{max}$}
    \end{subfigure}
    \caption{MDPO($h, D$) for different $h, D$ pairs}
    \label{fig:App4b}
\end{figure}

% %%% Change alpha scheduling 

\begin{figure}[htbp]
    \centering
    \begin{subfigure}[b]{0.24\textwidth}
        \centering
        \includegraphics[width=\textwidth]{images_supplemental/Heatmap_AE200a.pdf}
        \caption{$\alpha=\mathrm{const.}$, $\lambda=\mathrm{const.}$}
    \end{subfigure}
    \hfill
    \begin{subfigure}[b]{0.24\textwidth}
        \centering
        \includegraphics[width=\textwidth]{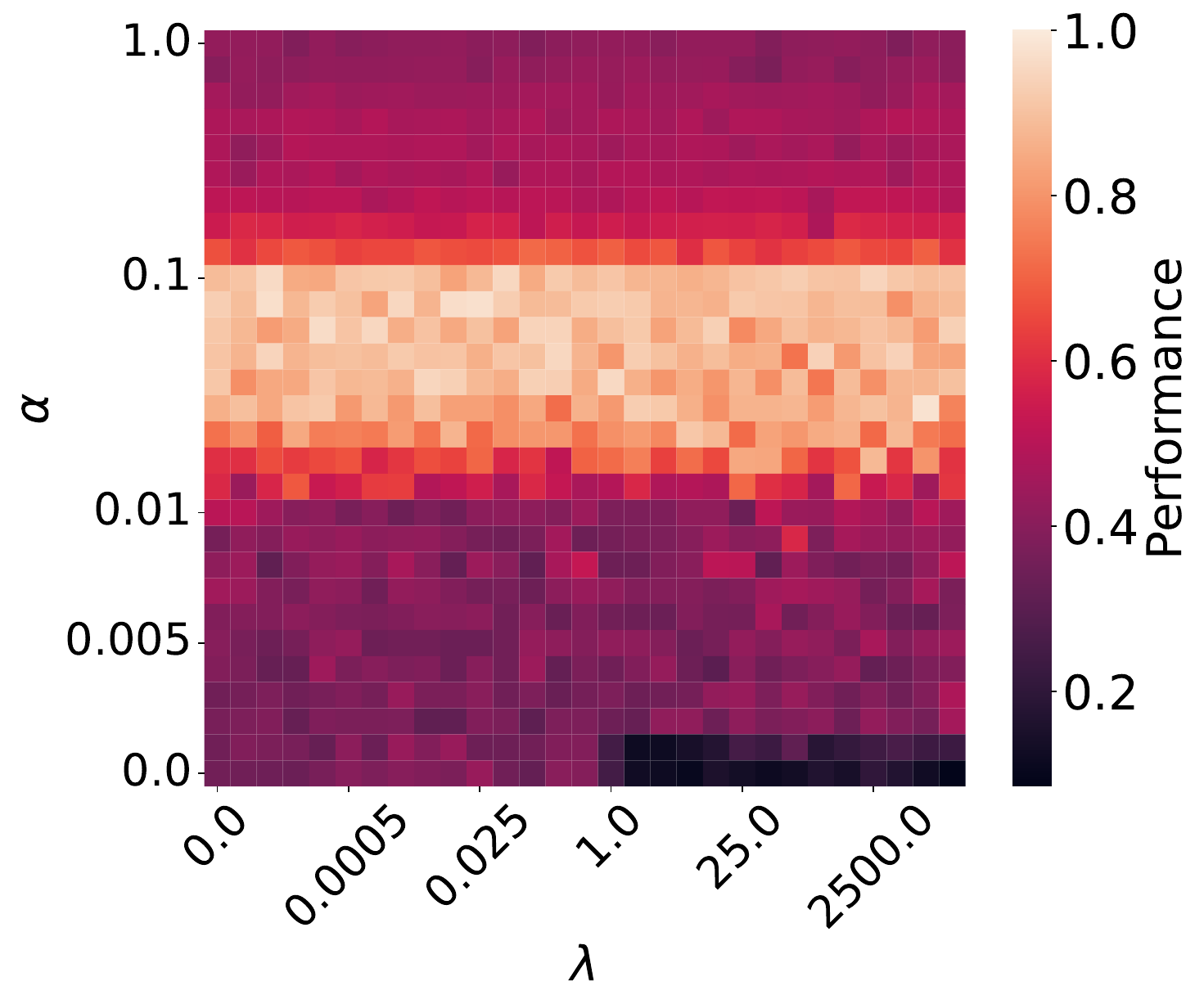}
        \caption{$\alpha=\mathrm{const.}$, $\lambda=\mathrm{lin.\,anneal}$}
    \end{subfigure}
    \hfill
    \begin{subfigure}[b]{0.24\textwidth}
        \centering
        \includegraphics[width=\textwidth]{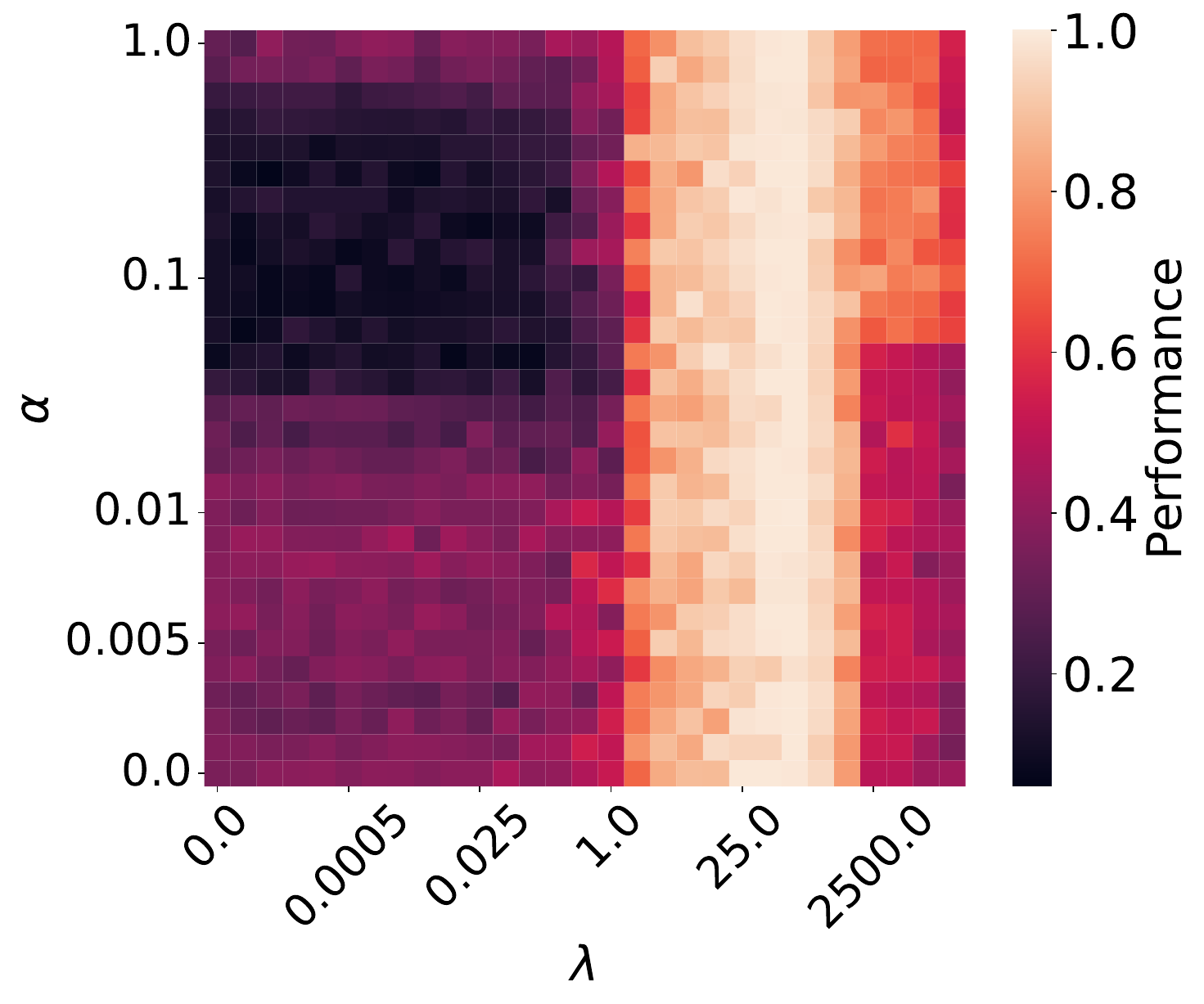}
        \caption{$\alpha=\mathrm{lin.\,anneal}$, $\lambda=\mathrm{const.}$}
    \end{subfigure}
    \hfill
    \begin{subfigure}[b]{0.24\textwidth}
        \centering
        \includegraphics[width=\textwidth]{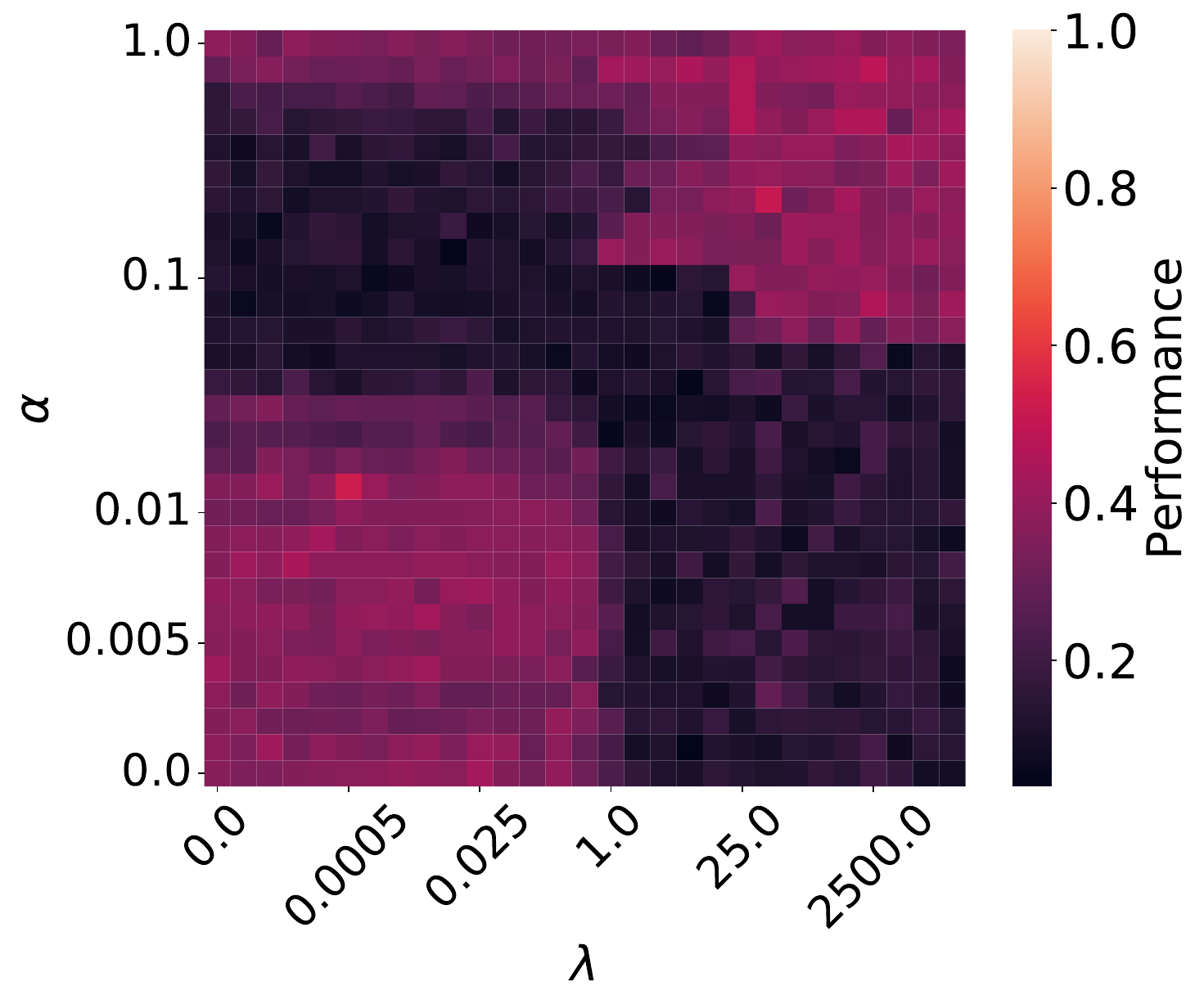}
        \caption{$\alpha=\mathrm{lin.\,anneal}$, $\lambda=\mathrm{lin.\,anneal}$}
    \end{subfigure}
    \caption{MDPO($-\mathcal{H}, D_\mathrm{KL}$) for different temperature scheduling schemes}
    \label{fig:App05}
\end{figure}

\begin{figure}[htbp]
    \centering
    \begin{subfigure}[b]{0.24\textwidth}
        \centering
        \includegraphics[width=\textwidth]{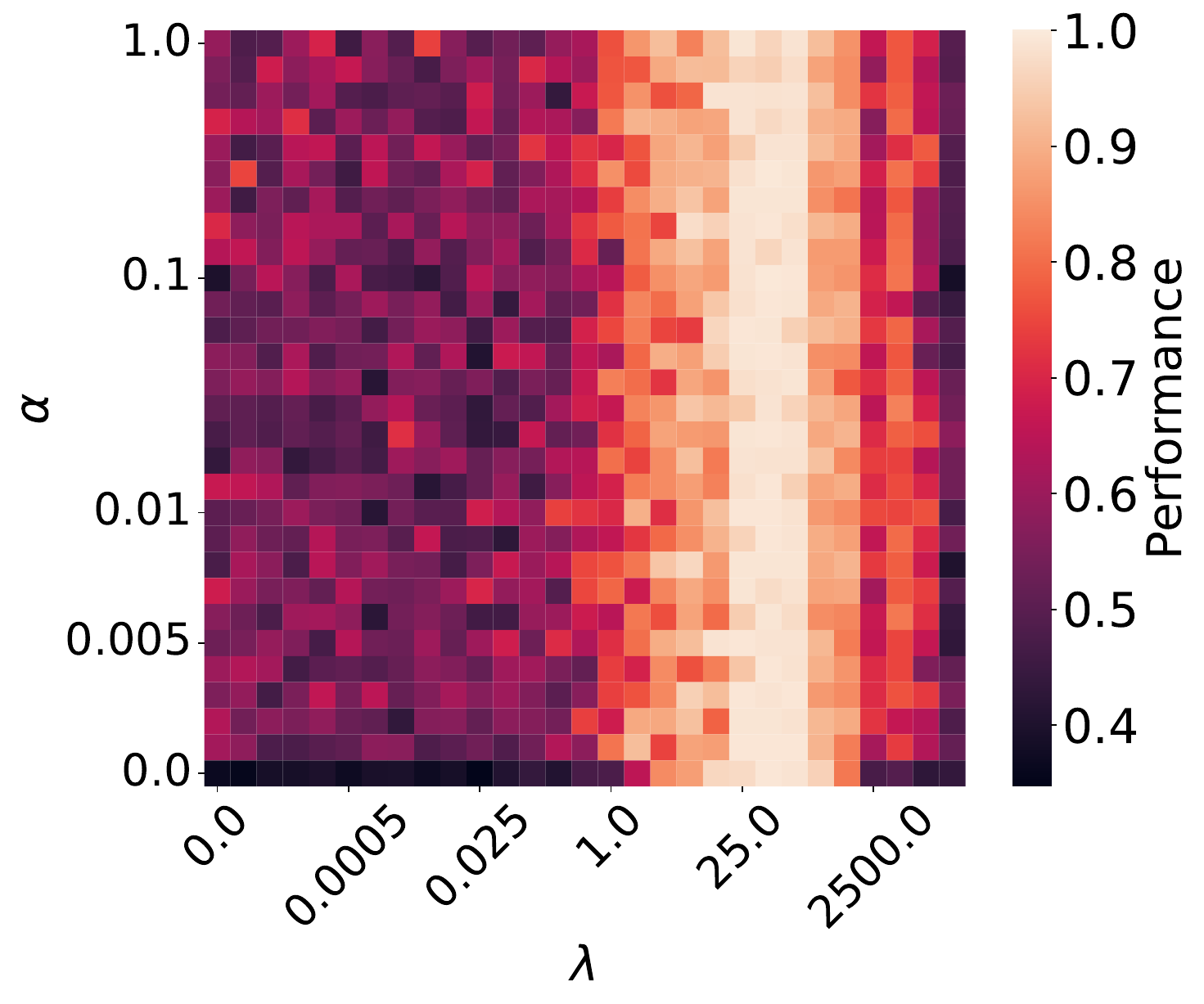}
        \caption{$\alpha=\mathrm{learned}$\\const., \\$\lambda=\mathrm{const.}$}
    \end{subfigure}
    \hfill
    \begin{subfigure}[b]{0.24\textwidth}
        \centering
        \includegraphics[width=\textwidth]{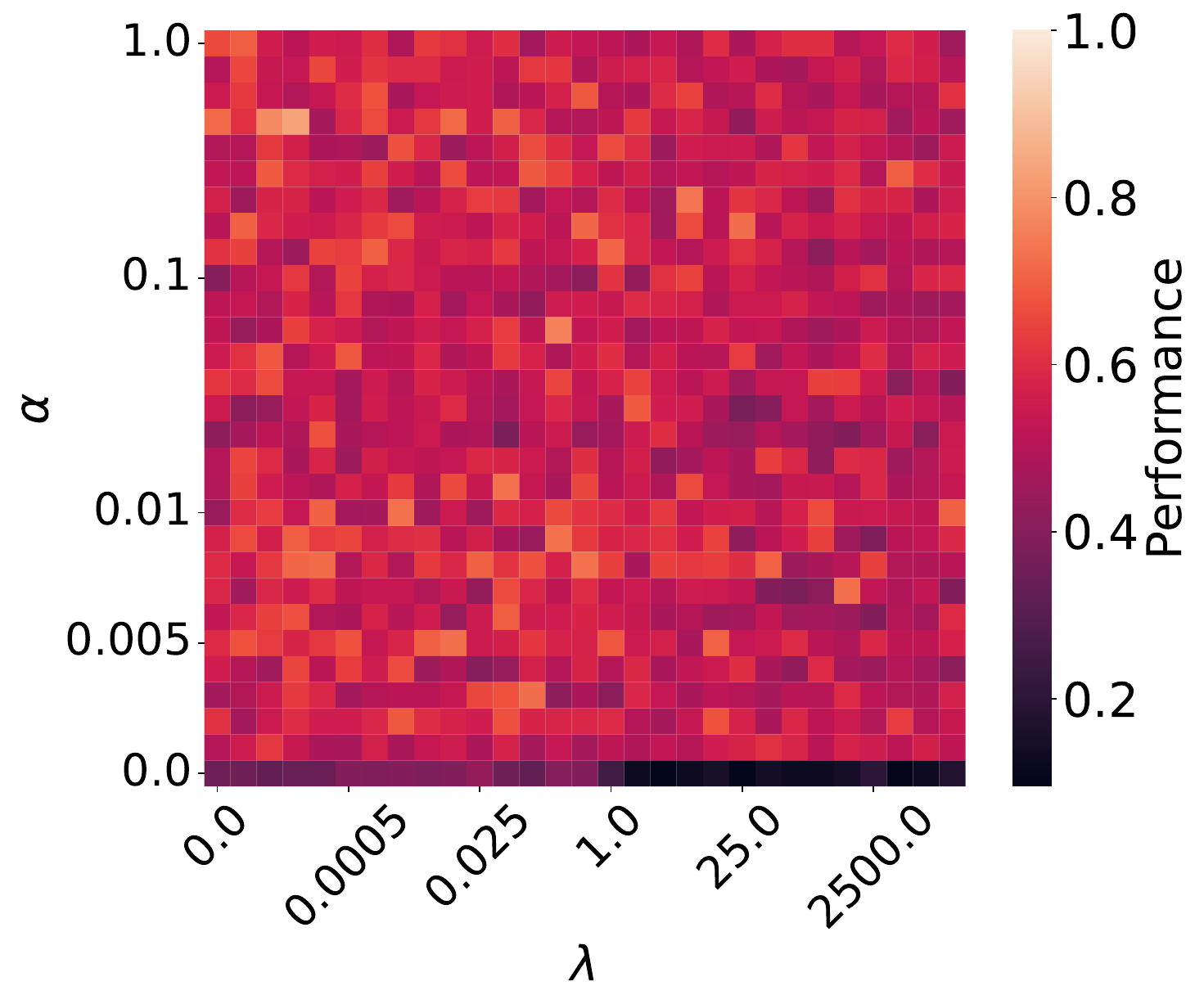}
        \caption{$\alpha=\mathrm{learned}$\\const., \\$\lambda=\mathrm{lin.\,anneal}$}
    \end{subfigure}
    \hfill
    \begin{subfigure}[b]{0.24\textwidth}
        \centering
        \includegraphics[width=\textwidth]{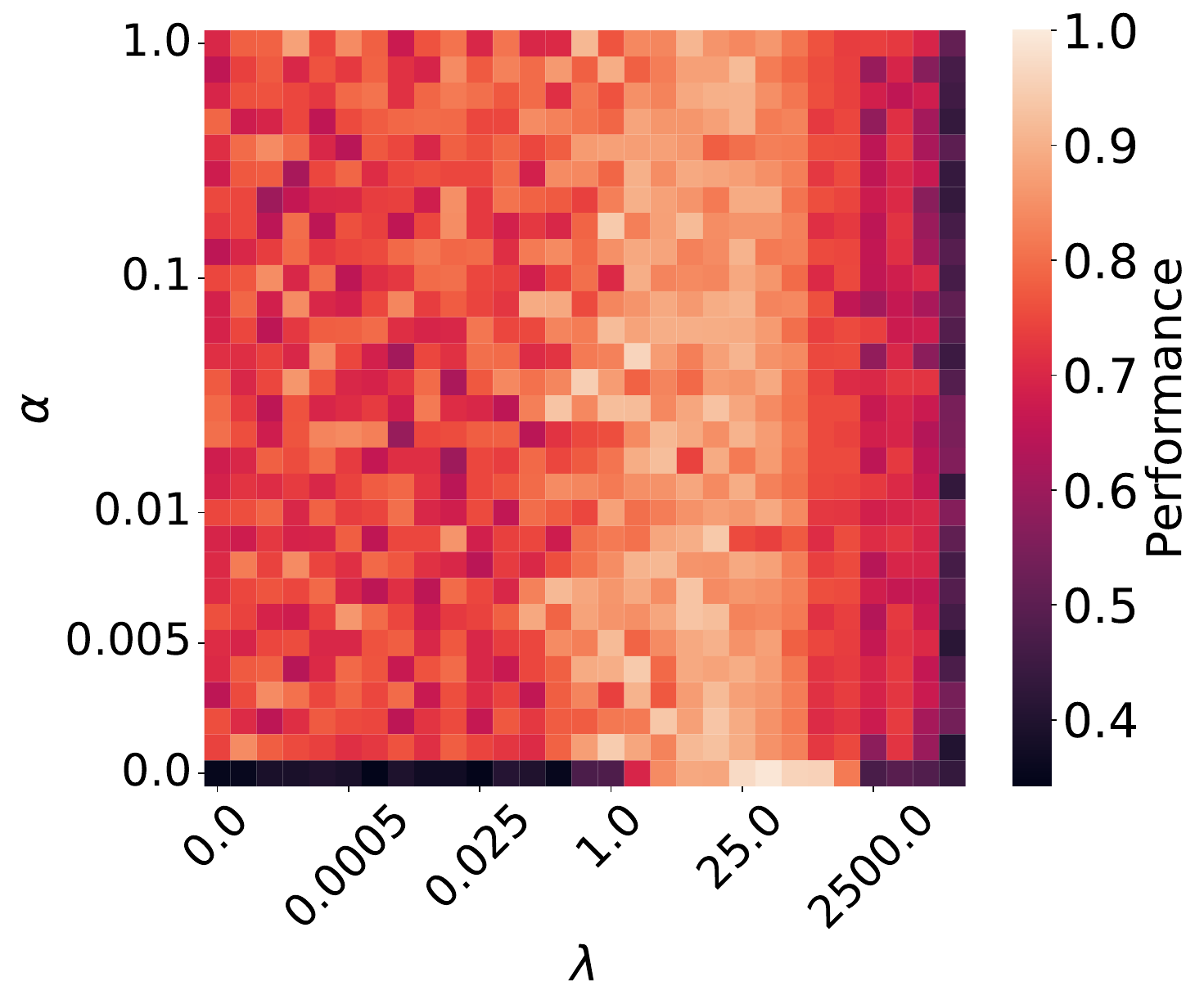}
        \caption{$\alpha=\mathrm{learned}$\\lin. anneal, \\$\lambda=\mathrm{const.}$}
    \end{subfigure}
    \hfill
    \begin{subfigure}[b]{0.24\textwidth}
        \centering
        \includegraphics[width=\textwidth]{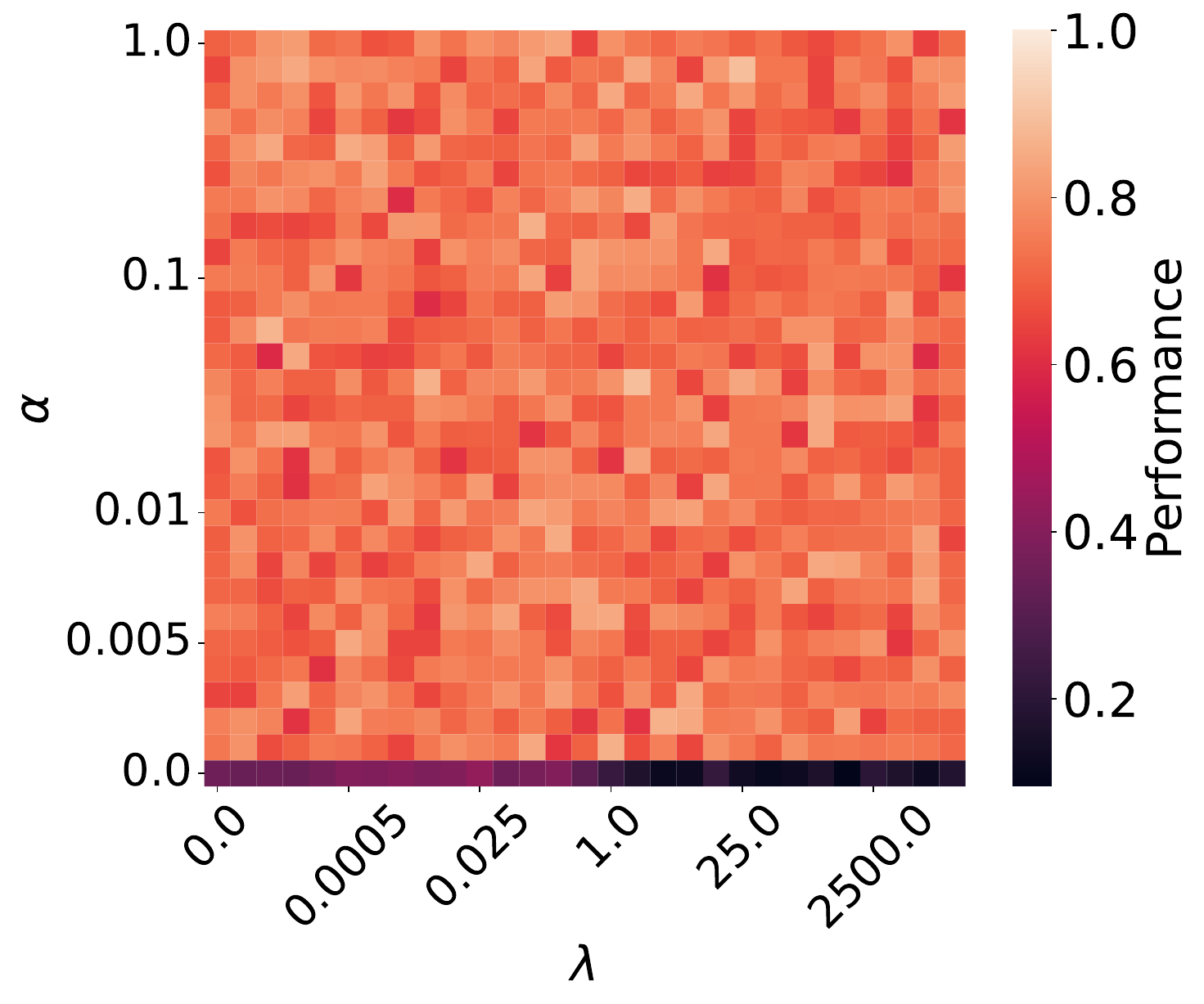}
        \caption{$\alpha=\mathrm{learned}$\\lin. anneal, \\$\lambda=\mathrm{lin.\,anneal}$}
    \end{subfigure}
    \caption{MDPO($-\mathcal{H}, D_\mathrm{KL}$) for different temperature scheduling schemes}
    \label{fig:App06}
\end{figure}

\begin{figure}[htbp]
    \centering
    \begin{subfigure}[b]{0.24\textwidth}
        \centering
        \includegraphics[width=\textwidth]{images_supplemental/Heatmap_AE204a.pdf}
        \caption{$\alpha=\mathrm{const.}$, $\lambda=\mathrm{const.}$}
    \end{subfigure}
    \hfill
    \begin{subfigure}[b]{0.24\textwidth}
        \centering
        \includegraphics[width=\textwidth]{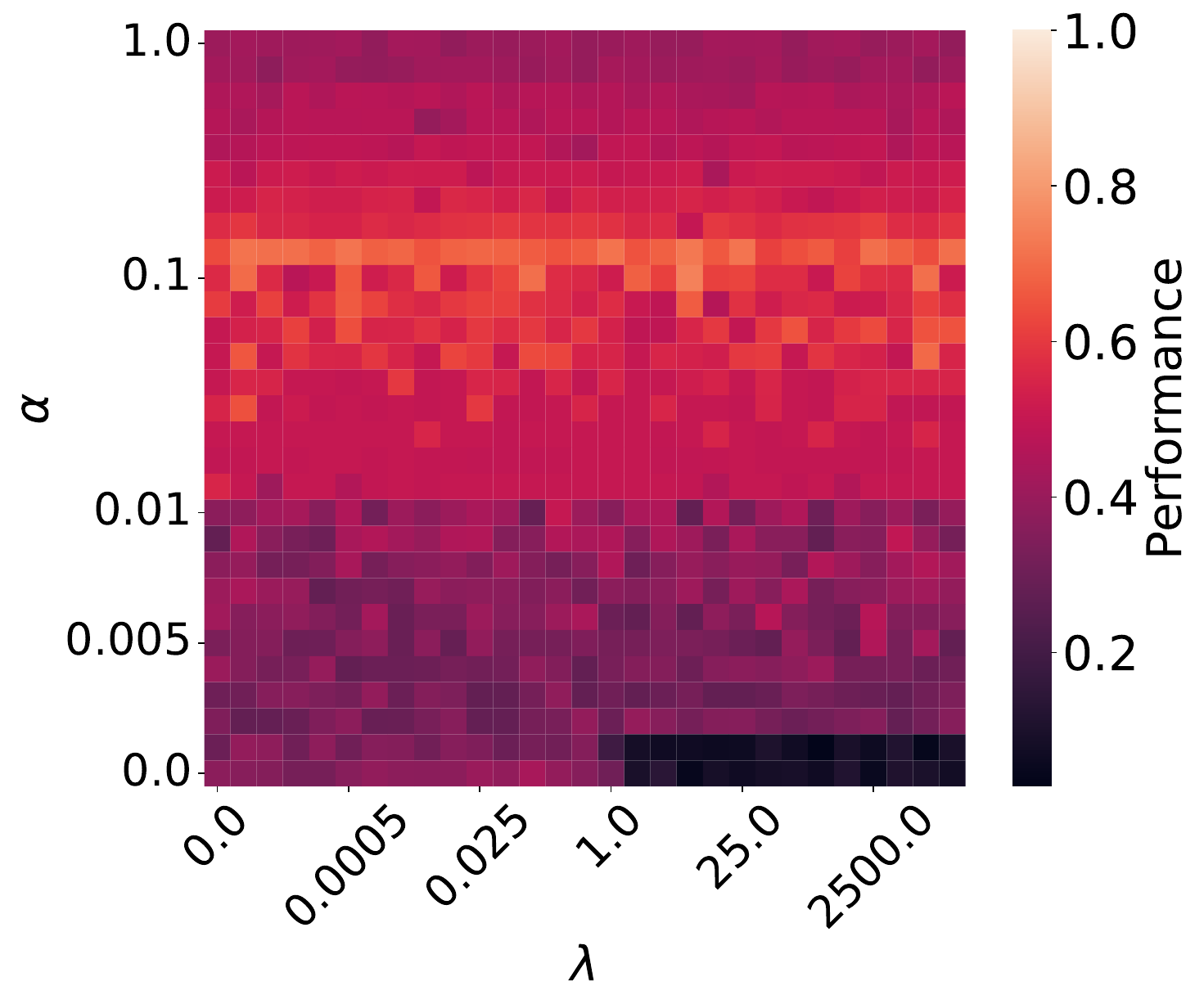}
        \caption{$\alpha=\mathrm{const.}$, $\lambda=\mathrm{lin.\,anneal}$}
    \end{subfigure}
    \hfill
    \begin{subfigure}[b]{0.24\textwidth}
        \centering
        \includegraphics[width=\textwidth]{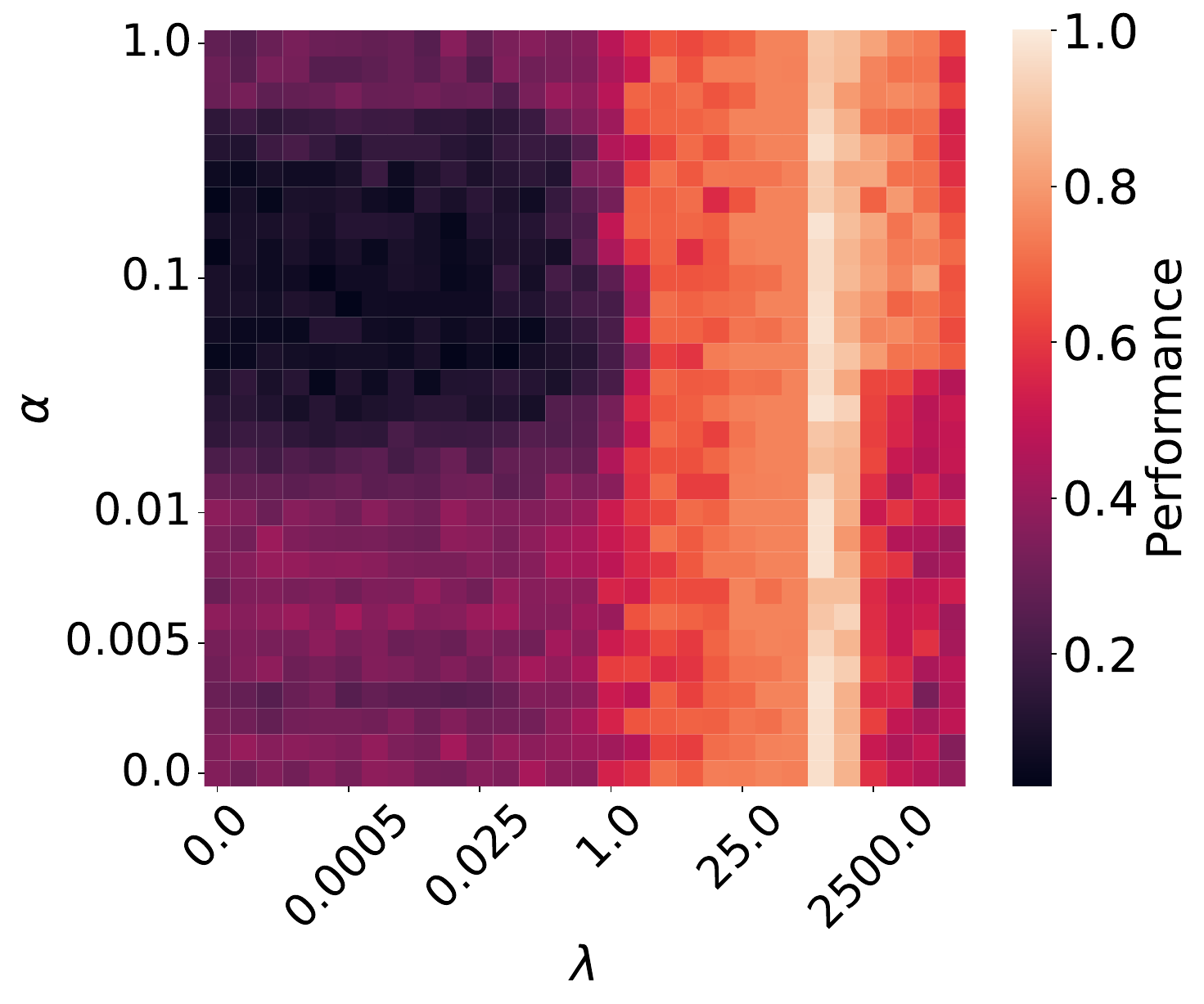}
        \caption{$\alpha=\mathrm{lin.\,anneal}$, $\lambda=\mathrm{const.}$}
    \end{subfigure}
    \hfill
    \begin{subfigure}[b]{0.24\textwidth}
        \centering
        \includegraphics[width=\textwidth]{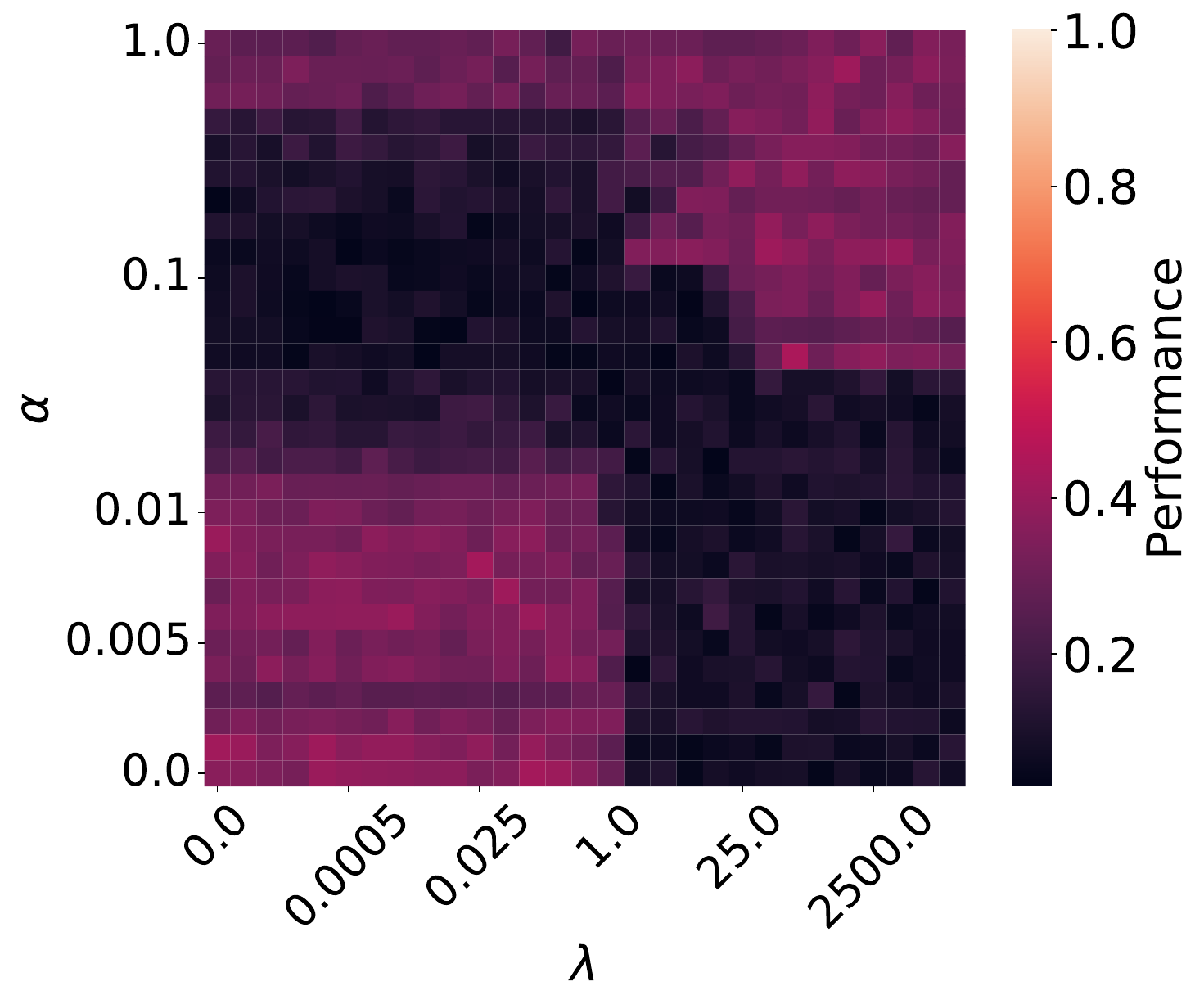}
        \caption{$\alpha=\mathrm{lin.\,anneal}$, $\lambda=\mathrm{lin.\,anneal}$}
    \end{subfigure}
    \caption{MDPO($-\mathcal{H}_{0.5}, B_{-\mathcal{H}_{0.5}}$) for different temperature scheduling schemes}
    \label{fig:App07}
\end{figure}

\begin{figure}[htbp]
    \centering
    \begin{subfigure}[b]{0.24\textwidth}
        \centering
        \includegraphics[width=\textwidth]{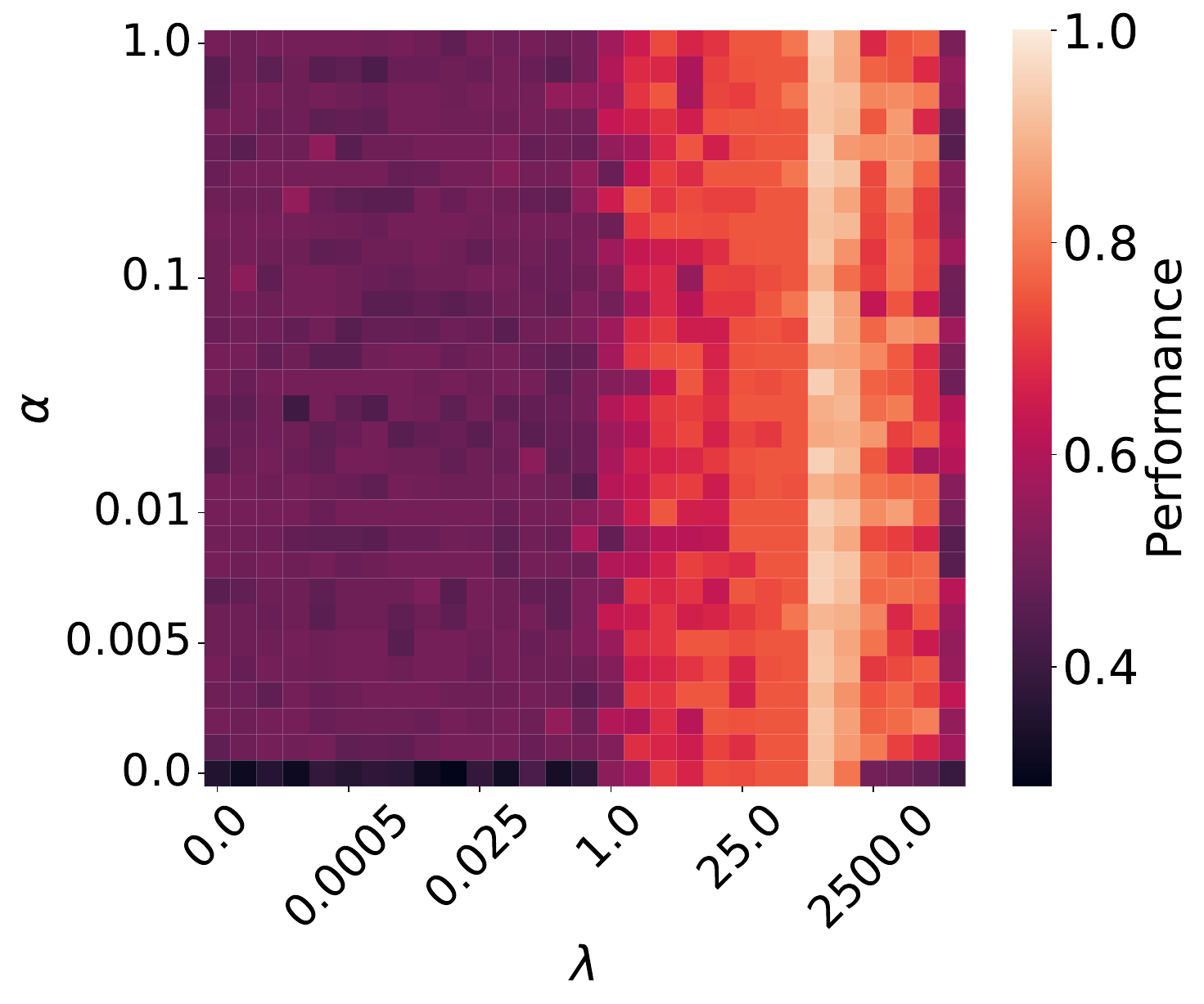}
        \caption{$\alpha=\mathrm{learned}$\\const., \\$\lambda=\mathrm{const.}$}
    \end{subfigure}
    \hfill
    \begin{subfigure}[b]{0.24\textwidth}
        \centering
        \includegraphics[width=\textwidth]{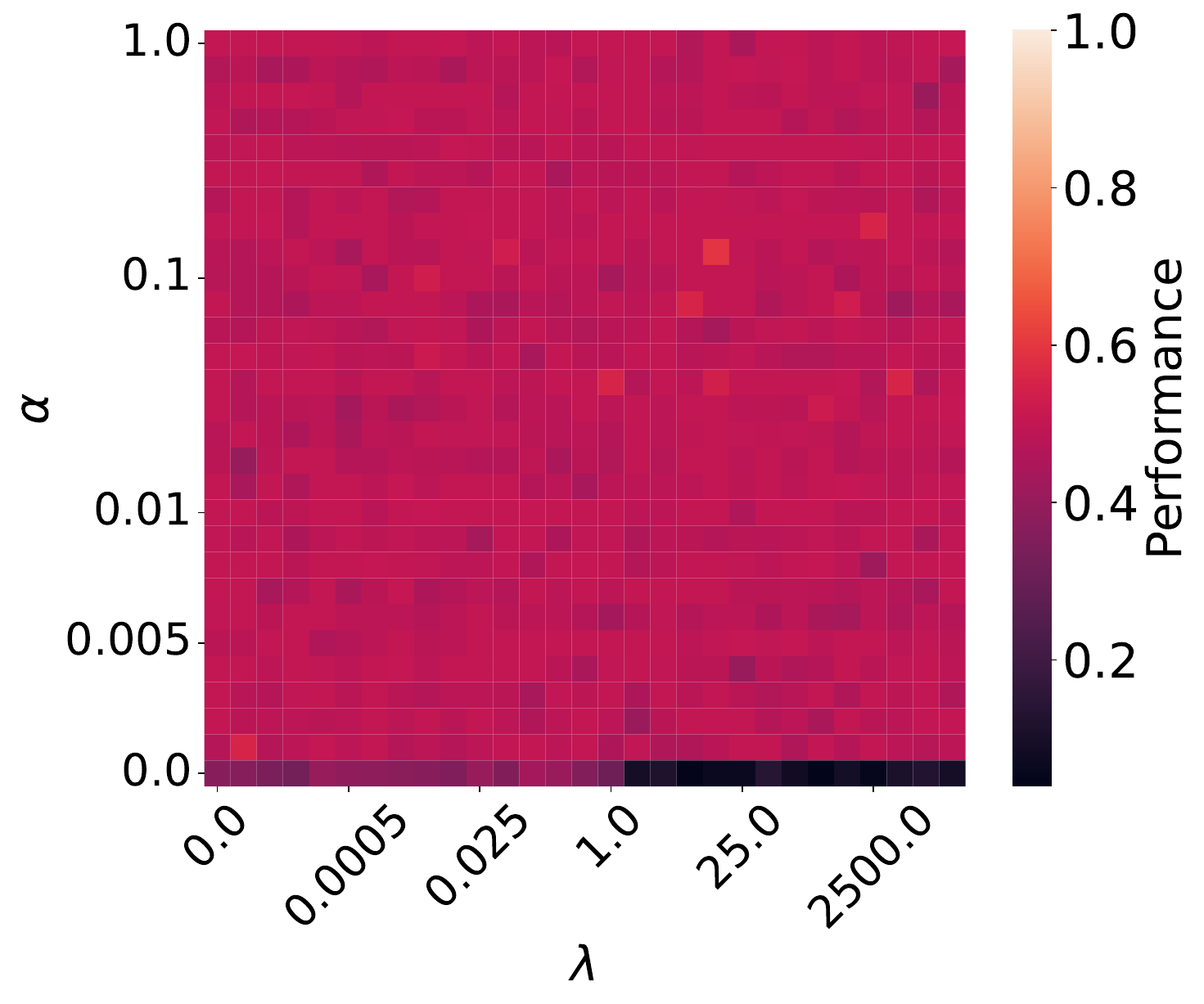}
        \caption{$\alpha=\mathrm{learned}$\\const., \\$\lambda=\mathrm{lin.\,anneal}$}
    \end{subfigure}
    \hfill
    \begin{subfigure}[b]{0.24\textwidth}
        \centering
        \includegraphics[width=\textwidth]{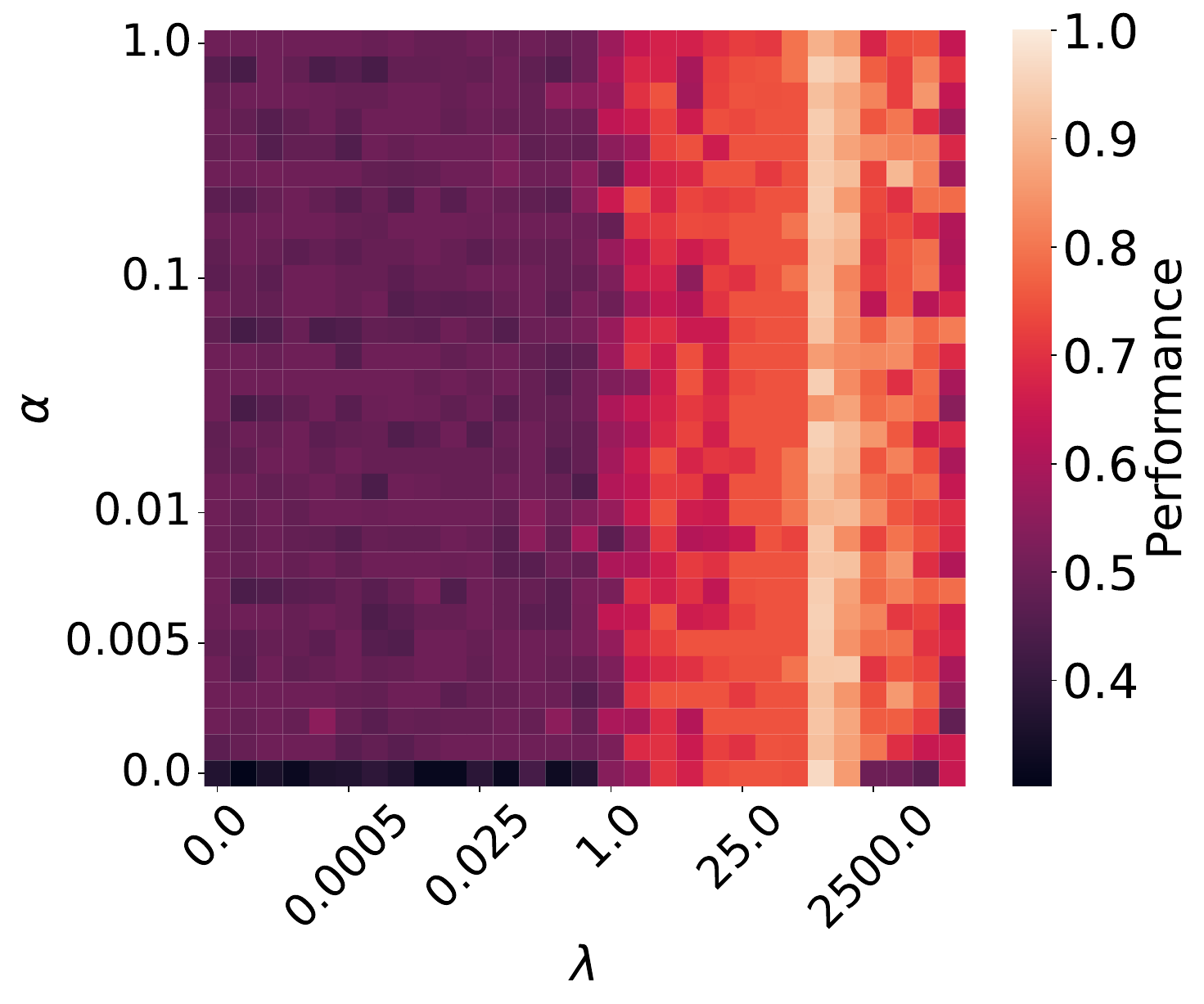}
        \caption{$\alpha=\mathrm{learned}$\\lin. anneal, \\$\lambda=\mathrm{const.}$}
    \end{subfigure}
    \hfill
    \begin{subfigure}[b]{0.24\textwidth}
        \centering
        \includegraphics[width=\textwidth]{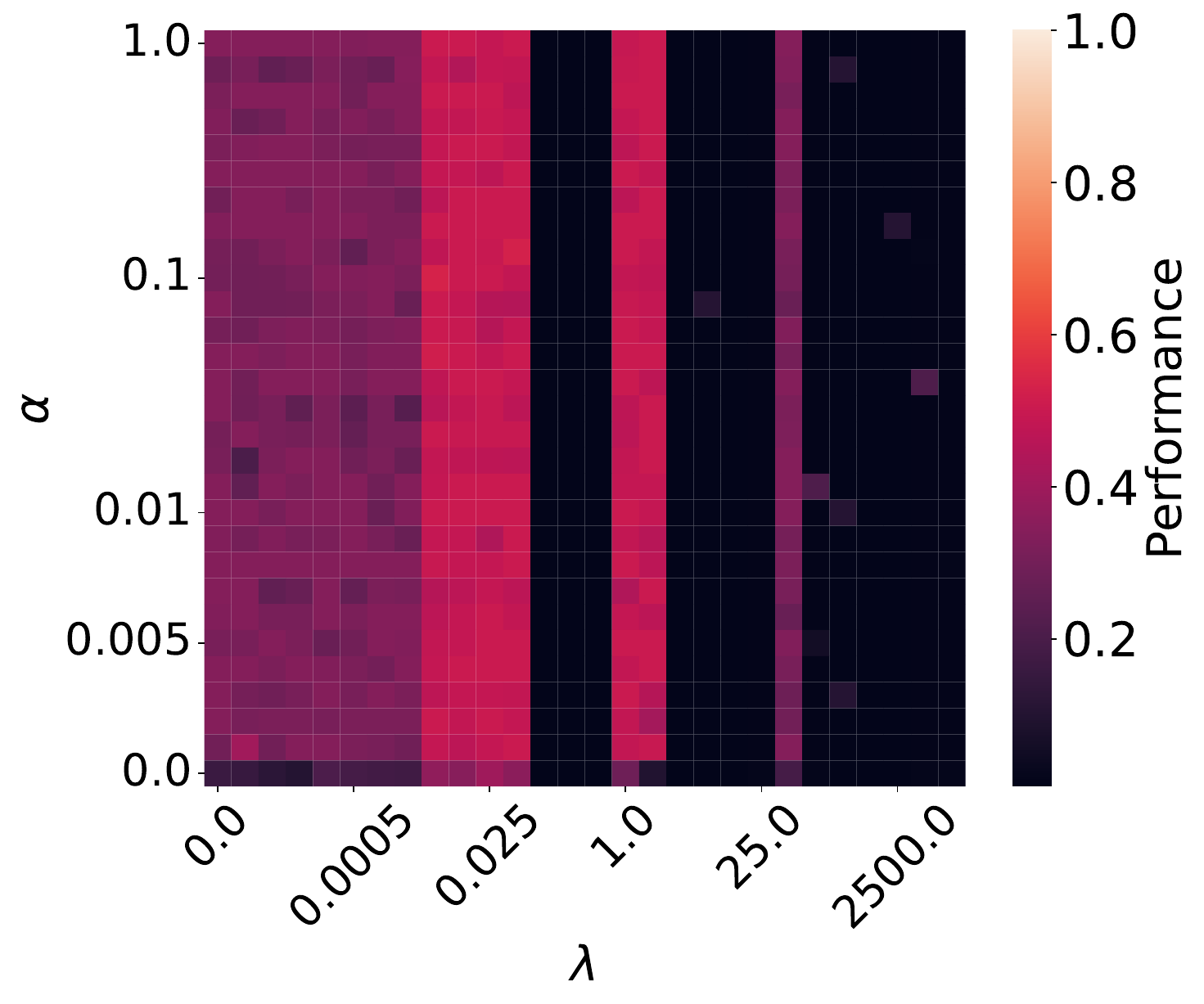}
        \caption{$\alpha=\mathrm{learned}$\\lin. anneal, \\$\lambda=\mathrm{lin.\,anneal}$}
    \end{subfigure}
    \caption{MDPO($-\mathcal{H}_{0.5}, B_{-\mathcal{H}_{0.5}}$) for different temperature scheduling schemes}
    \label{fig:App08}
\end{figure}

% % Rescaled CartPole on Different Environments

\begin{figure}[htbp]
    \centering
    \begin{subfigure}[b]{0.24\textwidth}
        \centering
        \includegraphics[width=\textwidth]{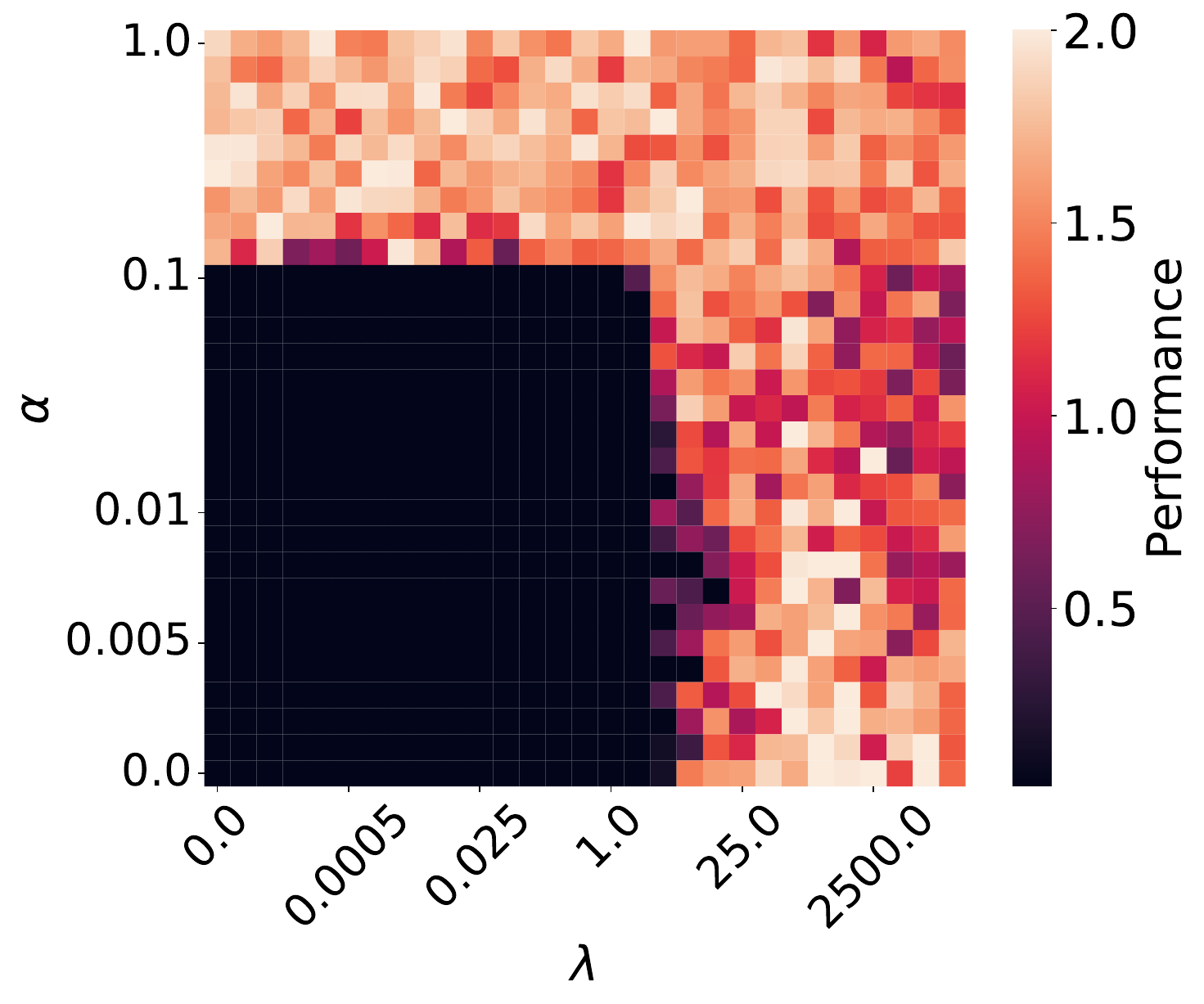}
        \caption{Maximum Return = $1000$}
    \end{subfigure}
    \hfill
    \begin{subfigure}[b]{0.24\textwidth}
        \centering
        \includegraphics[width=\textwidth]{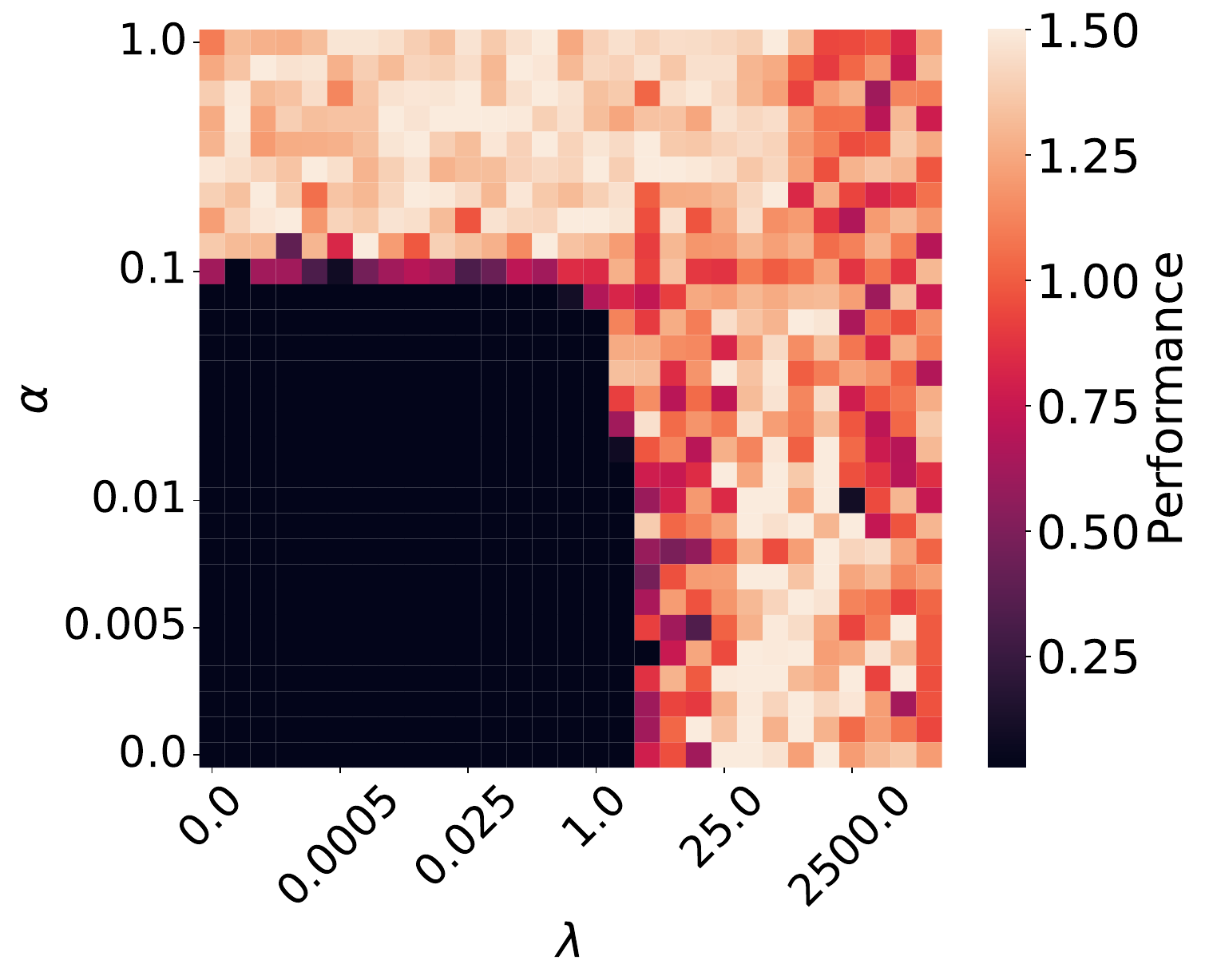}
        \caption{Maximum Return = $750$}
    \end{subfigure}
    \hfill
    \begin{subfigure}[b]{0.24\textwidth}
        \centering
        \includegraphics[width=\textwidth]{images_supplemental/Heatmap_AE200a-0.pdf}
        \caption{Maximum Return = $500$}
    \end{subfigure}
    \hfill
    \begin{subfigure}[b]{0.24\textwidth}
        \centering
        \includegraphics[width=\textwidth]{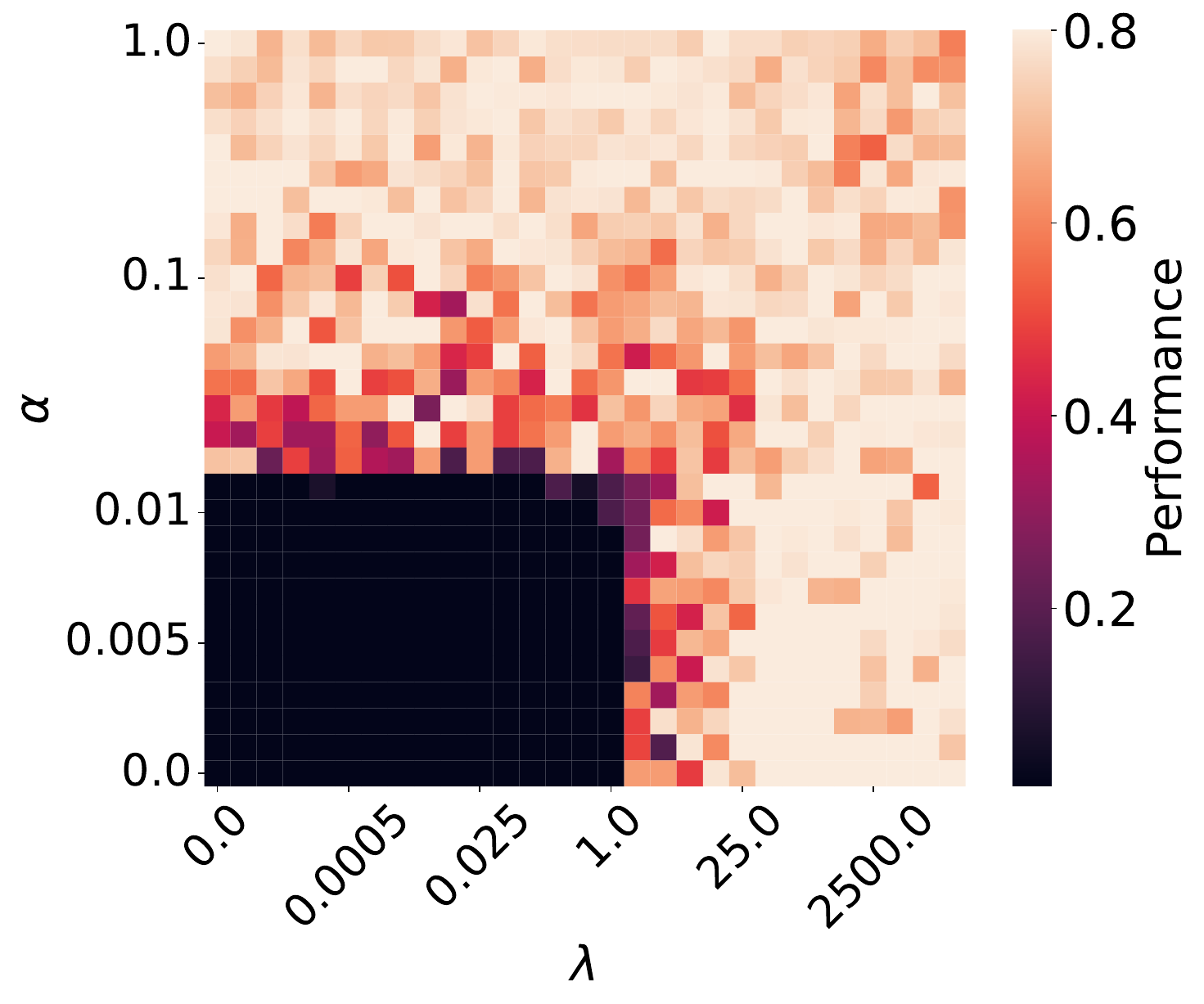}
        \caption{Maximum Return = $400$}
    \end{subfigure}
    \caption{MDPO($-\mathcal{H}, D_\mathrm{KL}$) with constant temperatures on CartPole with different maximum returns}
    \label{fig:App09}
\end{figure}

\begin{figure}[htbp]
    \centering
    \begin{subfigure}[b]{0.24\textwidth}
        \centering
        \includegraphics[width=\textwidth]{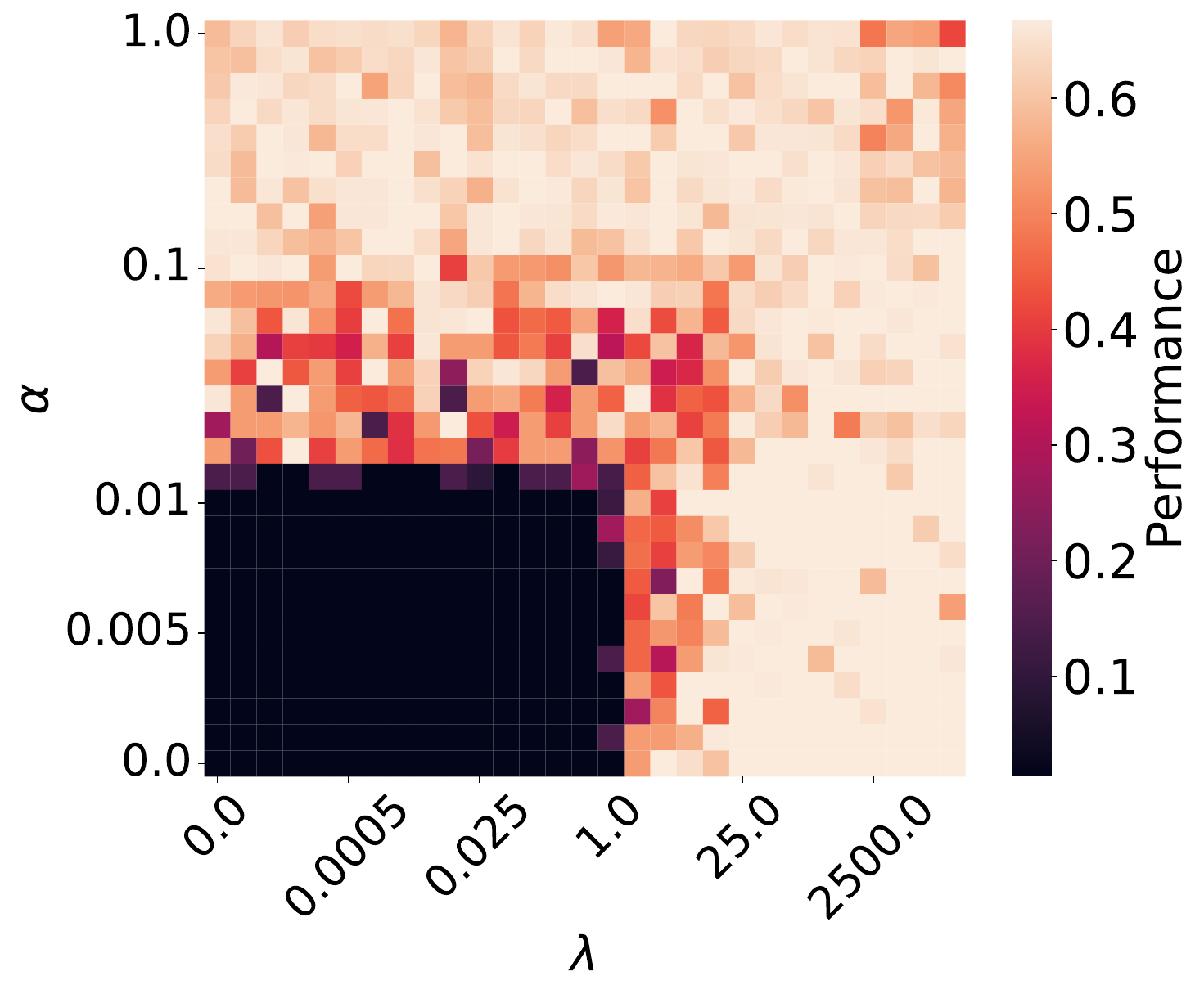}
        \caption{Maximum Return = $333$}
    \end{subfigure}
    \hfill
    \begin{subfigure}[b]{0.24\textwidth}
        \centering
        \includegraphics[width=\textwidth]{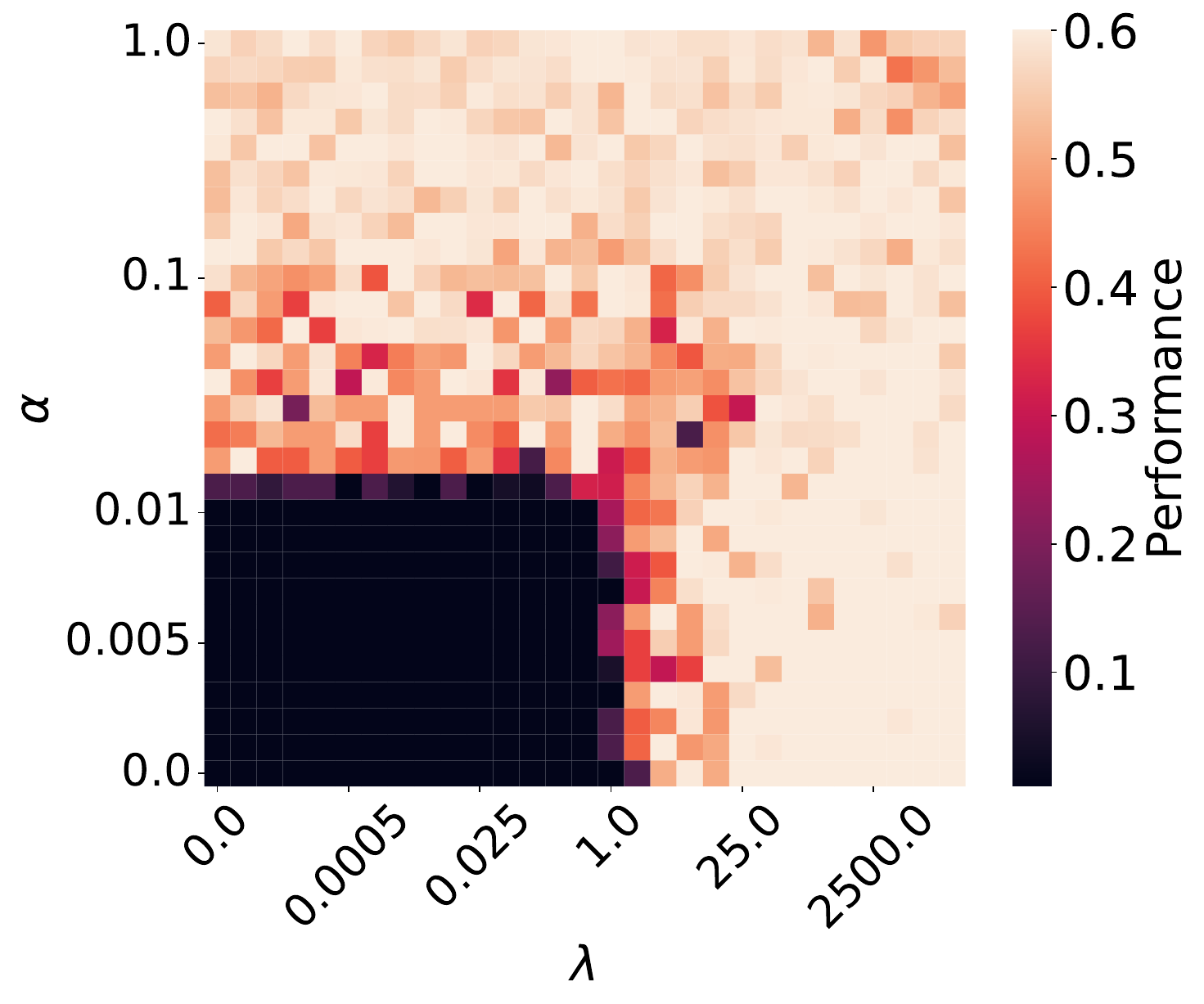}
        \caption{Maximum Return = $300$}
    \end{subfigure}
    \hfill
    \begin{subfigure}[b]{0.24\textwidth}
        \centering
        \includegraphics[width=\textwidth]{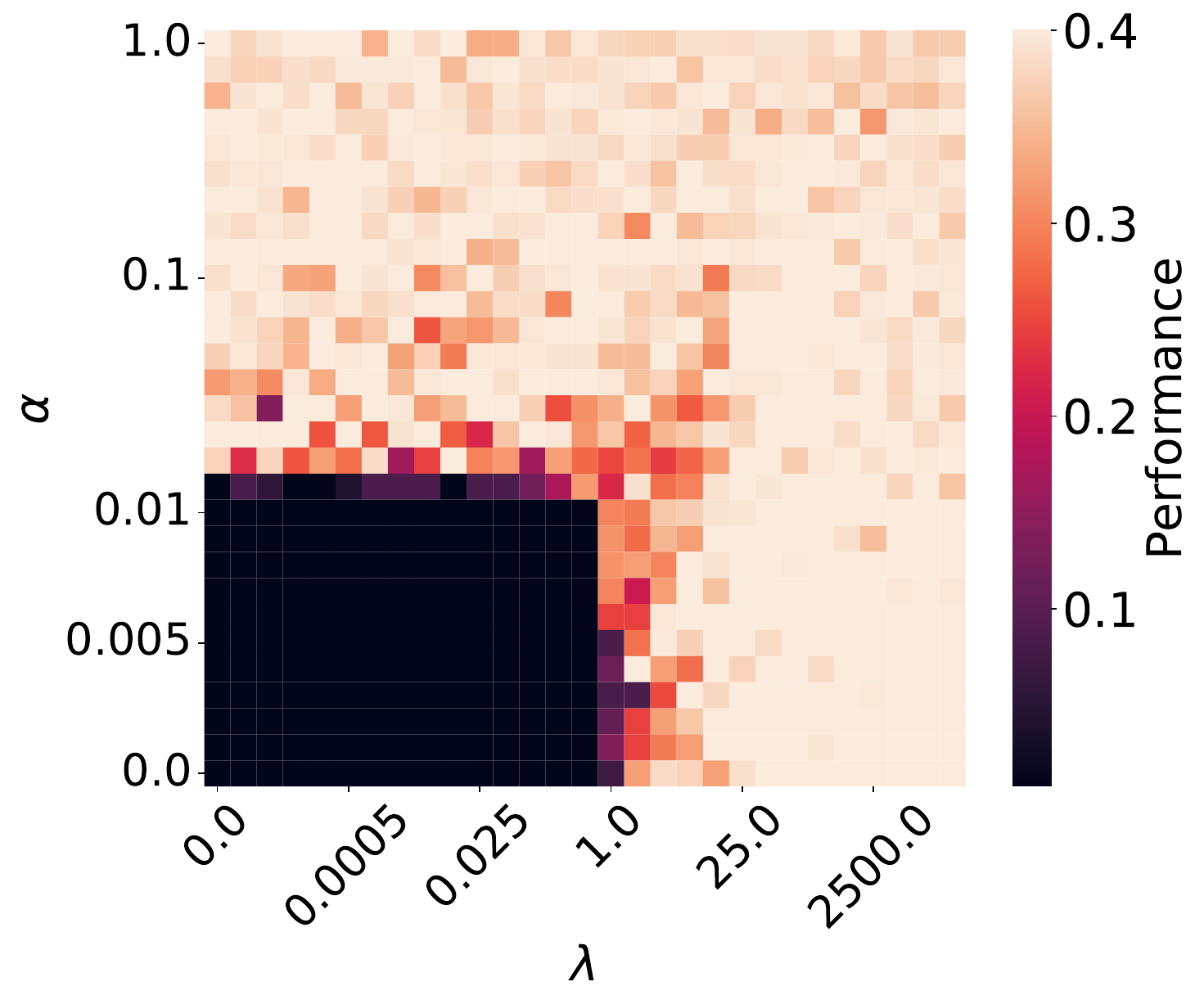}
        \caption{Maximum Return = $200$}
    \end{subfigure}
    \hfill
    \begin{subfigure}[b]{0.24\textwidth}
        \centering
        \includegraphics[width=\textwidth]{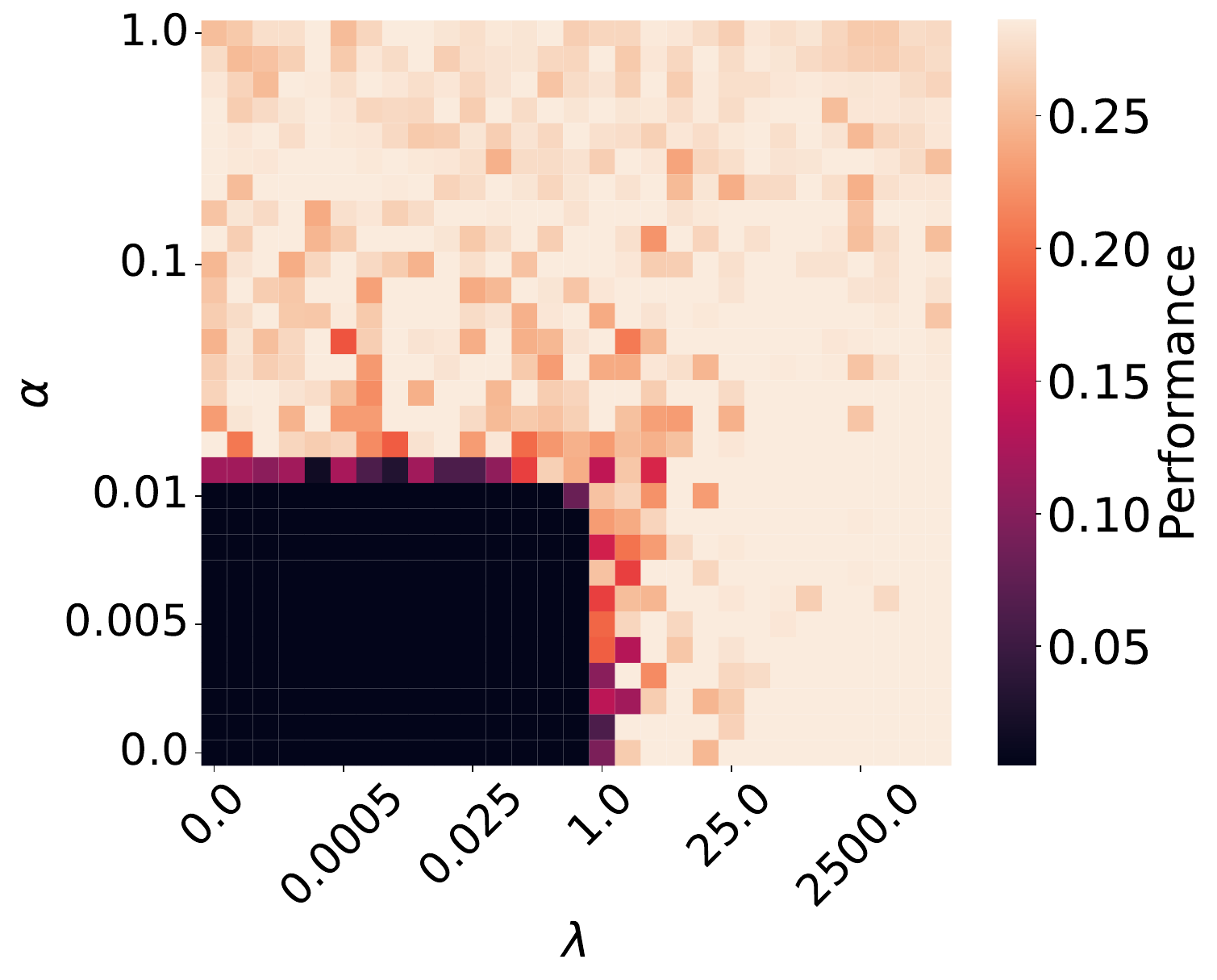}
        \caption{Maximum Return = $143$}
    \end{subfigure}
    \caption{MDPO($-\mathcal{H}, D_\mathrm{KL}$) with constant temperatures on CartPole with different maximum returns}
    \label{fig:App10}
\end{figure}

\begin{figure}[htbp]
    \centering
    \begin{subfigure}[b]{0.32\textwidth}
        \centering
        \includegraphics[width=\textwidth]{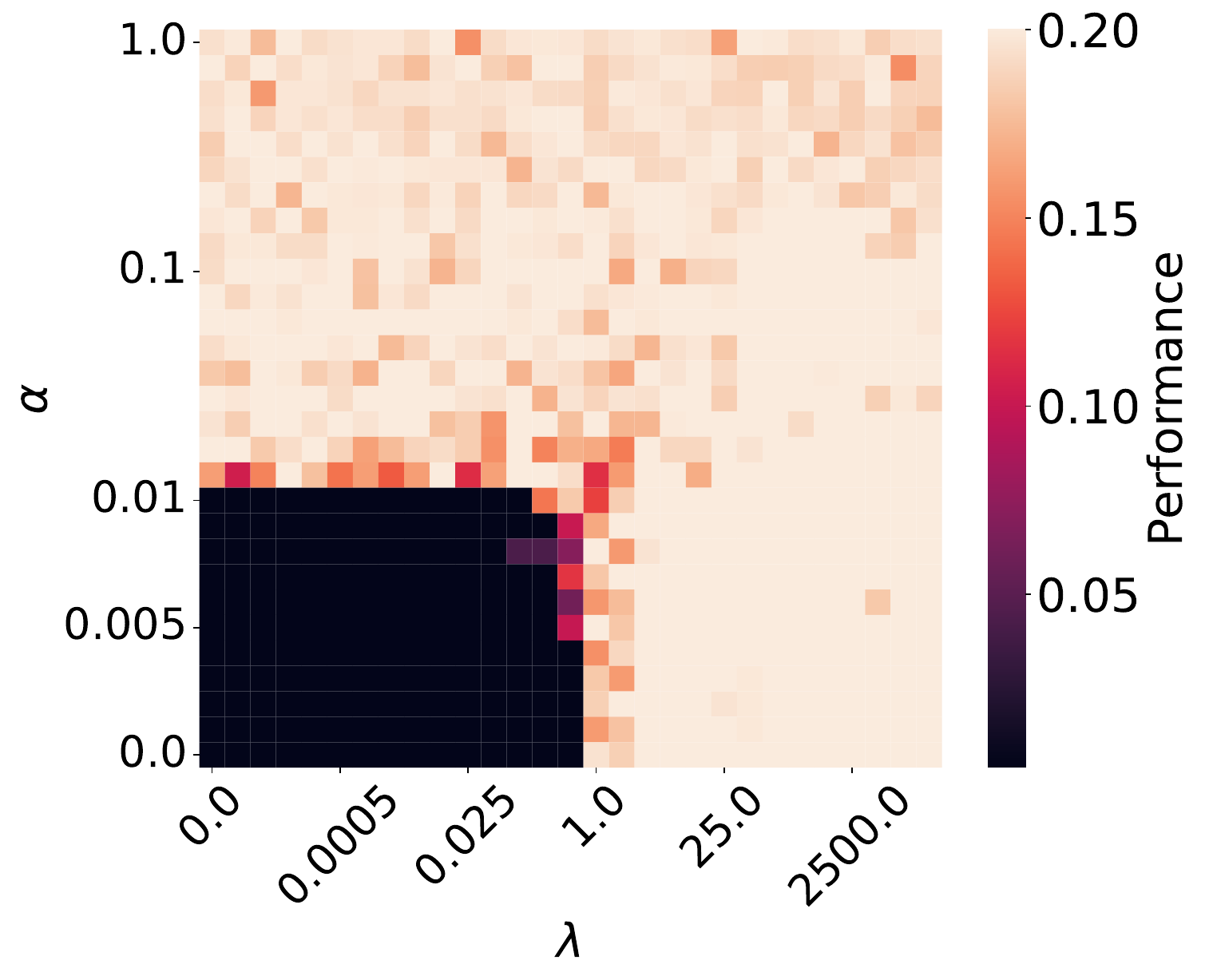}
        \caption{Maximum Return = $100$}
    \end{subfigure}
    \hfill
    \begin{subfigure}[b]{0.32\textwidth}
        \centering
        \includegraphics[width=\textwidth]{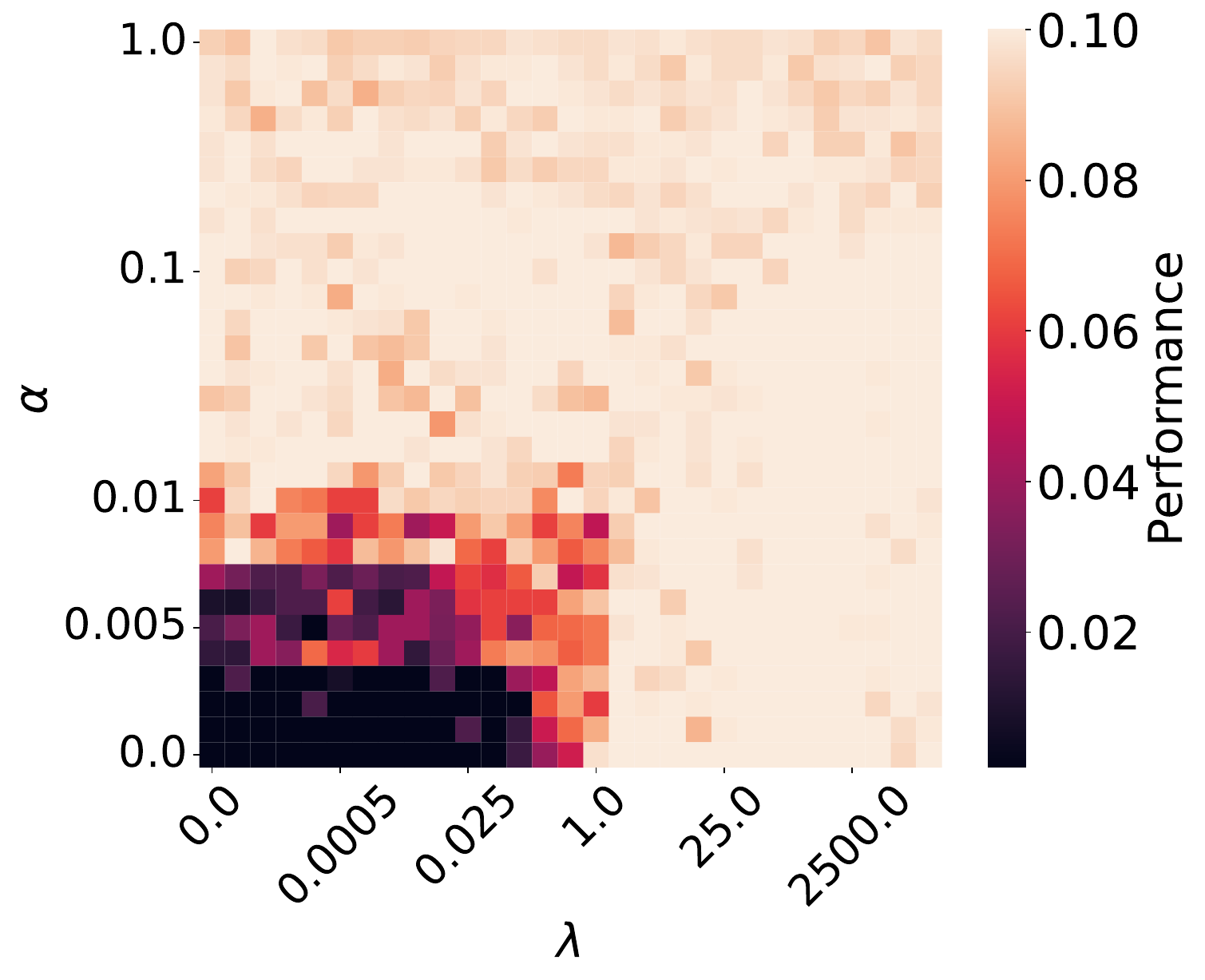}
        \caption{Maximum Return = $50$}
    \end{subfigure}
    \hfill
    \begin{subfigure}[b]{0.32\textwidth}
        \centering
        \includegraphics[width=\textwidth]{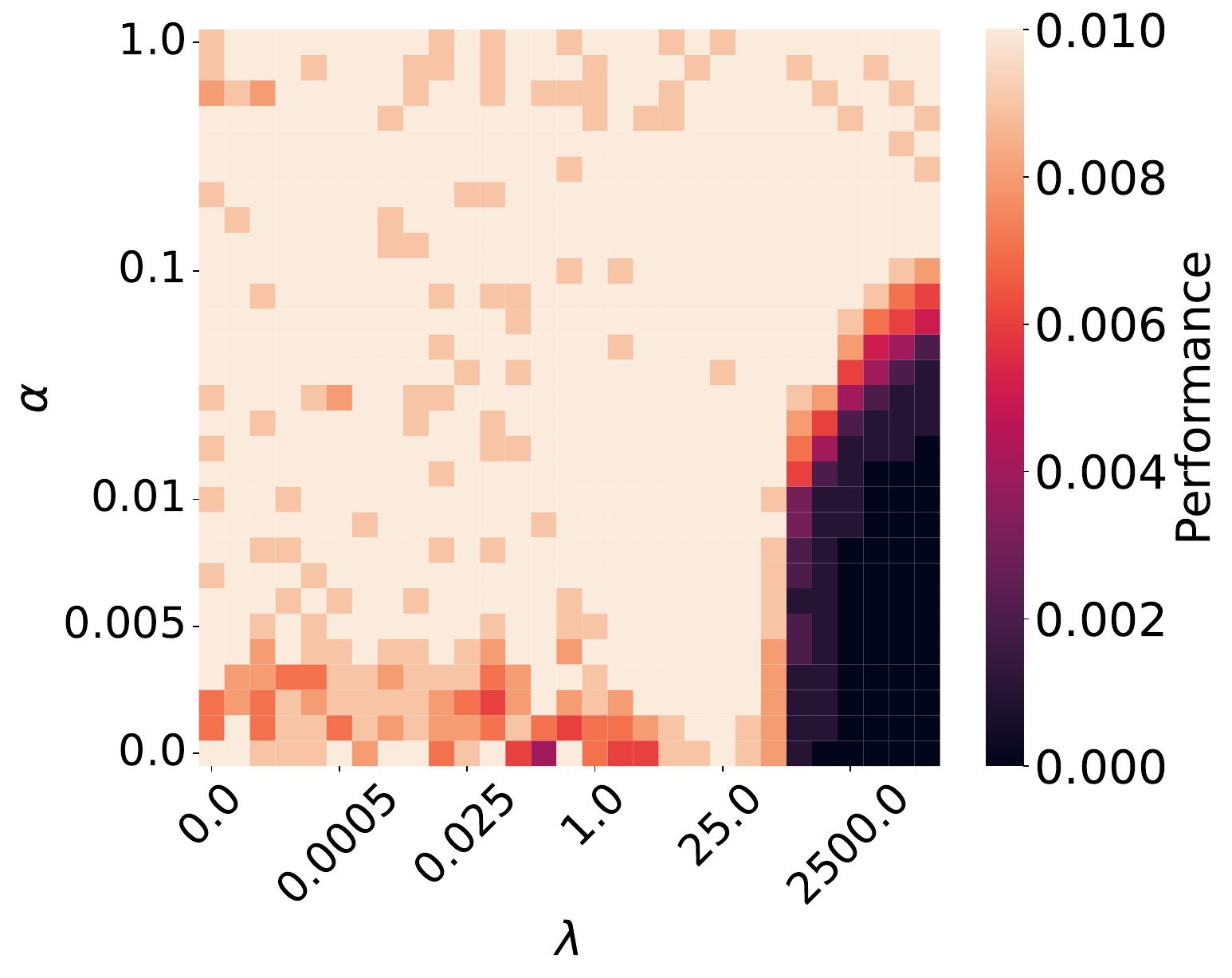}
        \caption{Maximum Return = $5$}
    \end{subfigure}
    \caption{MDPO($-\mathcal{H}, D_\mathrm{KL}$) with constant temperatures on CartPole with different maximum returns}
    \label{fig:App11}
\end{figure}

% \newpage
% \subsection*{Robustness}

% \begin{figure}[htbp]
%     \centering
%     \begin{subfigure}[b]{0.32\textwidth}
%         \centering
%         \includegraphics[width=\textwidth]{images_final/Rob01.pdf}
%         \caption{Starting at $\tau=0.4$}
%     \end{subfigure}
%     \hfill
%     \begin{subfigure}[b]{0.32\textwidth}
%         \centering
%         \includegraphics[width=\textwidth]{images_final/Rob02.pdf}
%         \caption{Starting at $\tau=0.6$}
%     \end{subfigure}
%     \hfill
%     \begin{subfigure}[b]{0.32\textwidth}
%         \centering
%         \includegraphics[width=\textwidth]{images_final/Rob03.pdf}
%         \caption{Starting at $\tau=0.8$}
%     \end{subfigure}
%     \caption{Performance frequency curves starting at different performance levels}
%     \label{fig:App12}
% \end{figure}

\end{document}